\pdfoutput=1 %
\documentclass{article} %

\usepackage{iclr2027_conference,times}

\usepackage[T1]{fontenc}    %
\usepackage{hyperref}
\usepackage{url}
\usepackage{fontawesome5}
\usepackage{uttt_link_isolation}
\hypersetup{
  colorlinks,
  linkcolor={blue!55!black},
  citecolor={blue!55!black},
  urlcolor={blue!55!black},
}

\usepackage{booktabs}       %
\usepackage{array}          %
\usepackage{amsmath}        %
\usepackage{amsfonts}       %
\newtheorem{proposition}{Proposition}
\usepackage{nicefrac}       %
\usepackage{graphicx}       %
\usepackage{subcaption}     %
\usepackage{xcolor}         %
\usepackage{colortbl}       %
\usepackage{tabularx}       %
\usepackage{placeins}       %
\usepackage{tikz}           %
\usepackage{amssymb}
\usetikzlibrary{backgrounds}

\newcommand{\floattile}[1]{%
  \begin{tikzpicture}
    \node[inner sep=6pt] (tile) {#1};
    \begin{scope}[on background layer]
      \foreach \i in {8,7,6,5,4,3,2,1} {
        \fill[black, opacity=0.012, rounded corners=1.2mm]
          ([xshift=-0.45pt*\i, yshift=-0.75pt*\i]tile.south west) rectangle
          ([xshift=0.45pt*\i, yshift=0.15pt*\i]tile.north east);
      }
      \fill[white, rounded corners=1.2mm] (tile.south west) rectangle (tile.north east);
    \end{scope}
  \end{tikzpicture}}

\definecolor{rowblock}{HTML}{E6F2EC}
\definecolor{rowhi}{HTML}{DFE9F7}
\definecolor{ttd-base}{HTML}{0B6B78}\definecolor{tth-base}{HTML}{E8F5F7}\definecolor{ttb-base}{HTML}{F4FAFB}
\definecolor{ttd-att}{HTML}{444444}\definecolor{tth-att}{HTML}{F0F0F0}\definecolor{ttb-att}{HTML}{F8F8F8}
\definecolor{ttd-lin}{HTML}{0B6B78}\definecolor{tth-lin}{HTML}{E8F5F7}\definecolor{ttb-lin}{HTML}{F4FAFB}
\definecolor{ttd-mamba}{HTML}{66701A}\definecolor{tth-mamba}{HTML}{F4F5EA}\definecolor{ttb-mamba}{HTML}{F9FAF4}
\definecolor{ttd-fw}{HTML}{5A3FA0}\definecolor{tth-fw}{HTML}{F2EFF9}\definecolor{ttb-fw}{HTML}{F8F7FC}\definecolor{ttx-fw}{HTML}{E5DEF2}
\definecolor{ttx-ours}{HTML}{FFD9EE}
\definecolor{ttd-moeA}{HTML}{006d2c}\definecolor{tth-moeA}{HTML}{E6F0EA}\definecolor{ttb-moeA}{HTML}{F2F8F4}
\definecolor{ttd-denA}{HTML}{3b1f8f}\definecolor{tth-denA}{HTML}{EBE9F4}\definecolor{ttb-denA}{HTML}{F5F4F9}
\definecolor{ttd-moeB}{HTML}{08519c}\definecolor{tth-moeB}{HTML}{E6EEF5}\definecolor{ttb-moeB}{HTML}{F3F7FA}\definecolor{ttx-moeB}{HTML}{CEDCEB}
\definecolor{ttd-denB}{HTML}{8c2d04}\definecolor{tth-denB}{HTML}{FBEDE6}\definecolor{ttb-denB}{HTML}{FDF6F3}
\definecolor{ttd-moeC}{HTML}{0e7c86}\definecolor{tth-moeC}{HTML}{E7F2F3}\definecolor{ttb-moeC}{HTML}{F3F8F9}
\definecolor{ttd-ref}{HTML}{444444}\definecolor{tth-ref}{HTML}{F0F0F0}\definecolor{ttb-ref}{HTML}{F8F8F8}
\newcommand{\tgroup}[3]{\rowcolor{tth-#1} \multicolumn{#2}{c}{\color{ttd-#1}\bfseries\boldmath #3} \\}
\newcommand{\trow}[1]{}%
\newcommand{\thl}[1]{\rowcolor{ttx-#1}}
\newcommand{\tsep}{\midrule}%
\makeatletter
\newcommand{\ttVgapwidth}{\dimexpr 12pt-2\tabcolsep\relax}
\newcommand{\ttVgap}{%
  \ifx\CT@row@color\relax
    \hskip\ttVgapwidth
  \else
    \begingroup
      \def\CT@color{\@ifnextchar[\ttVcolorA\ttVcolorB}%
      \CT@row@color
    \endgroup
  \fi}
\def\ttVcolorA[#1]#2{\ttVpaint{\color[#1]{#2}}}
\def\ttVcolorB#1{\ttVpaint{\color{#1}}}
\def\ttVpaint#1{\kern-.3pt{#1\vrule\@width\dimexpr\ttVgapwidth+.6pt\relax}\kern-.3pt}
\makeatother
\newcolumntype{V}{@{\hskip\tabcolsep\ttVgap\hskip\tabcolsep}}

\title{Universal Test-Time Training}

\author{
Zefan Cai$^{1,*}$ \quad Qinzhe Hu$^{1,*}$ \quad Ziqiao Ma$^{2}$ \quad Hao Tan$^{3}$ \quad Junjie Hu$^{1}$ \\[12pt]
\normalfont\normalsize
$^{1}$University of Wisconsin--Madison \quad $^{2}$University of Michigan--Ann Arbor \\
$^{3}$Adobe \\[5pt]
\multicolumn{1}{l}{\normalfont\small
\href{https://zefan-cai.github.io/uTTT.github.io/}{\faGlobe\enspace Project Page} \qquad
\href{https://github.com/Zefan-Cai/uTTT}{\faGithub\enspace GitHub}}
}

\iclrfinalcopy

\begin{document}

\maketitle
\addtocontents{toc}{\protect\setcounter{tocdepth}{-1}}   %
\lhead{Preprint.}
\begingroup
\renewcommand{\thefootnote}{\fnsymbol{footnote}}
\footnotetext[1]{Equal contribution.}
\endgroup

\begin{abstract}
Recent Test-Time Training (TTT) architectures compress context into fast weights that are updated online and queried as memory. Existing TTT designs keep this memory private to each layer: it recurs only over time, and depth merely indexes $L$ separate memories. We argue that memory ownership need not be tied to depth, and introduce \emph{Universal Test-Time Training (uTTT)}, in which all layers read and write one shared memory while retaining layer-specific backbone parameters. The shared memory thus recurs over two dimensions, time and depth, with chunks and layers as their units: a write by a deep layer in one chunk can be read by a shallow layer in the next. We instantiate this idea as \emph{uTTT-MoE} and \emph{uTTT-Dense}. uTTT-MoE routes each token head to a few experts in a pool shared by all layers; uTTT-Dense applies the whole shared memory at every layer without routing. In language modeling, uTTT-MoE reaches 15.5 and 27.9 RULER accuracy at 124M and 760M, 2.6 and 2.1 points above its layer-private counterpart at equal state and active compute, the highest among tested bounded-state models, with per-token loss matching or beating full attention. In novel view synthesis, sharing at fixed per-layer compute gains 0.92 dB in view-23 object PSNR in routed models and 0.76 dB in dense models.
\end{abstract}

\begingroup
\centering
\begin{minipage}{\textwidth}
\centering
\floattile{\includegraphics[width=\dimexpr\textwidth-20pt\relax]{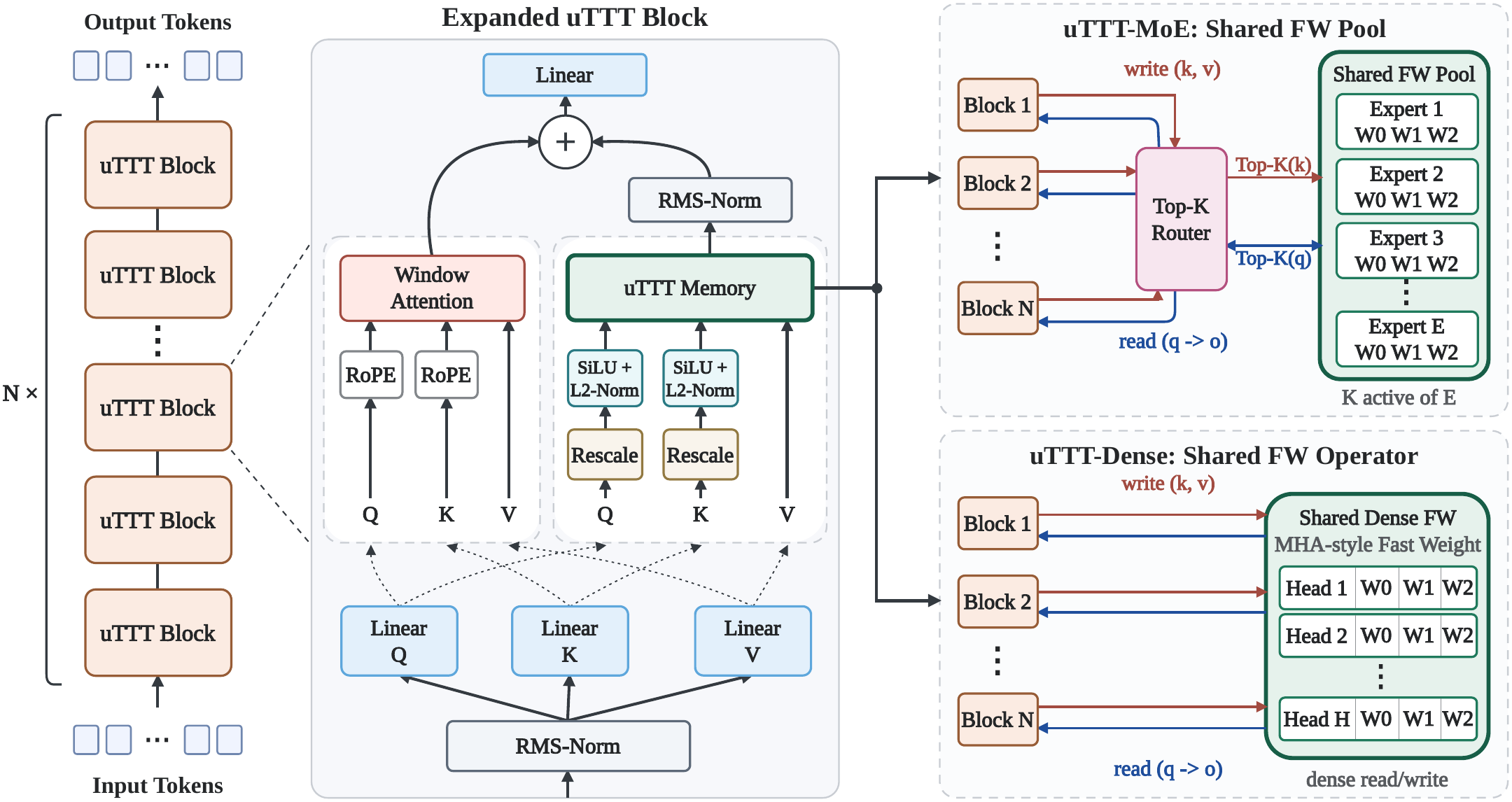}}
\captionof{figure}{\textbf{Universal Test-Time Training.} A block (\emph{middle}) keeps its normalization, projections, and window attention private and sends $q,k,v$ to a memory with state shared across the stack (\emph{left}). The state (\emph{right}) is a pool of $E$ fast-weight experts under a top-$K$ router (uTTT-MoE, top) or dense fast weights applied in full (uTTT-Dense, bottom). Blue arrows read \eqref{eq:uttt-read}; red arrows write \eqref{eq:uttt-update}.}
\label{fig:uttt-main}
\end{minipage}
\par\endgroup

\section{Introduction}
\label{sec:intro}

Long-context models trade explicit memory for compressed state. Attention keeps the full key-value history at growing cost~\citep{vaswani2017attention}, while recurrent networks, linear attention, and state-space models keep a bounded state~\citep{hochreiter1997long,katharopoulos2020transformers,gu2024mamba,dao2024transformersssms,yang2024parallelizing}. Recent Test-Time Training (TTT) architectures store context in \emph{fast weights} that are optimized online during the forward pass and queried as memory~\citep{hinton1987fastweights,schmidhuber1992learning,schlag2021linear,sun2025learninglearntesttime,behrouz2024titans}; large-chunk TTT amortizes these updates to make nonlinear fast-weight memory practical at long context~\citep{zhang2025testtimetrainingright}.

Once fast weights serve as memory, a design question is \emph{who owns them}. Layerwise KV-binding TTT~\citep{sun2025learninglearntesttime} and LaCT~\citep{zhang2025testtimetrainingright} give each layer its own state, written using its key-value pairs and read by its queries. TTT-E2E~\citep{tandon2025endtoendtesttimetraininglong} updates selected layers' MLP weights under one next-token prediction loss, but still keeps separate weight tensors across depth. In all of these designs, layer $\ell$'s memory takes only layer $\ell$'s writes, so it evolves along time alone and depth only multiplies memories: a layer can use only its own share of the budget, and no layer can read state written by another.

In this work, we call fast weights \emph{universal} when all layers read from and write to one shared state or one shared pool of states. \emph{Universal Test-Time Training (uTTT)} builds on such fast weights while keeping backbone parameters layer-specific (Figure~\ref{fig:uttt-main}). Its memory then evolves along both time and depth, measured in chunks along time and in layers along depth, and layer-private TTT is the special case that partitions it by layer. This opens two routes that layer-private TTT lacks: a shallow layer can directly read what a deep layer wrote in the previous chunk, a \emph{top-down route}, and the outer task loss at that read backpropagates through the write, so the slow weights learn how layers at different depths build and use one memory (Sections~\ref{sec:preliminary} and~\ref{sec:method:recurrent}). Unlike prior sharing across depth, uTTT shares a bounded fast-weight memory that every layer writes by test-time optimization within each sequence, and keeps these writes in the training graph.

uTTT comes in two forms. \emph{uTTT-MoE} gives every layer a shared pool of $E$ expert states, from which each token head selects $K$ per read or write. Its control, \emph{TTT-MoE}, uses the same router but restricts each layer to its own $E/L$ experts, so sharing widens each layer's choices from $E/L$ to $E$ experts at equal total state and active experts. \emph{uTTT-Dense} applies one dense shared state at every layer. Compared with \emph{TTT-Dense}, the layer-private dense design of LaCT, it lets every layer apply the whole state at a fixed budget, raising fast-weight compute, or matches per-layer compute with only $1/L$ of the state. Because all layers write the same tensors, each chunk's updates also merge into one operation, which speeds up training.

\paragraph{Related work.}
\label{sec:related}

TTT uses fast weights as a context-dependent memory state~\citep{sun2023learningtolearn,sun2025learninglearntesttime,vonoswald2025mesanet,wang2025testtimeregression}. MoM routes among linear memory states~\citep{du2025mom}; our TTT-MoE control adds routing to LaCT's nonlinear fast weights. Shared memory also appears in Feedback Transformers~\citep{fan2020feedback}, block-recurrent models~\citep{ju2021staircase,hutchins2022block}, and cross-layer key--value caches~\citep{sun2024yoco,wu2024layercondensed,brandon2024cla}. Universal Transformers, ALBERT and their MoE variants share slow parameters across depth~\citep{dehghani2019universal,lan2020albert,tan2023sut,csordas2024moeut}. UniPool and MoUE likewise share trained expert parameters~\citep{huang2026unipool,chen2026moue}; MoUE additionally shares an online-updated routing state. Appendix~\ref{app:related} gives the extended comparison.

\paragraph{Contributions.}
\begin{itemize}
    \setlength{\itemsep}{1pt}\setlength{\parsep}{0pt}
    \item We propose uTTT, a TTT architecture whose fast-weight memory is shared by all layers, in routed (uTTT-MoE) and dense (uTTT-Dense) forms, with one causal update per chunk for all layers' writes and a layer-private routed control, TTT-MoE, that isolates sharing from routing.
    \item We find that sharing outperforms layer-private TTT at equal or smaller state. uTTT-MoE gains $2.6$ and $2.1$ RULER points over TTT-MoE at 124M and 760M, leading the tested bounded-state models (Figure~\ref{fig:teaser-pareto}), and sharing adds $0.92$ and $0.76$~dB object PSNR in routed and dense view synthesis. uTTT-MoE also trains up to $1.39\times$ faster than TTT-MoE.
    \item We show through analysis and interventions that sharing unifies time and depth in one memory recurrence, with layer-private TTT as its partitioned special case (Section~\ref{sec:method:recurrent}). A read can depend on every layer's earlier writes: probes detect these cross-layer routes in trained models, and removing them, by knocking out cross-layer reads at inference or cutting cross-depth gradients in training, degrades performance (Figure~\ref{fig:core-evidence}).
\end{itemize}

\section{Preliminary}
\label{sec:preliminary}

\paragraph{LaCT and fast-weight state.}
Large-chunk test-time training (LaCT)~\citep{zhang2025testtimetrainingright} equips each layer with a module that stores context in fast weights. We use \emph{memory} for this function and \emph{state} for its concrete, updatable representation. We partition each sequence into contiguous token blocks, called chunks, $C_1,\ldots,C_N$, and process them in causal order. At layer $\ell\in\{1,\ldots,L\}$ and fast-weight head $h$ (a \emph{layer-head}), slow-parameter projections produce query, key and value vectors $\mathbf q_{\ell,h,i},\mathbf k_{\ell,h,i},\mathbf v_{\ell,h,i}\in\mathbb R^{d_h}$ for token $i$.

Suppressing these indices, one SwiGLU~\citep{shazeer2020glu} module has state $W=(W_0,W_1,W_2)$, with $W_0,W_2\in\mathbb R^{d_{\rm fw}\times d_h}$ and $W_1\in\mathbb R^{d_h\times d_{\rm fw}}$, and read function $f_W(\mathbf x)=W_1[\operatorname{SiLU}(W_0\mathbf x)\odot(W_2\mathbf x)]$. Here $d_{\rm fw}$ is its internal width and $\odot$ elementwise multiplication. A model's fast-weight state collects these matrix triples; its recurrent implementation also carries master accumulators and, where used, momentum buffers. LaCT keeps each layer's state private. Chunk-level writes amortize update costs; a read itself leaves the state unchanged and supplies features used for task predictions.

\paragraph{Inner loop: associative writes.}
A read evaluates $f_W$ on a query. A write differentiates the associative loss $\ell_{\rm in}(W;\mathbf k,\mathbf v)=-\langle f_W(\mathbf k),\mathbf v\rangle$. Within chunk $c$, reads use the state produced after chunk $c-1$; a boundary update aggregates token contributions:
\begin{equation}
\begin{aligned}
\mathbf o_i&=f_{W^{(c-1)}}(\mathbf q_i),\qquad i\in C_c,\\
W^{(c)}&=\operatorname{Update}\!\left(W^{(c-1)},\;
\left.\nabla_W\sum_{i\in C_c}\eta_i\ell_{\rm in}(W;\mathbf k_i,\mathbf v_i)\right|_{W=W^{(c-1)}}\right).
\end{aligned}
\label{eq:prelim-inner-loop}
\end{equation}
The effective step sizes $\eta_i$ come from slow-parameter predictors; matrix-specific coefficients and fixed reductions are suppressed. $\operatorname{Update}$ denotes a stateful optimizer carrying master accumulators and any momentum buffers, with predicted momentum coefficients also implicit. Language models use momentum, Muon and row renormalization; novel view synthesis (NVS) omits momentum and Muon (Appendix~\ref{app:fw-update}). Inner updates run during training and inference. Fast weights reset to their learned initialization between sequences; the final language-modeling chunk needs no write.

\paragraph{Outer loop: learning to write and read.}
During training, the task loss $\mathcal L_{\rm task}$ supervises predictions produced using the read outputs. Along the state chain, backpropagation follows $\mathcal L_{\rm task}\to\mathbf o\to W^{(c-1)}\to\operatorname{Update}_{s<c}\to\theta$, reaching slow parameters $\theta$ through earlier differentiable writes, including their inner-loss gradient computation. This trains the slow projections, coefficient predictors and initialization to produce useful future reads. The task loss supplies the outer learning signal; the associative loss determines each online write. AdamW applies these outer gradients across training sequences. At inference the slow parameters remain fixed: online inner updates still adapt the state, but no outer task-loss update is performed. Thus the two loops optimize different quantities: the inner loop changes sequence-specific state; the outer loop learns how that adaptation supports the task.

\section{Universal Fast Weights}
\label{sec:method}

Universal Test-Time Training (uTTT) shares the fast-weight state across depth while retaining layer-specific backbone weights, projections and local attention (Figure~\ref{fig:uttt-main}).

\subsection{Shared Fast Weights}
\label{sec:method:universal}
The layer-private construction in Section~\ref{sec:preliminary} gives each layer its own state. uTTT instead partitions the $L$ layers into $p$ contiguous groups and gives each group one shared state, a collection of the SwiGLU matrix triples defined above. Every layer in a group reads that state and contributes write inputs to it. At $p=L$ the groups are layer-private; at $p=1$ all depths share one state. The following instantiations specify how the modules $f_W$ access its components.

Let $\mathcal{B} = \{W_e\}_{e=1}^{E}$ be this pool, each $W_e$ one SwiGLU triple, either a routed expert or a dense head block. The group of layer $\ell$ owns a disjoint part of the pool, its \emph{visible domain} $\Omega_\ell\subseteq\{1,\dots,E\}$, and each layer-head $(\ell,h)$ routes a feature $z$ into it, $\mathcal{R}_{\ell,h}(z) \subseteq \Omega_\ell$. Groups are equal-sized, with integer $p$ satisfying $1\le p\le L$ and $p\mid L$; routed pools also require $p\mid E$ and integer $K$ with $1\le K\le E/p$. Table~\ref{tab:design-space-full} lists both families along $p$.

\begin{table}[tb]
    \centering
    \small
    \setlength{\tabcolsep}{4pt}
    \caption{\textbf{Designs along the group count $p$.} $H$ counts fast-weight heads and $S$ is one private dense layer's full state. $E$ counts SwiGLU triples of size $|W_e|$; layer $\ell$ reaches $|\Omega_\ell|\,|W_e|$, while Active counts the state applied by one token head. The fixed-compute dense row stores $pS$ at unchanged per-token fast-weight compute; the other dense rows preserve total state $LS$, and routed rows preserve $Eu$. The uTTT-Dense row is size-matched. A write-enabled chunk ends in one update per group, \eqref{eq:aggregated}.}
    \label{tab:design-space-full}
    \begin{tabular*}{\textwidth}{@{\extracolsep{\fill}}lccccccc@{}}
        \toprule
        Design & $p$ & $E$ & $|W_e|$ & $|\Omega_\ell|$ & Reachable & Active & Memories \\
        \midrule
        TTT-Dense (LaCT) & $L$ & $LH$ & $S/H$ & $H$ & $S$ & $S/H$ & $L$ \\
        Dense, fixed compute & $p$ & $pH$ & $S/H$ & $H$ & $S$ & $S/H$ & $p$ \\
        Dense, fixed state & $p$ & $pH$ & $LS/(pH)$ & $H$ & $LS/p$ & $LS/(pH)$ & $p$ \\
        uTTT-Dense & $1$ & $H$ & $LS/H$ & $H$ & $LS$ & $LS/H$ & $1$ \\
        \midrule
        TTT-MoE & $L$ & $E$ & $u$ & $E/L$ & $Eu/L$ & $Ku$ & $L$ \\
        Routed, $p$ groups & $p$ & $E$ & $u$ & $E/p$ & $Eu/p$ & $Ku$ & $p$ \\
        uTTT-MoE & $1$ & $E$ & $u$ & $E$ & $Eu$ & $Ku$ & $1$ \\
        \bottomrule
    \end{tabular*}
\end{table}

We use an \emph{aggregated} write schedule: all layers process chunk $c$ using the fast weights present before that chunk, and their inner-loss gradients are evaluated at those same weights. At the boundary, contributions are combined and one optimizer update is applied per group, including its recurrent buffers. Writing $G_{c,\ell}$ for the gradient term of \eqref{eq:prelim-inner-loop} contributed by layer $\ell$ in chunk $c$, evaluated at the chunk-start state, a group's state advances by the sum over its layers,
\begin{equation}
W^{(c)} = \operatorname{Update}\Big(W^{(c-1)},\, \textstyle\sum_{\ell} G_{c,\ell}\Big).
\label{eq:aggregated}
\end{equation}
No layer reads a write from its current chunk. Only write-enabled chunks update state; NVS target chunks containing only camera poses are read-only. Sharing changes which layers contribute to and use the state while retaining this causal protocol.

State size counts fast-weight elements, excluding master accumulators, optimizer buffers and attention caches. Total state is stored per sequence; reachable state is available to one layer, and active state is read or written by one token at one layer-head $(\ell,h)$. We distinguish these quantities because sharing can change their relationship without changing the read network.

\subsection{uTTT-MoE}
\label{sec:method:uttt-moe}
\label{sec:method:ttt-moe}
The routed instantiation makes every $W_e$ an expert state with $d_{\rm fw}=d_h$; a group owns $E/p$ experts. Each layer-head normalizes its routing feature $\hat z_{\ell,h}$ per head, the query for reads and the key for writes, and routes each token head to the $K$ experts of its visible domain with the largest inner product against learned expert addresses $r_{\ell,e}$,
\begin{equation}
\mathcal{R}_{\ell,h}(z) \;=\; \operatorname{TopK}\Big(\big\{ \hat z_{\ell,h}^{\!\top} r_{\ell,e} \big\}_{e \,\in\, \Omega_{\ell}},\; K\Big);
\label{eq:topk}
\end{equation}
the read and write selections are $\mathcal R^r_{\ell,h}(\mathbf q)=\mathcal R_{\ell,h}(\mathbf q)$ and $\mathcal R^w_{\ell,h}(\mathbf k)=\mathcal R_{\ell,h}(\mathbf k)$, and the language models learn the addresses per layer, while NVS shares them across layers, $r_{\ell,e}=r_e$. The softmax over these logits, kept for the selected experts, gives routing probabilities $p_{t,e}$ for token head $t=(\ell,h,i)$; the gates $g^r,g^w$ of \eqref{eq:uttt-read} and \eqref{eq:uttt-update} scale them by the \emph{routing-weight scale} $\gamma$,
\begin{equation}
g_{\ell,h,i,e} \;=\; \gamma \, p_{t,e} \qquad \text{for } e \in \mathcal{R}_{\ell,h}(z),\; t = (\ell,h,i),
\label{eq:gate-scale}
\end{equation}
with $\gamma=1$ everywhere except the sweep of Table~\ref{tab:lm-alpha}. The gates come from \emph{slow-parameter routers} and are not parameters of the online-updated expert state.
\begin{equation}
\mathbf o_{\ell,h,i}=\sum_{e\in\mathcal R^r_{\ell,h}(\mathbf q_{\ell,h,i})}
 g^r_{\ell,h,i,e}\,f_{W_e^{(c-1)}}(\mathbf q_{\ell,h,i}).
\label{eq:uttt-read}
\end{equation}
Let $\mathcal I_{c,e}$ collect the layer-head-token triples whose write selects expert $e$ in chunk $c$. Its aggregate gradient and update are
\begin{equation}
\begin{aligned}
G_{c,e}&=\left.\nabla_{W_e}\!\sum_{(\ell,h,i)\in\mathcal I_{c,e}}
 g^w_{\ell,h,i,e}\eta_{\ell,h,i}\ell_{\rm in}(W_e;\mathbf k_{\ell,h,i},\mathbf v_{\ell,h,i})\right|_{W_e=W_e^{(c-1)}},\\
W_e^{(c)}&=\operatorname{Update}\bigl(W_e^{(c-1)},G_{c,e}\bigr).
\end{aligned}
\label{eq:uttt-update}
\end{equation}
The inner derivative holds keys, values, gates and predicted step sizes fixed; outer backpropagation retains their producer graph. As in \eqref{eq:prelim-inner-loop}, $\eta$ absorbs fixed reductions and $\operatorname{Update}$ carries optimizer buffers (Appendix~\ref{app:fw-update}). Read and write selections can differ. At $p=L$ this is \textbf{TTT-MoE}; at $p=1$ it is \textbf{uTTT-MoE}. With expert size $u$, total state stays $Eu$ and active state per token head stays $Ku$, while each layer's candidate pool grows from $E/L$ to $E$ experts. TTT-MoE controls for routing; load balancing acts on expert utilization (Appendix~\ref{app:balancing-mechanisms}).

\subsection{uTTT-Dense}
\label{sec:method:dense}
\label{sec:method:ttt-dense}
A dense group holds one state with $H$ head blocks, each applied in full without expert selection. In the notation of \eqref{eq:topk}, the route is a fixed singleton, $\mathcal{R}_{\ell,h}\equiv\{e(\ell,h)\}$ with unit gate, where $e(\ell,h)$ is the group's block for head $h$. Each block reads its query using $f_W$ and receives writes from the corresponding heads of the group's layers. At $p=L$ this is \textbf{TTT-Dense}, with $L$ layer-private states of size $S$ ($S/H$ active per token head); at $p=1$ it is \textbf{uTTT-Dense}.

Two controls separate sharing from widening. Sharing a state of size $S$ preserves per-token fast-weight compute while reducing total state from $LS$ to $S$. Widening the shared state to $LS$ instead preserves total state and increases active state per head from $S/H$ to $LS/H$, multiplying fast-weight compute by $L$. For intermediate $p$, these paths store $pS$ or $LS$ respectively. Main language comparisons match total state ($Eu=LS$); view synthesis tests both controls. Table~\ref{tab:design-space-full} gives the full matching conditions.

\subsection{Unifying Time and Depth}
\label{sec:method:schedule}
\label{sec:method:recurrent}
\begin{figure}[tb]
    \centering
    \includegraphics[width=\textwidth]{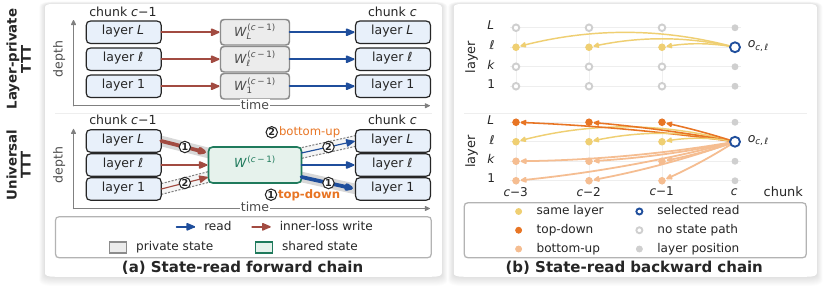}
    \caption{\textbf{State-read forward and backward chains.} Layer-private TTT (top) and universal TTT (bottom). \emph{(a)}~Forward reads use the chunk-start state. Inner-loop writes prepare the state for later reads during both training and inference (Section~\ref{sec:preliminary}). Bands trace top-down (1) and bottom-up (2) routes. \emph{(b)}~The restricted read-output Jacobian traces backward through historical inner updates to write inputs; it is a factor of outer task-loss backpropagation. Each update includes the inner-loss gradient computation. Queries and predicted coefficient producers are held fixed here; full training retains those additional paths. Hollow nodes lack this restricted state-chain path; writers below layer $\ell$ in earlier chunks can still reach the read through the residual stream. Colors identify the forward writer-to-reader relation; arrows point backward. Routing determines which paths are active (Appendix~\ref{app:gr:proof}).}
    \label{fig:recurrence}
\end{figure}

Sharing fast-weight state extends the two learning loops in Section~\ref{sec:preliminary} across depth. In the \emph{inner loop}, layers jointly update their group's state at write-enabled chunk boundaries. A deep layer's write in chunk $c$ can then support a shallower layer's read in chunk $c+1$, a \emph{top-down route} absent from private state; the converse, a shallow layer's write read by a deeper layer, is a \emph{bottom-up route} (Figure~\ref{fig:recurrence}a). In the \emph{outer loop}, task-loss gradients from later reads propagate through these inner updates to train earlier writes at other depths. Figure~\ref{fig:recurrence}b shows the restricted state-chain Jacobian used by this backward pass; full training also retains query and hidden-state producer paths.

Writing $\mathcal S_c$ for the complete recurrent state (fast weights, master accumulators, optimizer buffers and attention caches), the aggregated model obeys $\mathcal S_c=f_\theta(\mathcal S_{c-1},C_c)$. Each transition runs all layers on chunk $C_c$ using the chunk-start fast weights and combines each group's writes at the boundary, \eqref{eq:aggregated}. Our runtime therefore advances chunk by chunk; this execution order differs from the two optimization loops above (Appendix~\ref{app:fw-schedules}). A direct routed path requires writer and reader to select a common expert. Proposition~\ref{prop:reach} (Appendix~\ref{app:gr:proof}) states the reach: in a universal pool, a read's gradient reaches the writes of every earlier write-enabled chunk at any depth, a routed pool requiring a common expert, whereas with private memories it never reaches a deeper layer's writes. The \emph{cross-layer gradient cut} keeps this forward pass and, in the backward pass, stops each layer's read gradient through the other layers' write contributions (Appendix~\ref{app:gr:cut}); it tests the training role of these routes.

\section{Experiments: Language Modeling}
\label{sec:exp-llm}

Long-context language modeling tests how well a fixed fast-weight state retains information beyond the local attention window.

\subsection{Setup}

\paragraph{Data, chunks, and metrics.}
Following the large-chunk TTT language-modeling protocol~\citep{zhang2025testtimetrainingright}, we train autoregressive models at 32K context on Long-Data-Collections~\citep{long_data_collection} at 124M and 760M parameters. The token stream is split into causal chunks of 4{,}096 tokens, with sliding-window attention over 4{,}096 tokens for local context, so a token reads only fast weights written by earlier chunks whichever schedule writes them (Section~\ref{sec:method:schedule}). We report two quantities. Per-token loss (PTL) is the mean next-token loss over positions at or beyond 30K in 32K-token sequences of Books3 from The Pile~\citep{gao2020pile}, following the same protocol and the per-position convention of the Forgetting Transformer~\citep{linforgetting}; RULER retrieval accuracy~\citep{hsieh2024ruler} is averaged over the 4K, 8K, 16K, and 32K settings. The two are not redundant: a model can lower average loss without improving retrieval. Appendix~\ref{app:lm-seeds} reports six training seeds of uTTT-MoE at 124M.

\paragraph{Models.}
The four fast-weight models in the main comparison have the \emph{same total fast-weight state} (Table~\ref{tab:design-space-full}). TTT-Dense implements the layer-private dense design of LaCT, also reported from its published code. TTT-MoE routes each token head into a layer-local pool of four experts ($E/L = 4$, $K = 1$); uTTT-MoE shares the pool across depth, with $E = 48$ at 124M and $E = 96$ at 760M, keeping $K = 1$ and a router per layer. Both routed models use loss-free balancing and the aggregated schedule; uTTT-MoE aggregates utilization statistics across the pool and backpropagates the outer task loss through state reads and differentiable inner-loss writes. All four share the tokenizer, chunk size, sliding-window attention, depth, width, optimizer and token budget. The seven baselines are full-attention Transformer, Transformer-SWA, DeltaNet-SWA~\citep{yang2024parallelizing}, Gated DeltaNet-SWA~\citep{yang2025gateddeltanet}, KDA-SWA~\citep{team2025kimilinear}, Mamba-2-SWA~\citep{dao2024transformersssms} and MoM-SWA~\citep{du2025mom}; all SWA models use the same 4{,}096-token window. Appendix~\ref{app:configs} gives the configurations, and Appendix~\ref{app:uttt-dense-llm} evaluates the dense shared branch at 124M, where it retrieves as well as TTT-Dense with one sixth to one half of the state.

\begin{figure}[t]
    \centering
    \includegraphics[width=\linewidth]{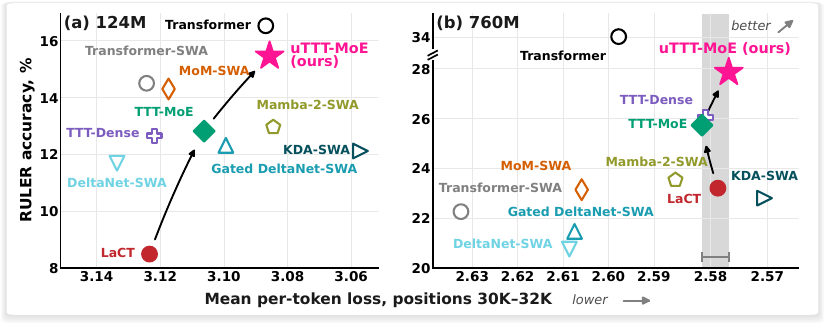}
    \caption{\textbf{Loss--retrieval Pareto plot}, at (a)~124M and (b)~760M parameters. One point per model (Section~\ref{sec:exp-llm}); the horizontal axis is mean per-token loss beyond 30K in 32K-token sequences, and the vertical axis is mean RULER accuracy over 4K--32K. Arrows run through the fast-weight models from LaCT to uTTT-MoE. The shaded band in (b) spans the four fast-weight models, whose losses differ by $0.0047$ nat. The vertical-axis break in (b) marks a compressed empty stretch.}
    \label{fig:teaser-pareto}
\end{figure}

\subsection{Results}

uTTT-MoE leads bounded-state RULER at both scales while matching or beating full attention's per-token loss (Table~\ref{tab:lm-results}, Figure~\ref{fig:teaser-pareto}). At 124M, KDA-SWA and Mamba-2-SWA attain lower loss but retrieve $3.3$ and $2.5$ points less; at 760M only KDA-SWA does, retrieving $5.0$ points less. Appendix~\ref{app:lm-curves} gives retrieval-task and per-position loss curves.

\begin{table}[tb]
    \centering
    \scriptsize
    \caption{\textbf{Long-context language modeling at 32K.} PTL is the mean next-token loss at positions $\geq 30$K on Books3, and the five columns beside it are RULER accuracy (\%): the mean over 4K, 8K, 16K and 32K, then each of those lengths. Bold marks the best value in each column among the bounded-state models; the full-attention Transformer ($\dagger$) keeps a key--value cache that grows with context and is excluded from that comparison. The lower block holds the fast-weight models, of which uTTT-MoE is the model this paper proposes.}
    \label{tab:lm-results}
    \setlength{\tabcolsep}{3.5pt}\renewcommand{\ttVgapwidth}{17pt}%
    \setlength{\cmidrulekern}{\tabcolsep}
    \renewcommand{\arraystretch}{1.15}
    \begin{tabular}{l<{\hspace{10.67pt}}ccccccVcccccc}
        \toprule
        & \multicolumn{6}{cV}{\textbf{124M}} & \multicolumn{6}{@{}c}{\textbf{760M}} \\
        \cmidrule(l{\tabcolsep}r{\dimexpr2\tabcolsep+\ttVgapwidth\relax}){2-7} \cmidrule(l{0pt}r{\tabcolsep}){8-13}
        Model & PTL $\downarrow$ & RULER $\uparrow$ & 4K & 8K & 16K & 32K & PTL $\downarrow$ & RULER $\uparrow$ & 4K & 8K & 16K & 32K \\
        \midrule
        \tgroup{base}{13}{Baselines}
        \trow{base} Transformer $^\dagger$ & 3.0869 & 16.52 & 23.82 & 20.24 & 13.10 & 8.93 & 2.5978 & 34.16 & 41.70 & 37.95 & 33.36 & 23.62 \\
        \trow{base} Transformer-SWA & 3.1243 & 14.50 & \textbf{26.93} & 14.73 & 9.94 & 6.41 & 2.6325 & 22.27 & 45.23 & 22.64 & 12.67 & 8.57 \\
        \trow{base} Gated DeltaNet-SWA & 3.0994 & 12.32 & 23.78 & 12.93 & 8.30 & 4.26 & 2.6075 & 21.48 & 40.40 & 23.12 & 13.66 & 8.76 \\
        \trow{base} DeltaNet-SWA & 3.1336 & 11.69 & 22.52 & 11.23 & 7.63 & 5.36 & 2.6086 & 20.77 & 43.43 & 20.56 & 11.79 & 7.28 \\
        \trow{base} KDA-SWA & \textbf{3.0571} & 12.12 & 20.86 & 12.98 & 8.36 & 6.28 & \textbf{2.5705} & 22.81 & 43.47 & 24.17 & 14.65 & 8.96 \\
        \trow{base} Mamba-2-SWA & 3.0845 & 12.96 & 24.21 & 13.93 & 8.32 & 5.37 & 2.5862 & 23.54 & 48.42 & 25.49 & 12.60 & 7.65 \\
        \trow{base} MoM-SWA & 3.1174 & 14.30 & 26.11 & 14.10 & 10.41 & 6.58 & 2.6059 & 23.15 & 43.35 & 23.69 & 15.97 & 9.60 \\
        \tsep
        \tgroup{fw}{13}{Fast-weight family}
        \trow{fw} LaCT (published) & 3.1234 & 8.50 & 13.75 & 11.37 & 5.93 & 2.94 & 2.5787 & 23.21 & 43.65 & 23.97 & 14.29 & \textbf{10.92} \\
        \trow{fw} TTT-Dense & 3.1219 & 12.65 & 26.31 & 13.58 & 6.00 & 4.72 & 2.5809 & 26.05 & 47.73 & 30.07 & 16.49 & 9.92 \\
        \trow{fw} TTT-MoE & 3.1062 & 12.82 & 25.35 & 13.26 & 7.60 & 5.08 & 2.5815 & 25.73 & 45.27 & 29.44 & \textbf{17.85} & 10.35 \\
        \thl{ours} \textbf{uTTT-MoE} (ours) & 3.0857 & \textbf{15.46} & 26.50 & \textbf{17.32} & \textbf{11.23} & \textbf{6.78} & 2.5768 & \textbf{27.85} & \textbf{54.15} & \textbf{31.00} & 16.03 & 10.23 \\
        \bottomrule
    \end{tabular}
\end{table}

Sharing adds $2.6$ and $2.1$ RULER points over TTT-MoE at 124M and 760M, with the same state and no loss penalty; TTT-MoE stays near TTT-Dense. uTTT-MoE leads TTT-MoE at every length at 124M; at 760M it leads at 4K and 8K and trails by less than two points at 16K and 32K (Table~\ref{tab:lm-results}). Its loss advantage over Transformer-SWA persists to 128K, but retrieval past 32K falls to the window's level for every bounded-state model (Appendix~\ref{app:extrapolation}).

\label{sec:llm:balancing}
Load balancing keeps the pool in use: without it the busiest expert receives $5.9\times$ its even share of the token heads at 124M and $9.6\times$ at 760M, against $1.4\times$ and $1.8\times$ for the reported uTTT-MoE (Appendix~\ref{app:balancing-mechanisms}). At 124M removing it costs uTTT-MoE $4.6$ RULER points and TTT-MoE $2.8$, both to below TTT-Dense. At 760M the unbalanced run has the higher mean, $28.46$ against $27.85$, but the lead comes from one easy needle task (S-NIAH-1), it loses $8$ to $9$ points on five others, and it reads mostly from $22$ of the $96$ experts. We balance the pool rather than each layer: the pool is what is shared, and requiring every layer to use every expert equally would remove the depth allocation that sharing is for (Appendix~\ref{app:balancing-mechanisms}). We use the loss-free bias for both routed models; the auxiliary-loss results are in Appendix~\ref{app:balancing-tasks}.

\begin{figure}[tb]
    \centering
    \includegraphics[width=\textwidth]{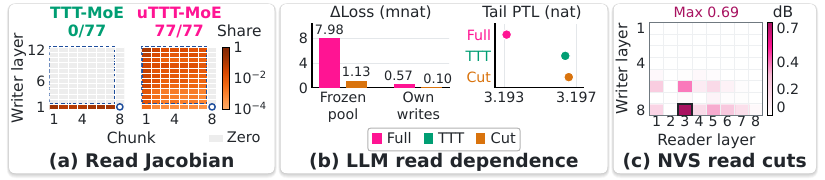}
    \caption{\textbf{State-read backward and forward dependence.}
LLM diagnostics use 124M models and individual Books3 documents' first 32K tokens.
\emph{(a)}~Squared-gradient shares from a layer-1, chunk-8 read-output Jacobian probe, normalized per document and averaged over 300 documents. This probes a local factor of outer task-loss backpropagation. Colors are logarithmic; gray is zero. Boxed counts require nonzero gradients in every document; circles mark the read.
\emph{(b)}~On 1{,}024 documents, bars show loss increases after freezing state writes or reading only own writes, relative to each model's normal predictions over chunks~2--8; dots show unmodified tail PTL at positions $\geq30{,}000$, measured on these unpacked documents and therefore not on the scale of Table~\ref{tab:lm-results}. Full uses full uTTT-MoE backpropagation; Cut stops cross-layer read-to-write gradients during training; TTT is TTT-MoE.
\emph{(c)}~PSNR drop after removing one writer--reader route in uTTT-MoE-e64-p1, averaged over 1{,}019 GSO objects with 23 target views each. Diagonal routes are included. Colors are linear, with drops below $0.01$~dB white; the outline marks the maximum. Panels~(b,c) use fixed trained checkpoints.}
    \label{fig:core-evidence}
\end{figure}

The read-output probe's local Jacobian reaches all 77 earlier deep-layer write locations in uTTT-MoE and none in TTT-MoE across 300 Books3 documents (Figure~\ref{fig:core-evidence}a). This probes a local factor used by outer task-loss backpropagation. On 1{,}024 documents, freezing state writes raises loss by $7.98$~mnat in Full and $1.13$~mnat in Cut, indicating weaker read dependence on written state in Cut (Figure~\ref{fig:core-evidence}b). Appendix~\ref{app:gradient-routes} gives the probe, training and intervention protocols.

\subsection{Measured cost}
\label{sec:llm:cost}

\begin{figure}[tb]
    \centering
    \includegraphics[width=\textwidth]{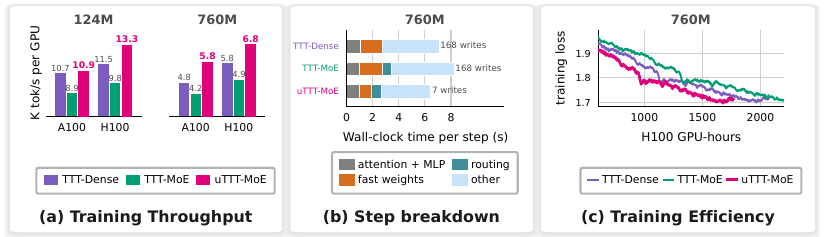}
    \caption{\textbf{Measured training efficiency.} \emph{(a)}~Training throughput of the three ownership regimes on one node of eight A100 or H100 GPUs (32K context, batch 1 per GPU), in thousands of tokens per second per GPU. \emph{(b)}~Where a 760M training step's time goes (one A100, batch 1); attention and MLP are identical across the three models, and \emph{other} is the backward pass and the optimizer. \emph{(c)}~Training loss of the three 760M models against H100 GPU-hours.}
    \label{fig:efficiency-main}
\end{figure}

uTTT-MoE trains fastest in every setting of Figure~\ref{fig:efficiency-main}a and reaches TTT-Dense's final training loss in $1{,}540$ H100 GPU-hours, $15\%$ fewer than TTT-Dense and $27\%$ fewer than TTT-MoE (Figure~\ref{fig:efficiency-main}c). Both routed models use the same fused kernels (Table~\ref{tab:eff-kernels}). uTTT-MoE aggregates cross-layer contributions into one update of the shared expert pool per non-final chunk boundary, while TTT-MoE updates its layer-private expert pools separately. This permits fewer, larger kernel launches; the observed speedup and step-time breakdown (Figure~\ref{fig:efficiency-main}b) are consistent with reduced launch overhead. Appendix~\ref{app:efficiency} reports kernel ablations, sweeps over the number of experts, inference and memory measurements.

\begin{figure}[tb]
    \centering
    \includegraphics[width=\textwidth]{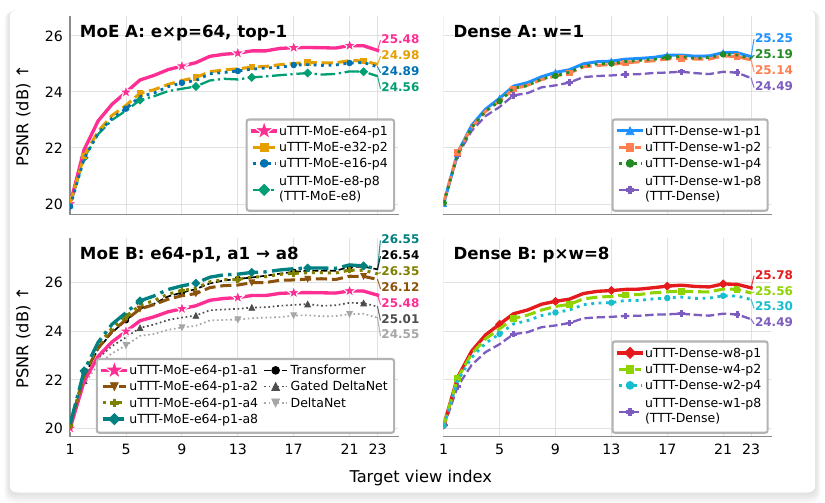}
    \caption{\textbf{Sharing (row A) and compute scaling (row B) on Objaverse/GSO}: PSNR against the target-view index, i.e.\ the number of posed views written into the pool; the number at the right end of each curve is its value at view~23. The $e$, $p$, $w$, and $a$ in model names are defined in Section~\ref{sec:nvs:setup}. Row~A holds per-layer fast-weight compute fixed ($a{=}1$, $w{=}1$) and widens sharing from $p{=}8$ to $p{=}1$; row~B holds total fast-weight state fixed (64 experts, width 8) and raises per-layer compute from one unit to eight. The MoE panel of row~B also shows the three baselines at the same protocol: the full-attention reference (black dashed) and the linear-attention Gated DeltaNet and DeltaNet (grey dotted). Full metrics and 95\% bootstrap intervals are in Table~\ref{tab:nvs-gso-full}.}
    \label{fig:nvs-psnr-by-view}
\end{figure}

\section{Experiments: Novel View Synthesis}
\label{sec:exp-nvs}

Novel view synthesis tests a bounded scene memory: attention stays within each view, so information from other views must pass through the fast-weight state. Training uses eleven writes and evaluation uses 23, testing beyond the trained horizon. We separate sharing at fixed per-layer compute from compute scaling at fixed total state.

\subsection{Setup}
\label{sec:nvs:setup}

\paragraph{Task and protocol.}
Object models train on Objaverse~\citep{deitke2022objaverse} and evaluate on Google Scanned Objects~\citep{downs2022googlescannedobjects} at $256\times256$; scene models use DL3DV~\citep{ling2024dl3dv}, with about 10K training and 140 test scenes. Following LVSM~\citep{jin2025lvsm}, a posed image patchified with ray encodings forms a source chunk that reads and writes the state; a target camera encoded by rays alone forms a read-only chunk that renders the held-out view. Reads and writes use \eqref{eq:uttt-read}--\eqref{eq:uttt-update}. Training views and their order are sampled randomly.

\paragraph{Models.}
The $L{=}8$ layers form $p$ pools ($p{=}8$ private, $p{=}1$ universal). Each routed pool holds $e=E/p$ experts, with $a$ active per token head (the $K$ of Section~\ref{sec:method:uttt-moe}); dense pools have width $w$. An active expert and a width unit cost the same per-layer compute, but a width unit stores eight times as much state: 64 experts match width 8. Names read uTTT-MoE-e$e$-p$p$-a$a$ ($a{=}1$ omitted) and uTTT-Dense-w$w$-p$p$; e8-p8 and w1-p8 are TTT-MoE-e8 and TTT-Dense. Routed models use aggregated writes. Full attention accumulates keys and values across views with other settings matched. Table~\ref{tab:nvs-config} gives the recipe shared by objects and scenes except for the dataset.

\subsection{From private pools to one shared pool}
\label{sec:nvs:ownership}

Row~A of Figure~\ref{fig:nvs-psnr-by-view} merges eight private pools into four, two and one at fixed per-layer compute. Routed models retain 64 experts at fixed parameters, state and active compute; dense models retain width 1 per pool, reducing total state eightfold. Full sharing performs best in both families, gaining $0.92$~dB for routed models and $0.76$~dB for dense models at view~23, with disjoint intervals for the all-view means (Table~\ref{tab:nvs-main}). The curves separate as views accumulate and the gains persist beyond the trained horizon; the dense gain needs neither routing nor more state.

A shared pool of only 8 experts also beats the private pools by $0.99$~dB at view~23, with disjoint intervals for the all-view means, despite storing one eighth of the state and offering the same eight candidates per token head. This isolates shared writes from a wider routing choice. Appendix~\ref{app:nvs-perview} repeats the sharing progression at 32 experts; Appendix~\ref{app:redundancy} examines reuse across views and layers.

Single-route knockouts reveal the strongest dependence on top-down state-read routes: removing writes from layer 8 to reads at layer 3 reduces PSNR by $0.69$~dB, compared with at most $0.06$~dB for same-layer routes (Figure~\ref{fig:core-evidence}c). Appendix~\ref{app:tp:knockout} gives the intervention protocol.

\begin{table}[tb]
    \centering
    \scriptsize
    \caption{\textbf{The configurations of Figure~\ref{fig:nvs-psnr-by-view} on Objaverse/GSO} ($n{=}1{,}019$ objects). Cells: mean over objects of each object's mean over target views 1--23, with its 95\% bootstrap interval over objects (Appendix~\ref{app:nvs-perview}); the second PSNR column is view~23 alone. Bold: best per column among fast-weight models; $^\dagger$full-attention reference, not bolded. The tinted row is the best configuration in the table.}
    \label{tab:nvs-main}
    \setlength{\tabcolsep}{3.0pt}
    \renewcommand{\arraystretch}{1.12}
    \begin{tabularx}{\textwidth}{l*{4}{>{\centering\arraybackslash}X}}
        \toprule
        Model & PSNR $\uparrow$ & View 23 PSNR $\uparrow$ & SSIM $\uparrow$ & LPIPS $\downarrow$ \\
        \midrule
        \tgroup{moeA}{5}{MoE A: $e{\times}p{=}64$, top-1}
        \trow{moeA} uTTT-MoE-e64-p1 & 24.68 {\tiny[24.51, 24.86]} & 25.48 {\tiny[25.27, 25.68]} & 0.860 {\tiny[0.855, 0.864]} & 0.173 {\tiny[0.168, 0.177]} \\
        \trow{moeA} uTTT-MoE-e32-p2 & 24.20 {\tiny[24.03, 24.37]} & 24.98 {\tiny[24.78, 25.18]} & 0.854 {\tiny[0.850, 0.859]} & 0.181 {\tiny[0.177, 0.186]} \\
        \trow{moeA} uTTT-MoE-e16-p4 & 24.11 {\tiny[23.94, 24.29]} & 24.89 {\tiny[24.69, 25.10]} & 0.854 {\tiny[0.849, 0.858]} & 0.182 {\tiny[0.177, 0.186]} \\
        \trow{moeA} uTTT-MoE-e8-p8 (TTT-MoE-e8) & 23.90 {\tiny[23.73, 24.08]} & 24.56 {\tiny[24.36, 24.77]} & 0.850 {\tiny[0.845, 0.855]} & 0.186 {\tiny[0.181, 0.191]} \\
        \tsep
        \tgroup{denA}{5}{Dense A: $w{=}1$}
        \trow{denA} uTTT-Dense-w1-p1 & 24.45 {\tiny[24.27, 24.62]} & 25.25 {\tiny[25.05, 25.46]} & 0.858 {\tiny[0.853, 0.862]} & 0.176 {\tiny[0.172, 0.181]} \\
        \trow{denA} uTTT-Dense-w1-p2 & 24.35 {\tiny[24.17, 24.53]} & 25.14 {\tiny[24.93, 25.34]} & 0.856 {\tiny[0.852, 0.861]} & 0.179 {\tiny[0.175, 0.184]} \\
        \trow{denA} uTTT-Dense-w1-p4 & 24.37 {\tiny[24.20, 24.55]} & 25.19 {\tiny[24.98, 25.40]} & 0.858 {\tiny[0.853, 0.862]} & 0.177 {\tiny[0.172, 0.182]} \\
        \trow{denA} uTTT-Dense-w1-p8 (TTT-Dense) & 23.99 {\tiny[23.81, 24.16]} & 24.49 {\tiny[24.29, 24.69]} & 0.852 {\tiny[0.847, 0.856]} & 0.185 {\tiny[0.180, 0.189]} \\
        \tsep
        \tgroup{moeB}{5}{MoE B: e64-p1, $a{=}2{\to}8$ (a1 above)}
        \trow{moeB} uTTT-MoE-e64-p1-a2 & 25.18 {\tiny[25.00, 25.36]} & 26.12 {\tiny[25.91, 26.33]} & 0.867 {\tiny[0.862, 0.871]} & 0.164 {\tiny[0.159, 0.168]} \\
        \trow{moeB} uTTT-MoE-e64-p1-a4 & 25.38 {\tiny[25.20, 25.56]} & 26.35 {\tiny[26.14, 26.56]} & 0.869 {\tiny[0.864, 0.873]} & 0.161 {\tiny[0.156, 0.165]} \\
        \thl{moeB} \textbf{uTTT-MoE-e64-p1-a8} & \textbf{25.56} {\tiny[25.38, 25.74]} & \textbf{26.55} {\tiny[26.34, 26.76]} & \textbf{0.871} {\tiny[0.867, 0.876]} & \textbf{0.158} {\tiny[0.154, 0.163]} \\
        \tsep
        \tgroup{denB}{5}{Dense B: $p{\times}w{=}8$ (w1-p8 above)}
        \trow{denB} uTTT-Dense-w8-p1 & 24.95 {\tiny[24.77, 25.13]} & 25.78 {\tiny[25.57, 25.99]} & 0.864 {\tiny[0.859, 0.868]} & 0.167 {\tiny[0.163, 0.172]} \\
        \trow{denB} uTTT-Dense-w4-p2 & 24.75 {\tiny[24.58, 24.93]} & 25.56 {\tiny[25.36, 25.77]} & 0.861 {\tiny[0.857, 0.866]} & 0.172 {\tiny[0.167, 0.176]} \\
        \trow{denB} uTTT-Dense-w2-p4 & 24.53 {\tiny[24.35, 24.71]} & 25.30 {\tiny[25.09, 25.51]} & 0.859 {\tiny[0.854, 0.863]} & 0.174 {\tiny[0.169, 0.179]} \\
        \tsep
        \tgroup{ref}{5}{Baseline}
        \trow{ref} Transformer$^\dagger$ & 25.36 {\tiny[25.19, 25.54]} & 26.54 {\tiny[26.33, 26.74]} & 0.870 {\tiny[0.865, 0.874]} & 0.162 {\tiny[0.158, 0.166]} \\
        \trow{ref} Gated DeltaNet & 24.32 {\tiny[24.14, 24.50]} & 25.01 {\tiny[24.80, 25.21]} & 0.856 {\tiny[0.851, 0.860]} & 0.178 {\tiny[0.173, 0.182]} \\
        \trow{ref} DeltaNet & 23.94 {\tiny[23.76, 24.12]} & 24.55 {\tiny[24.34, 24.75]} & 0.851 {\tiny[0.846, 0.856]} & 0.185 {\tiny[0.180, 0.189]} \\
        \bottomrule
    \end{tabularx}
\end{table}

\subsection{More compute at fixed state}
\label{sec:nvs:capacity}

Row~B fixes total state at 64 experts or width 8 and matches per-layer compute: routed models activate $a{=}1,2,4,8$ experts; dense models widen to $w{=}1,2,4,8$ as pools merge ($p{=}8,4,2,1$), changing sharing as well. Routed view-23 PSNR improves monotonically by $1.07$~dB. The dense gain is $1.29$~dB: $0.76$ from sharing and $0.53$ from widening, consistent with the width sweep at full sharing (Appendix~\ref{app:nvs-perview}). Routed models lead their compute-matched dense partners at every level, by $0.77$~dB at eight units, with disjoint all-view intervals. At $a{=}8$, the fixed state matches full attention at view~23 ($26.55$ versus $26.54$~dB) and leads at earlier views (Table~\ref{tab:nvs-main}). All three metrics improve; LPIPS falls from $0.185$ for TTT-Dense to $0.158$ for uTTT-MoE at $a{=}8$.

\FloatBarrier

Appendix~\ref{app:nvs-perview} confirms monotone activation scaling at 32 experts ($25.30\to26.39$). Growing the pool from 8 to 64 experts at $a{=}1$ changes view-23 PSNR by at most $0.25$~dB without a consistent trend. On DL3DV, increasing $a{=}1$ to $a{=}8$ raises all-view PSNR from $16.12$ to $16.66$.

\section{Conclusion}

Fast-weight memory need not be owned by individual layers. uTTT shares it across depth, so the memory evolves along time and depth together, and layer-private TTT becomes one partition of a larger design space. At equal or smaller state, uTTT improves both language modeling and view synthesis, its routed form trains faster, and trained models depend on the cross-layer routes that sharing opens. Depth then adds to test-time optimization instead of multiplying separate memories.

\bibliographystyle{iclr2027_conference}
\bibliography{references}

\begin{thebibliography}{68}
\providecommand{\natexlab}[1]{#1}
\providecommand{\url}[1]{\texttt{#1}}
\expandafter\ifx\csname urlstyle\endcsname\relax
  \providecommand{\doi}[1]{doi: #1}\else
  \providecommand{\doi}{doi: \begingroup \urlstyle{rm}\Url}\fi

\bibitem[Ba et~al.(2016)Ba, Hinton, Mnih, Leibo, and
  Ionescu]{ba2016fastweights}
Jimmy Ba, Geoffrey~E Hinton, Volodymyr Mnih, Joel~Z Leibo, and Catalin Ionescu.
\newblock Using fast weights to attend to the recent past.
\newblock In \emph{Advances in Neural Information Processing Systems}, 2016.

\bibitem[Behrouz et~al.(2025)Behrouz, Zhong, and Mirrokni]{behrouz2024titans}
Ali Behrouz, Peilin Zhong, and Vahab Mirrokni.
\newblock Titans: Learning to memorize at test time.
\newblock In \emph{Advances in Neural Information Processing Systems}, 2025.

\bibitem[Beltagy et~al.(2020)Beltagy, Peters, and Cohan]{beltagy2020longformer}
Iz~Beltagy, Matthew~E. Peters, and Arman Cohan.
\newblock Longformer: The long-document transformer.
\newblock \emph{arXiv preprint arXiv:2004.05150}, 2020.

\bibitem[Berges et~al.(2025)Berges, O{\u{g}}uz, Haziza, Yih, Zettlemoyer, and
  Ghosh]{berges2025memorylayers}
Vincent-Pierre Berges, Barlas O{\u{g}}uz, Daniel Haziza, Wen-tau Yih, Luke
  Zettlemoyer, and Gargi Ghosh.
\newblock Memory layers at scale.
\newblock In \emph{International Conference on Machine Learning}, 2025.

\bibitem[Brandon et~al.(2024)Brandon, Mishra, Nrusimha, Panda, and
  Ragan-Kelley]{brandon2024cla}
William Brandon, Mayank Mishra, Aniruddha Nrusimha, Rameswar Panda, and
  Jonathan Ragan-Kelley.
\newblock Reducing transformer key-value cache size with cross-layer attention.
\newblock In \emph{Advances in Neural Information Processing Systems}, 2024.

\bibitem[Chen et~al.(2026)Chen, Gu, Shang, Zhang, Feng, Sheng, Liu, Wang, Sun,
  Wu, and Wang]{chen2026moue}
Yilong Chen, Naibin Gu, Junyuan Shang, Zhenyu Zhang, Yuchen Feng, Jiawei Sheng,
  Tingwen Liu, Shuohuan Wang, Yu~Sun, Hua Wu, and Haifeng Wang.
\newblock Mixture of universal experts: Scaling virtual width via depth-width
  transformation.
\newblock \emph{arXiv preprint arXiv:2603.04971}, 2026.

\bibitem[Clark et~al.(2022)Clark, Guu, Chang, Pasupat, Hinton, and
  Norouzi]{clark2022fastweightlayers}
Kevin Clark, Kelvin Guu, Ming-Wei Chang, Panupong Pasupat, Geoffrey Hinton, and
  Mohammad Norouzi.
\newblock Meta-learning fast weight language models.
\newblock In \emph{Conference on Empirical Methods in Natural Language
  Processing}, 2022.

\bibitem[Csord\'{a}s et~al.(2024)Csord\'{a}s, Irie, Schmidhuber, Potts, and
  Manning]{csordas2024moeut}
R\'{o}bert Csord\'{a}s, Kazuki Irie, J\"{u}rgen Schmidhuber, Christopher Potts,
  and Christopher~D. Manning.
\newblock {MoEUT}: Mixture-of-experts universal transformers.
\newblock In \emph{Advances in Neural Information Processing Systems}, 2024.

\bibitem[Dao \& Gu(2024)Dao and Gu]{dao2024transformersssms}
Tri Dao and Albert Gu.
\newblock Transformers are {SSM}s: Generalized models and efficient algorithms
  through structured state space duality.
\newblock In \emph{International Conference on Machine Learning}, 2024.

\bibitem[Dehghani et~al.(2019)Dehghani, Gouws, Vinyals, Uszkoreit, and
  Kaiser]{dehghani2019universal}
Mostafa Dehghani, Stephan Gouws, Oriol Vinyals, Jakob Uszkoreit, and {\L}ukasz
  Kaiser.
\newblock Universal transformers.
\newblock In \emph{International Conference on Learning Representations}, 2019.

\bibitem[Deitke et~al.(2023)Deitke, Schwenk, Salvador, Weihs, Michel,
  VanderBilt, Schmidt, Ehsani, Kembhavi, and Farhadi]{deitke2022objaverse}
Matt Deitke, Dustin Schwenk, Jordi Salvador, Luca Weihs, Oscar Michel, Eli
  VanderBilt, Ludwig Schmidt, Kiana Ehsani, Aniruddha Kembhavi, and Ali
  Farhadi.
\newblock Objaverse: A universe of annotated {3D} objects.
\newblock In \emph{Proceedings of the IEEE/CVF Conference on Computer Vision
  and Pattern Recognition}, 2023.

\bibitem[Downs et~al.(2022)Downs, Francis, Koenig, Kinman, Hickman, Reymann,
  McHugh, and Vanhoucke]{downs2022googlescannedobjects}
Laura Downs, Anthony Francis, Nate Koenig, Brandon Kinman, Ryan Hickman, Krista
  Reymann, Thomas~B McHugh, and Vincent Vanhoucke.
\newblock {Google Scanned Objects}: A high-quality dataset of {3D} scanned
  household items.
\newblock In \emph{IEEE International Conference on Robotics and Automation},
  2022.

\bibitem[Du et~al.(2026)Du, Sun, Lan, Hu, Zhang, and Cheng]{du2025mom}
Jusen Du, Weigao Sun, Disen Lan, Jiaxi Hu, Tao Zhang, and Yu~Cheng.
\newblock {MoM}: Linear sequence modeling with mixture-of-memories.
\newblock In \emph{International Conference on Learning Representations}, 2026.

\bibitem[Fan et~al.(2020)Fan, Lavril, Grave, Joulin, and
  Sukhbaatar]{fan2020feedback}
Angela Fan, Thibaut Lavril, Edouard Grave, Armand Joulin, and Sainbayar
  Sukhbaatar.
\newblock Addressing some limitations of transformers with feedback memory.
\newblock \emph{arXiv preprint arXiv:2002.09402}, 2020.

\bibitem[Fedus et~al.(2022)Fedus, Zoph, and Shazeer]{fedus2022switch}
William Fedus, Barret Zoph, and Noam Shazeer.
\newblock {Switch Transformers}: Scaling to trillion parameter models with
  simple and efficient sparsity.
\newblock \emph{Journal of Machine Learning Research}, 23\penalty0
  (120):\penalty0 1--39, 2022.

\bibitem[Finn et~al.(2017)Finn, Abbeel, and Levine]{finn2017maml}
Chelsea Finn, Pieter Abbeel, and Sergey Levine.
\newblock Model-agnostic meta-learning for fast adaptation of deep networks.
\newblock In \emph{International Conference on Machine Learning}, 2017.

\bibitem[Gao et~al.(2020)Gao, Biderman, Black, Golding, Hoppe, Foster, Phang,
  He, Thite, Nabeshima, Presser, and Leahy]{gao2020pile}
Leo Gao, Stella Biderman, Sid Black, Laurence Golding, Travis Hoppe, Charles
  Foster, Jason Phang, Horace He, Anish Thite, Noa Nabeshima, Shawn Presser,
  and Connor Leahy.
\newblock {The Pile}: An {800GB} dataset of diverse text for language modeling.
\newblock \emph{arXiv preprint arXiv:2101.00027}, 2020.

\bibitem[Gao et~al.(2024)Gao, Tow, Abbasi, Biderman, Black, DiPofi, Foster,
  Golding, Hsu, Le~Noac'h, Li, McDonell, Muennighoff, Ociepa, Phang, Reynolds,
  Schoelkopf, Skowron, Sutawika, Tang, Thite, Wang, Wang, and
  Zou]{eval-harness}
Leo Gao, Jonathan Tow, Baber Abbasi, Stella Biderman, Sid Black, Anthony
  DiPofi, Charles Foster, Laurence Golding, Jeffrey Hsu, Alain Le~Noac'h,
  Haonan Li, Kyle McDonell, Niklas Muennighoff, Chris Ociepa, Jason Phang,
  Laria Reynolds, Hailey Schoelkopf, Aviya Skowron, Lintang Sutawika, Eric
  Tang, Anish Thite, Ben Wang, Kevin Wang, and Andy Zou.
\newblock The language model evaluation harness.
\newblock Zenodo, 2024.
\newblock URL \url{https://doi.org/10.5281/zenodo.5371628}.

\bibitem[Gu \& Dao(2024)Gu and Dao]{gu2024mamba}
Albert Gu and Tri Dao.
\newblock Mamba: Linear-time sequence modeling with selective state spaces.
\newblock In \emph{Conference on Language Modeling}, 2024.

\bibitem[Gu et~al.(2026)Gu, Likhomanenko, Thilak, Ramapuram, and
  Jaitly]{gu2026pathmoe}
Zijin Gu, Tatiana Likhomanenko, Vimal Thilak, Jason Ramapuram, and Navdeep
  Jaitly.
\newblock Path-constrained mixture-of-experts.
\newblock \emph{arXiv preprint arXiv:2603.18297}, 2026.

\bibitem[Hinton \& Plaut(1987)Hinton and Plaut]{hinton1987fastweights}
Geoffrey~E. Hinton and David~C. Plaut.
\newblock Using fast weights to deblur old memories.
\newblock In \emph{Annual Conference of the Cognitive Science Society}, 1987.

\bibitem[Hochreiter \& Schmidhuber(1997)Hochreiter and
  Schmidhuber]{hochreiter1997long}
Sepp Hochreiter and J{\"u}rgen Schmidhuber.
\newblock Long short-term memory.
\newblock \emph{Neural Computation}, 9\penalty0 (8):\penalty0 1735--1780, 1997.

\bibitem[Hong et~al.(2024)Hong, Zhang, Gu, Bi, Zhou, Liu, Liu, Sunkavalli, Bui,
  and Tan]{hong2024lrm}
Yicong Hong, Kai Zhang, Jiuxiang Gu, Sai Bi, Yang Zhou, Difan Liu, Feng Liu,
  Kalyan Sunkavalli, Trung Bui, and Hao Tan.
\newblock {LRM}: Large reconstruction model for single image to {3D}.
\newblock In \emph{International Conference on Learning Representations}, 2024.

\bibitem[Hsieh et~al.(2024)Hsieh, Sun, Kriman, Acharya, Rekesh, Jia, Zhang, and
  Ginsburg]{hsieh2024ruler}
Cheng-Ping Hsieh, Simeng Sun, Samuel Kriman, Shantanu Acharya, Dima Rekesh, Fei
  Jia, Yang Zhang, and Boris Ginsburg.
\newblock {RULER}: What's the real context size of your long-context language
  models?
\newblock In \emph{Conference on Language Modeling}, 2024.

\bibitem[Huang et~al.(2026)Huang, Shi, Zheng, Wu, Chen, Yu, Yin, and
  Cheng]{huang2026unipool}
Minbin Huang, Han Shi, Chuanyang Zheng, Yimeng Wu, Guoxuan Chen, Xintong Yu,
  Yichun Yin, and Hong Cheng.
\newblock {UniPool}: A globally shared expert pool for mixture-of-experts.
\newblock \emph{arXiv preprint arXiv:2605.06665}, 2026.

\bibitem[Hutchins et~al.(2022)Hutchins, Schlag, Wu, Dyer, and
  Neyshabur]{hutchins2022block}
DeLesley Hutchins, Imanol Schlag, Yuhuai Wu, Ethan Dyer, and Behnam Neyshabur.
\newblock Block-recurrent transformers.
\newblock In \emph{Advances in Neural Information Processing Systems}, 2022.

\bibitem[Jaggi(2026)]{jaggi2026tying}
Martin Jaggi.
\newblock Tying the loop -- tied expert layers in mixture-of-experts language
  models.
\newblock \emph{arXiv preprint arXiv:2606.16825}, 2026.

\bibitem[Jin et~al.(2025)Jin, Jiang, Tan, Zhang, Bi, Zhang, Luan, Snavely, and
  Xu]{jin2025lvsm}
Haian Jin, Hanwen Jiang, Hao Tan, Kai Zhang, Sai Bi, Tianyuan Zhang, Fujun
  Luan, Noah Snavely, and Zexiang Xu.
\newblock {LVSM}: A large view synthesis model with minimal {3D} inductive
  bias.
\newblock In \emph{International Conference on Learning Representations}, 2025.

\bibitem[Jordan et~al.(2024)Jordan, Jin, Boza, You, Cesista, Newhouse, and
  Bernstein]{jordan2024muon}
Keller Jordan, Yuchen Jin, Vlado Boza, Jiacheng You, Franz Cesista, Laker
  Newhouse, and Jeremy Bernstein.
\newblock {Muon}: An optimizer for hidden layers in neural networks.
\newblock Blog post, 2024.
\newblock URL \url{https://kellerjordan.github.io/posts/muon/}.

\bibitem[Ju et~al.(2022)Ju, Roller, Sukhbaatar, and Weston]{ju2021staircase}
Da~Ju, Stephen Roller, Sainbayar Sukhbaatar, and Jason Weston.
\newblock Staircase attention for recurrent processing of sequences.
\newblock In \emph{Advances in Neural Information Processing Systems}, 2022.

\bibitem[Katharopoulos et~al.(2020)Katharopoulos, Vyas, Pappas, and
  Fleuret]{katharopoulos2020transformers}
Angelos Katharopoulos, Apoorv Vyas, Nikolaos Pappas, and Fran{\c{c}}ois
  Fleuret.
\newblock Transformers are {RNNs}: Fast autoregressive transformers with linear
  attention.
\newblock In \emph{International Conference on Machine Learning}, 2020.

\bibitem[Kerbl et~al.(2023)Kerbl, Kopanas, Leimk{\"u}hler, and
  Drettakis]{kerbl20233dgs}
Bernhard Kerbl, Georgios Kopanas, Thomas Leimk{\"u}hler, and George Drettakis.
\newblock {3D} {Gaussian} splatting for real-time radiance field rendering.
\newblock \emph{ACM Transactions on Graphics}, 42\penalty0 (4):\penalty0 1--14,
  2023.

\bibitem[{Kimi Team}(2025)]{team2025kimilinear}
{Kimi Team}.
\newblock {Kimi Linear}: An expressive, efficient attention architecture.
\newblock \emph{arXiv preprint arXiv:2510.26692}, 2025.

\bibitem[Krause et~al.(2018)Krause, Kahembwe, Murray, and
  Renals]{krause2018dynamic}
Ben Krause, Emmanuel Kahembwe, Iain Murray, and Steve Renals.
\newblock Dynamic evaluation of neural sequence models.
\newblock In \emph{International Conference on Machine Learning}, 2018.

\bibitem[Lample et~al.(2019)Lample, Sablayrolles, Ranzato, Denoyer, and
  J{\'e}gou]{lample2019productkeys}
Guillaume Lample, Alexandre Sablayrolles, Marc'Aurelio Ranzato, Ludovic
  Denoyer, and Herv{\'e} J{\'e}gou.
\newblock Large memory layers with product keys.
\newblock In \emph{Advances in Neural Information Processing Systems}, 2019.

\bibitem[Lan et~al.(2020)Lan, Chen, Goodman, Gimpel, Sharma, and
  Soricut]{lan2020albert}
Zhenzhong Lan, Mingda Chen, Sebastian Goodman, Kevin Gimpel, Piyush Sharma, and
  Radu Soricut.
\newblock {ALBERT}: A lite {BERT} for self-supervised learning of language
  representations.
\newblock In \emph{International Conference on Learning Representations}, 2020.

\bibitem[Lin et~al.(2025)Lin, Nikishin, He, and Courville]{linforgetting}
Zhixuan Lin, Evgenii Nikishin, Xu~He, and Aaron Courville.
\newblock Forgetting transformer: Softmax attention with a forget gate.
\newblock In \emph{International Conference on Learning Representations}, 2025.

\bibitem[Ling et~al.(2024)Ling, Sheng, Tu, Zhao, Xin, Wan, Yu, Guo, Yu, Lu, Li,
  Sun, Ashok, Mukherjee, Kang, Kong, Hua, Zhang, Benes, and
  Bera]{ling2024dl3dv}
Lu~Ling, Yichen Sheng, Zhi Tu, Wentian Zhao, Cheng Xin, Kun Wan, Lantao Yu,
  Qianyu Guo, Zixun Yu, Yawen Lu, Xuanmao Li, Xingpeng Sun, Rohan Ashok,
  Aniruddha Mukherjee, Hao Kang, Xiangrui Kong, Gang Hua, Tianyi Zhang, Bedrich
  Benes, and Aniket Bera.
\newblock {DL3DV-10K}: A large-scale scene dataset for deep learning-based {3D}
  vision.
\newblock In \emph{Proceedings of the IEEE/CVF Conference on Computer Vision
  and Pattern Recognition}, 2024.

\bibitem[Loshchilov \& Hutter(2019)Loshchilov and Hutter]{loshchilov2019adamw}
Ilya Loshchilov and Frank Hutter.
\newblock Decoupled weight decay regularization.
\newblock In \emph{International Conference on Learning Representations}, 2019.

\bibitem[Mikolov et~al.(2010)Mikolov, Karafi{\'a}t, Burget, {\v{C}}ernock{\'y},
  and Khudanpur]{mikolov2010recurrent}
Tom{\'a}{\v{s}} Mikolov, Martin Karafi{\'a}t, Luk{\'a}{\v{s}} Burget, Jan
  {\v{C}}ernock{\'y}, and Sanjeev Khudanpur.
\newblock Recurrent neural network based language model.
\newblock In \emph{Interspeech}, 2010.

\bibitem[Mildenhall et~al.(2020)Mildenhall, Srinivasan, Tancik, Barron,
  Ramamoorthi, and Ng]{mildenhall2020nerf}
Ben Mildenhall, Pratul~P Srinivasan, Matthew Tancik, Jonathan~T Barron, Ravi
  Ramamoorthi, and Ren Ng.
\newblock {NeRF}: Representing scenes as neural radiance fields for view
  synthesis.
\newblock In \emph{European Conference on Computer Vision}, 2020.

\bibitem[Rajpurkar et~al.(2018)Rajpurkar, Jia, and Liang]{rajpurkar2018know}
Pranav Rajpurkar, Robin Jia, and Percy Liang.
\newblock Know what you don't know: Unanswerable questions for {SQuAD}.
\newblock In \emph{Annual Meeting of the Association for Computational
  Linguistics}, 2018.

\bibitem[Sajjadi et~al.(2022)Sajjadi, Meyer, Pot, Bergmann, Greff, Radwan,
  Vora, Lu\v{c}i\'c, Duckworth, Dosovitskiy, Uszkoreit, Funkhouser, and
  Tagliasacchi]{sajjadi2022scene}
Mehdi S.~M. Sajjadi, Henning Meyer, Etienne Pot, Urs Bergmann, Klaus Greff,
  Noha Radwan, Suhani Vora, Mario Lu\v{c}i\'c, Daniel Duckworth, Alexey
  Dosovitskiy, Jakob Uszkoreit, Thomas Funkhouser, and Andrea Tagliasacchi.
\newblock Scene representation transformer: Geometry-free novel view synthesis
  through set-latent scene representations.
\newblock In \emph{Proceedings of the IEEE/CVF Conference on Computer Vision
  and Pattern Recognition}, 2022.

\bibitem[Schlag et~al.(2021)Schlag, Irie, and Schmidhuber]{schlag2021linear}
Imanol Schlag, Kazuki Irie, and J{\"u}rgen Schmidhuber.
\newblock Linear transformers are secretly fast weight programmers.
\newblock In \emph{International Conference on Machine Learning}, 2021.

\bibitem[Schmidhuber(1992)]{schmidhuber1992learning}
J{\"u}rgen Schmidhuber.
\newblock Learning to control fast-weight memories: An alternative to dynamic
  recurrent networks.
\newblock \emph{Neural Computation}, 4\penalty0 (1):\penalty0 131--139, 1992.

\bibitem[Shazeer(2020)]{shazeer2020glu}
Noam Shazeer.
\newblock {GLU} variants improve {Transformer}.
\newblock \emph{arXiv preprint arXiv:2002.05202}, 2020.

\bibitem[Shazeer et~al.(2017)Shazeer, Mirhoseini, Maziarz, Davis, Le, Hinton,
  and Dean]{shazeer2017outrageously}
Noam Shazeer, Azalia Mirhoseini, Krzysztof Maziarz, Andy Davis, Quoc Le,
  Geoffrey Hinton, and Jeff Dean.
\newblock Outrageously large neural networks: The sparsely-gated
  mixture-of-experts layer.
\newblock In \emph{International Conference on Learning Representations}, 2017.

\bibitem[Su et~al.(2024)Su, Ahmed, Lu, Pan, Bo, and Liu]{su2024roformer}
Jianlin Su, Murtadha Ahmed, Yu~Lu, Shengfeng Pan, Wen Bo, and Yunfeng Liu.
\newblock {RoFormer}: Enhanced transformer with rotary position embedding.
\newblock \emph{Neurocomputing}, 568:\penalty0 127063, 2024.

\bibitem[Sun et~al.(2020)Sun, Wang, Liu, Miller, Efros, and Hardt]{sun2020ttt}
Yu~Sun, Xiaolong Wang, Zhuang Liu, John Miller, Alexei~A. Efros, and Moritz
  Hardt.
\newblock Test-time training with self-supervision for generalization under
  distribution shifts.
\newblock In \emph{International Conference on Machine Learning}, 2020.

\bibitem[Sun et~al.(2023)Sun, Li, Dalal, Hsu, Koyejo, Guestrin, Wang,
  Hashimoto, and Chen]{sun2023learningtolearn}
Yu~Sun, Xinhao Li, Karan Dalal, Chloe Hsu, Sanmi Koyejo, Carlos Guestrin,
  Xiaolong Wang, Tatsunori Hashimoto, and Xinlei Chen.
\newblock Learning to ({L}earn at test time).
\newblock \emph{arXiv preprint arXiv:2310.13807}, 2023.

\bibitem[Sun et~al.(2025)Sun, Li, Dalal, Xu, Vikram, Zhang, Dubois, Chen, Wang,
  Koyejo, Hashimoto, and Guestrin]{sun2025learninglearntesttime}
Yu~Sun, Xinhao Li, Karan Dalal, Jiarui Xu, Arjun Vikram, Genghan Zhang, Yann
  Dubois, Xinlei Chen, Xiaolong Wang, Sanmi Koyejo, Tatsunori Hashimoto, and
  Carlos Guestrin.
\newblock Learning to ({L}earn at test time): {RNN}s with expressive hidden
  states.
\newblock In \emph{International Conference on Machine Learning}, 2025.

\bibitem[Sun et~al.(2024)Sun, Dong, Zhu, Huang, Wang, Ma, Zhang, Wang, and
  Wei]{sun2024yoco}
Yutao Sun, Li~Dong, Yi~Zhu, Shaohan Huang, Wenhui Wang, Shuming Ma, Quanlu
  Zhang, Jianyong Wang, and Furu Wei.
\newblock You only cache once: Decoder-decoder architectures for language
  models.
\newblock In \emph{Advances in Neural Information Processing Systems}, 2024.

\bibitem[Tan et~al.(2023)Tan, Shen, Chen, Courville, and Gan]{tan2023sut}
Shawn Tan, Yikang Shen, Zhenfang Chen, Aaron Courville, and Chuang Gan.
\newblock Sparse universal transformer.
\newblock In \emph{Conference on Empirical Methods in Natural Language
  Processing}, 2023.

\bibitem[Tan et~al.(2025)Tan, Li, Yuan, Zhou, Liu, Zhuang, Li, Niu, Qin, Yao,
  Liu, Xu, Li, Dai, Zhao, and Wang]{tan2025rexmoe}
Zheyue Tan, Zhiyuan Li, Tao Yuan, Dong Zhou, Weilin Liu, Yueqing Zhuang, Yadong
  Li, Guowei Niu, Cheng Qin, Zhuyu Yao, Congyi Liu, Haiyang Xu, Boxun Li,
  Guohao Dai, Bo~Zhao, and Yu~Wang.
\newblock {ReXMoE}: Reusing experts with minimal overhead in
  mixture-of-experts.
\newblock \emph{arXiv preprint arXiv:2510.17483}, 2025.

\bibitem[Tandon et~al.(2025)Tandon, Dalal, Li, Koceja, R{\o}d, Buchanan, Wang,
  Leskovec, Koyejo, Hashimoto, Guestrin, McCaleb, Choi, and
  Sun]{tandon2025endtoendtesttimetraininglong}
Arnuv Tandon, Karan Dalal, Xinhao Li, Daniel Koceja, Marcel R{\o}d, Sam
  Buchanan, Xiaolong Wang, Jure Leskovec, Sanmi Koyejo, Tatsunori Hashimoto,
  Carlos Guestrin, Jed McCaleb, Yejin Choi, and Yu~Sun.
\newblock End-to-end test-time training for long context.
\newblock \emph{arXiv preprint arXiv:2512.23675}, 2025.

\bibitem[{Together AI}(2023)]{long_data_collection}
{Together AI}.
\newblock {Long-Data-Collections}.
\newblock Hugging Face dataset, no longer publicly accessible; data mixture
  documented in the LLaMA-2-7B-32K model card, 2023.
\newblock URL
  \url{https://huggingface.co/togethercomputer/LLaMA-2-7B-32K#training-and-fine-tuning}.

\bibitem[Vaswani et~al.(2017)Vaswani, Shazeer, Parmar, Uszkoreit, Jones, Gomez,
  Kaiser, and Polosukhin]{vaswani2017attention}
Ashish Vaswani, Noam Shazeer, Niki Parmar, Jakob Uszkoreit, Llion Jones,
  Aidan~N Gomez, {\L}ukasz Kaiser, and Illia Polosukhin.
\newblock Attention is all you need.
\newblock In \emph{Advances in Neural Information Processing Systems}, 2017.

\bibitem[von Oswald et~al.(2026)von Oswald, Scherrer, Kobayashi, Versari, Yang,
  Schlegel, Maile, Schimpf, Sieberling, Meulemans, Lajoie, Saurous, Frenkel,
  Pascanu, {Ag{\"u}era y Arcas}, and Sacramento]{vonoswald2025mesanet}
Johannes von Oswald, Nino Scherrer, Seijin Kobayashi, Luca Versari, Songlin
  Yang, Maximilian Schlegel, Kaitlin Maile, Yanick Schimpf, Oliver Sieberling,
  Alexander Meulemans, Guillaume Lajoie, Rif~A. Saurous, Charlotte Frenkel,
  Razvan Pascanu, Blaise {Ag{\"u}era y Arcas}, and Jo{\~a}o Sacramento.
\newblock {MesaNet}: Sequence modeling by locally optimal test-time training.
\newblock In \emph{International Conference on Learning Representations}, 2026.

\bibitem[Wang et~al.(2026)Wang, Shi, and Fox]{wang2025testtimeregression}
Ke~Alexander Wang, Jiaxin Shi, and Emily~B. Fox.
\newblock Test-time regression: a unifying framework for designing sequence
  models with associative memory.
\newblock \emph{Journal of Machine Learning Research}, 27\penalty0
  (169):\penalty0 1--41, 2026.

\bibitem[Wang et~al.(2024)Wang, Gao, Zhao, Sun, and Dai]{wang2024auxlossfree}
Lean Wang, Huazuo Gao, Chenggang Zhao, Xu~Sun, and Damai Dai.
\newblock Auxiliary-loss-free load balancing strategy for mixture-of-experts.
\newblock \emph{arXiv preprint arXiv:2408.15664}, 2024.

\bibitem[Wang et~al.(2025)Wang, Zhang, Holynski, Efros, and
  Kanazawa]{wang2025continuous}
Qianqian Wang, Yifei Zhang, Aleksander Holynski, Alexei~A. Efros, and Angjoo
  Kanazawa.
\newblock Continuous {3D} perception model with persistent state.
\newblock In \emph{Proceedings of the IEEE/CVF Conference on Computer Vision
  and Pattern Recognition}, 2025.

\bibitem[Wu \& Tu(2024)Wu and Tu]{wu2024layercondensed}
Haoyi Wu and Kewei Tu.
\newblock Layer-condensed {KV} cache for efficient inference of large language
  models.
\newblock In \emph{Annual Meeting of the Association for Computational
  Linguistics}, 2024.

\bibitem[Yang et~al.(2024)Yang, Wang, Zhang, Shen, and
  Kim]{yang2024parallelizing}
Songlin Yang, Bailin Wang, Yu~Zhang, Yikang Shen, and Yoon Kim.
\newblock Parallelizing linear transformers with the delta rule over sequence
  length.
\newblock In \emph{Advances in Neural Information Processing Systems}, 2024.

\bibitem[Yang et~al.(2025)Yang, Kautz, and Hatamizadeh]{yang2025gateddeltanet}
Songlin Yang, Jan Kautz, and Ali Hatamizadeh.
\newblock Gated delta networks: Improving {Mamba2} with delta rule.
\newblock In \emph{International Conference on Learning Representations}, 2025.

\bibitem[Yang et~al.(2018)Yang, Qi, Zhang, Bengio, Cohen, Salakhutdinov, and
  Manning]{yang2018hotpotqa}
Zhilin Yang, Peng Qi, Saizheng Zhang, Yoshua Bengio, William~W. Cohen, Ruslan
  Salakhutdinov, and Christopher~D. Manning.
\newblock {H}otpot{QA}: A dataset for diverse, explainable multi-hop question
  answering.
\newblock In \emph{Conference on Empirical Methods in Natural Language
  Processing}, 2018.

\bibitem[Yu et~al.(2021)Yu, Ye, Tancik, and Kanazawa]{yu2021pixelnerf}
Alex Yu, Vickie Ye, Matthew Tancik, and Angjoo Kanazawa.
\newblock {pixelNeRF}: Neural radiance fields from one or few images.
\newblock In \emph{Proceedings of the IEEE/CVF Conference on Computer Vision
  and Pattern Recognition}, 2021.

\bibitem[Zhang \& Sennrich(2019)Zhang and Sennrich]{zhang2019rmsnorm}
Biao Zhang and Rico Sennrich.
\newblock Root mean square layer normalization.
\newblock In \emph{Advances in Neural Information Processing Systems}, 2019.

\bibitem[Zhang et~al.(2026)Zhang, Bi, Hong, Zhang, Luan, Yang, Sunkavalli,
  Freeman, and Tan]{zhang2025testtimetrainingright}
Tianyuan Zhang, Sai Bi, Yicong Hong, Kai Zhang, Fujun Luan, Songlin Yang,
  Kalyan Sunkavalli, William~T Freeman, and Hao Tan.
\newblock Test-time training done right.
\newblock In \emph{International Conference on Learning Representations}, 2026.

\end{thebibliography}

\clearpage
\appendix
\makeatletter
\@addtoreset{figure}{section}\@addtoreset{table}{section}
\makeatother
\renewcommand{\thefigure}{\thesection\arabic{figure}}
\renewcommand{\thetable}{\thesection\arabic{table}}
\renewcommand{\theHfigure}{\thesection.\arabic{figure}}
\renewcommand{\theHtable}{\thesection.\arabic{table}}
\addtocontents{toc}{\protect\setcounter{tocdepth}{2}}
\renewcommand{\contentsname}{Appendix Contents}
\tableofcontents
\clearpage
\section{Extended Related Work}
\label{app:related}

\subsection{Fast weights and test-time training}
\label{sec:related:ttt}
Fast weights update rapidly on recent data while slow weights change rarely~\citep{hinton1987fastweights,schmidhuber1992learning,ba2016fastweights}; Fast Weight Layers express dynamic-evaluation updates~\citep{mikolov2010recurrent,krause2018dynamic} as linear attention trained end to end~\citep{clark2022fastweightlayers}. Besides the long-memory use of Section~\ref{sec:related}, test-time training can improve generalization on novel instances by optimizing a self-supervised objective at test time~\citep{sun2020ttt}. Our TTT layers are meta-learned: the chunk writes form the inner loop, and ordinary outer-loop training learns the initial fast weights, as in MAML~\citep{finn2017maml}, together with the step-size and, where used, momentum-coefficient predictors.

\subsection{Sharing parameters and experts across depth}
\label{sec:related:universal}
Sparse mixture-of-experts layers route each token to one or a few experts~\citep{shazeer2017outrageously,fedus2022switch}. Universal Transformers~\citep{dehghani2019universal} and ALBERT~\citep{lan2020albert} tie a block's parameters across layers; SUT and MoEUT tie the whole layer, experts and router included, per depth step~\citep{tan2023sut} or per repeated group of layers~\citep{csordas2024moeut}. ReXMoE shares experts across adjacent layers~\citep{tan2025rexmoe}; other designs tie the expert sub-block within groups of layers~\citep{jaggi2026tying} or share router parameters within blocks of consecutive layers while keeping per-layer experts~\citep{gu2026pathmoe}. UniPool replaces layer-private expert sets with one shared pool, addressed by per-layer routers and balanced at the pool level~\citep{huang2026unipool}. MoUE keeps layer-local experts beside a pool of universal experts, of which each group of consecutive layers reaches a window that shifts with depth, counting the expert paths this opens across depth as virtual width; its router also scores experts against a state matrix, which MoUE calls fast weights over depth, shared by layers with the same reachable experts and updated after routing, without backpropagation, from the current batch's keys and detached routing targets~\citep{chen2026moue}.

Product-key memory and Memory Layers use sparse lookup into key--value parameters learned during ordinary training~\citep{lample2019productkeys,berges2025memorylayers}, and Memory Layers shares these parameters across its memory layers. Apart from MoUE's routing state, everything the designs above route to, look up or share stays fixed at inference; uTTT instead shares the fast-weight memory itself, which all layers write within each sequence and whose updates are retained in the training graph (Section~\ref{sec:method:recurrent}).

\subsection{Novel view synthesis}
\label{sec:related:nvs}
Scene representations are fitted per scene~\citep{mildenhall2020nerf,kerbl20233dgs}, which is itself test-time optimization, or inferred from one or a few images by a prior learned across scenes~\citep{yu2021pixelnerf,hong2024lrm}. SRT and encoder--decoder LVSM replace the 3D representation with a set of latent tokens that a decoder queries with target rays~\citep{sajjadi2022scene,jin2025lvsm}; decoder-only LVSM, whose tokenization Section~\ref{sec:exp-nvs} adopts, drops even that, running input-view and target-ray tokens through one transformer. LaCT applies large-chunk, layer-private fast weights to novel view synthesis~\citep{zhang2025testtimetrainingright}, and the novel-view block of Section~\ref{sec:exp-nvs} follows its layer design (Appendix~\ref{app:configs}). A geometry transformer can also fold unposed images one by one into a persistent state~\citep{wang2025continuous}.

\section{Datasets and Licenses}
\label{app:assets}

Table~\ref{tab:assets-terms} lists the external datasets used in the experiments with their license or access terms.

\begin{table}[!t]
    \centering
    \small
    \caption{External datasets, with license or access terms.}
    \label{tab:assets-terms}
    \renewcommand{\arraystretch}{1.12}
    \begin{tabularx}{\textwidth}{@{}>{\raggedright\arraybackslash}p{0.31\textwidth}>{\raggedright\arraybackslash}p{0.27\textwidth}>{\raggedright\arraybackslash}X@{}}
        \toprule
        Asset & Use in this paper & License or access \\
        \midrule
        Objaverse~\citep{deitke2022objaverse} & NVS training & ODC-By v1.0; per-object CC licenses, some non-commercial \\
        Google Scanned Objects~\citep{downs2022googlescannedobjects} & NVS evaluation & CC BY 4.0 \\
        DL3DV-10K~\citep{ling2024dl3dv} & NVS training and evaluation & CC BY-NC 4.0; gated access \\
        Long-Data-Collections~\citep{long_data_collection} & LM training & Mixed sources, each under its own license \\
        Books3 from The Pile~\citep{gao2020pile} & LM evaluation (Section~\ref{sec:exp-llm}; Appendices~\ref{app:lm-results}, \ref{app:extrapolation} and~\ref{app:gradient-routes}) & Copyrighted books; withdrawn from its original host in 2023 \\
        RULER~\citep{hsieh2024ruler} & Retrieval evaluation, via the LM Evaluation Harness~\citep{eval-harness} & Apache-2.0 (harness: MIT); some needle tasks use Paul Graham essays as filler, the QA tasks use the two datasets below, the rest are synthetic \\
        SQuAD~2.0~\citep{rajpurkar2018know} & RULER single-hop QA (development set) & CC BY-SA 4.0 \\
        HotpotQA~\citep{yang2018hotpotqa} & RULER multi-hop QA (development set, distractor setting) & CC BY-SA 4.0 \\
        \bottomrule
    \end{tabularx}
\end{table}
\FloatBarrier

\section{Fast-Weight Update and Write Schedules}
\label{app:preliminaries}

\subsection{Read network and write update}
\label{app:fw-update}
Expert blocks use $d_{\rm fw}=d_h$, while dense widening increases $d_{\rm fw}$. For matrix $m\in\{0,1,2\}$ of $W_e$, the aggregated write direction is
\begin{equation}
D_{c,e}^{(m)}
=\rho\!\sum_{(\ell,h,i)\in\mathcal I_{c,e}}
g^w_{\ell,h,i,e}\,\eta_{\ell,h,i}^{(m)}
\nabla_{W_e^{(m)}}\bigl[v_{\ell,h,i}^{\top}f_{W_e}(k_{\ell,h,i})\bigr]
\Big|_{W_e=W_e^{(c-1)}}.
\label{eq:fw-write-direction}
\end{equation}
It has the opposite sign to the loss gradient supplied to $\operatorname{Update}$ in \eqref{eq:uttt-update}, which suppresses the matrix index on $\eta$ and absorbs the reduction factor $\rho$ into it. LaCT, TTT-Dense and both routed language models use $\rho=1$, as does routed NVS. Language-modeling uTTT-Dense averages across layers, giving $\rho=1/L$; dense NVS averages over chunk tokens, giving $\rho=1/|C_c|$, and sums across sharing layers.

\paragraph{Predicted step sizes and momentum.}
The input $\bar x_{\ell,i}$ to the coefficient predictors is the normalized layer input supplied to attention in the main language-modeling configurations. In NVS, it is obtained by applying an additional RMS normalization without learned affine parameters ($\epsilon=10^{-5}$) to the same layer's normalized memory input. Step sizes take the form
\begin{equation}
\eta_{\ell,h,i}^{(m)}
=\operatorname{softplus}\!\left(\psi_{\ell,h}^{(m)}(\bar x_{\ell,i})
+\operatorname{softplus}^{-1}(\eta_0)\right),
\qquad
\eta_0=10^{-3}\ \text{(LM)},\quad 10^{-2}\ \text{(NVS)}.
\label{eq:fw-step-predictor}
\end{equation}
The $\psi$ are learned affine projections, bias-free in NVS. In the routed language models, a layer predicts a scalar momentum coefficient for each token and averages over the chunk. A shared pool further averages over the layers $\mathcal G$ of its group:
\begin{equation}
\mu_{c,\ell}=\frac{1}{|C_c|}\sum_{i\in C_c}
\sigma\!\left(\pi_\ell(\bar x_{\ell,i})\right),
\qquad
\mu_c=\frac{1}{|\mathcal G|}\sum_{\ell\in\mathcal G}\mu_{c,\ell}.
\label{eq:fw-momentum-predictor}
\end{equation}
Here $\sigma$ is the sigmoid and $\pi_\ell$ a learned linear projection. This mean is over the pool's layers and tokens, independent of which experts they select; $|\mathcal G|=1$ in a private pool. Dense language models instead predict a separate coefficient per head and average it over tokens.

\paragraph{Routed language-modeling update.}
Let $U$ be the master accumulators, $W$ the fast weights used by both reads and write-gradient evaluations, and $M$ the momentum buffers. Each sequence starts from the learned matrices $U_{\rm init}$, parameterized as in Table~\ref{tab:lm-memory}, with $U_{c=0}=W_{c=0}=U_{\rm init}$ and $M_{c=0}=0$; the initial matrices are not row-normalized. For each expert and each matrix, with these indices suppressed, the boundary update is
\begin{equation}
\begin{aligned}
M_c&=D_c+\mu_c M_{c-1},
& B_c&=U_{c-1}+\mathcal O_5(M_c),\\
U_c&=\mathcal P_{\rm row}(B_c;U_{c-1}),
& W_c&=U_c.
\end{aligned}
\label{eq:fw-routed-update}
\end{equation}
The row rescaling uses the pre-update master accumulators as its reference:
\begin{equation}
\bigl[\mathcal P_{\rm row}(B;R)\bigr]_{j,:}
=\frac{\|R_{j,:}\|_2}{\|B_{j,:}\|_2+10^{-5}}B_{j,:}.
\label{eq:fw-row-rescale}
\end{equation}
The buffer stores $M_c$ before the Muon transform $\mathcal O_5$. The routed configurations apply no further step-size multiplier at the boundary: the token step sizes have already entered $D_c$. An expert with no new write can still change through its previous momentum.

\paragraph{Five-step Muon transform.}
For each update matrix, transpose $M$ if it has more rows than columns, then form $X_0=M/(\|M\|_F+10^{-7})$ and iterate
\begin{equation}
A_t=X_tX_t^\top,\qquad
X_{t+1}=a_tX_t+(b_tA_t+c_tA_t^2)X_t,
\quad t=0,\ldots,4.
\label{eq:fw-muon}
\end{equation}
It uses the five coefficient triples
\begin{equation*}
\begin{array}{rrr}
a_t&b_t&c_t\\
4.0848&-6.8946&2.9270\\
3.9505&-6.3029&2.6377\\
3.7418&-5.5913&2.3037\\
2.8769&-3.1427&1.2046\\
2.8366&-3.0525&1.2012
\end{array}
\end{equation*}
and transposes $X_5$ back when necessary. This is the matrix orthogonalization step used by the LaCT-derived write implementation with Muon~\citep{jordan2024muon}; it is applied separately to the three matrices of each expert or head, with the width-block convention for language-modeling uTTT-Dense described below.

\paragraph{Dense language modeling and NVS.}
TTT-Dense retains an unnormalized master accumulator: after the same momentum and Muon steps, $U_c=U_{c-1}+\mathcal O_5(M_c)$ and $W_c=\mathcal P_{\rm row}(U_c;W_{c-1})$. Thus its rescaling is applied to the fast copy using the previous fast weights' row norms, while \eqref{eq:fw-routed-update} rescales the master itself. LaCT uses the sequence-initial row norms in this rescaling, replacing $W_{c-1}$ with the initial fast weights $W_{\rm init}$ as the reference.
Language-modeling uTTT-Dense uses the unnormalized-master transition of TTT-Dense, with each head's write directions and momentum coefficients averaged across layers. At width multiplier $r$, Muon and row rescaling act independently on its $r$ square width blocks.
The NVS configurations omit both momentum and Muon and use
\begin{equation}
U_c=U_{c-1}+D_c,\qquad
(W_c)_{j,:}=\frac{(U_c)_{j,:}}{\max(\|(U_c)_{j,:}\|_2,10^{-5})}.
\label{eq:fw-nvs-update}
\end{equation}
At each sequence start, $U_{c=0}$ is the learned initial master state and $W_{c=0}$ is its row-normalized copy, using the normalization in~\eqref{eq:fw-nvs-update}. The unnormalized master persists across chunks; the normalized fast copy is used for reads and for evaluating the next write gradient. Only source-view chunks write, while target pose-only chunks read.

\subsection{Write schedules and execution order}
\label{app:fw-schedules}
Under both schedules all layers read the chunk-start state $W^{(c-1)}$; they differ in how the $L$ per-layer contributions $G_{c,\ell}$ are applied at the chunk boundary. The \emph{sequential} schedule applies them as $L$ successive steps, evaluating each gradient at the state left by the preceding update,
\begin{equation}
W^{(c,\ell)} = \operatorname{Update}\big(W^{(c,\ell-1)},\, G_{c,\ell}\big), \qquad W^{(c,0)} = W^{(c-1)}, \quad W^{(c)} = W^{(c,L)}.
\label{eq:sequential}
\end{equation}
The \emph{aggregated} schedule of Section~\ref{sec:method:universal}, \eqref{eq:aggregated}, evaluates all gradients at $W^{(c-1)}$ and applies their sum in one step. With $p$ groups, both schedules act within each group. A group's next chunk waits for all of its layers, so our runtime runs chunk-first (Section~\ref{sec:method:recurrent}); private states also permit LaCT's layer-first order. The aggregated schedule reduces the optimizer updates per write-enabled chunk from $L$ to $p$; routing and raw-gradient computation remain layer-specific. We train with it; Appendix~\ref{app:nvs-schedules} compares the two schedules on view synthesis.

\section{Model Configurations}
\label{app:configs}

This appendix records how every model we train is built, read from the training code and configurations. For each experiment it gives the settings all compared models share, then what distinguishes each model, and, for language modeling, the fast-weight block itself.

\subsection{Language modeling}

Every language model follows one recipe (Table~\ref{tab:lm-config}) and differs only in the token mixer beside the 4K window (Table~\ref{tab:lm-models}); Table~\ref{tab:lm-memory} specifies the fast-weight block, whose read and update equations are given in Appendix~\ref{app:fw-update}.

\paragraph{How the models are aligned.}
Table~\ref{tab:lm-models} groups the models by what they are matched in and cites the work each mixer comes from. Gated DeltaNet-SWA's narrower FFN pays exactly for its output-gate projection, and Mamba-2-SWA's state size $N = 0.75\,d$ is set so that its state equals the fast-weight state, while its FFN is narrowed to $1{,}152 / 2{,}304$ to pay for the window branch's own $q,k,v$ projections, which every other mixer reuses from the window, so it is matched in parameters as well. MoM-SWA routes each token to one of four Gated DeltaNet memories that share one $k,v$ projection; it keeps Gated DeltaNet-SWA's FFN and pays for its router with $36{,}864$ further weights at 124M and $147{,}456$ at 760M, so it matches the other linear-attention rows in parameters while carrying four times their state.

\begin{table}[!htbp]
    \centering
    \small
    \caption{Settings shared by every language model. Where one value spans both columns it holds at both scales.}
    \label{tab:lm-config}
    \begin{tabular*}{\textwidth}{@{\hspace{\tabcolsep}}l@{\extracolsep{\fill}\hspace{2\tabcolsep}}c@{\extracolsep{0pt}\hspace{2\tabcolsep}}c}
        \toprule
        & 124M & 760M \\
        \midrule
        Width $d$ / layers $L$ / attention heads & 768 / 12 / 12 & 1536 / 24 / 24 \\
        Attention head dimension & \multicolumn{2}{c}{64} \\
        FFN (SwiGLU~\citep{shazeer2020glu}) hidden width & 2{,}048 & 4{,}096 \\
        Normalization & \multicolumn{2}{c}{pre-norm RMSNorm~\citep{zhang2019rmsnorm}} \\
        Vocabulary & \multicolumn{2}{c}{32{,}000; untied embeddings} \\
        RoPE~\citep{su2024roformer} $\theta$ & \multicolumn{2}{c}{$10^6$} \\
        Sliding window & \multicolumn{2}{c}{4{,}096 (every model but the Transformer)} \\
        Context length & \multicolumn{2}{c}{32{,}768} \\
        \midrule
        Data & \multicolumn{2}{c}{Long-Data-Collections} \\
        Optimizer & \multicolumn{2}{c}{AdamW~\citep{loshchilov2019adamw}} \\
        $\beta$, $\varepsilon$, weight decay & \multicolumn{2}{c}{$(0.9, 0.95)$, $10^{-15}$, 0.1} \\
        Learning rate & \multicolumn{2}{c}{$10^{-3}$; 1{,}024 warmup steps, cosine decay to $10^{-4}$} \\
        Gradient clip & \multicolumn{2}{c}{1.0} \\
        Global batch & \multicolumn{2}{c}{32 sequences (1{,}048{,}576 tokens)} \\
        Training steps & 10{,}240 & 40{,}960 \\
        Training tokens & 10.7B & 42.9B \\
        \bottomrule
    \end{tabular*}
\end{table}

\begin{table}[!htbp]
    \centering
    \small
    \caption{What distinguishes each language model. Every mixer but full attention runs beside the 4K window of Table~\ref{tab:lm-config} and, except Mamba-2's, reuses its $q,k,v$ projections; the linear-attention mixers apply SiLU and L2 to $q,k$ and use no short convolution. State is what a layer carries across chunks; uTTT-MoE pools it, $4L$ experts shared by all layers. Parameters exclude the input embedding and output head (24.6M / 49.2M each).}
    \label{tab:lm-models}
    \setlength{\tabcolsep}{10.32pt}%
    \begin{tabular}{lccrr}
        \toprule
        & & & \multicolumn{2}{c}{Non-emb.\ params} \\
        \cmidrule(l){4-5}
        Model & FFN width & State / layer & 124M & 760M \\
        \midrule
        \tgroup{att}{5}{Attention references}
        \trow{att} Transformer~\citep{vaswani2017attention} & 2{,}048 / 4{,}096 & KV cache & 84.95M & 679.6M \\
        \trow{att} Transformer-SWA~\citep{beltagy2020longformer} & 2{,}048 / 4{,}096 & --- & 84.95M & 679.6M \\
        \tsep
        \tgroup{lin}{5}{Linear attention, matched in parameters}
        \trow{lin} DeltaNet-SWA~\citep{yang2024parallelizing} & 2{,}048 / 4{,}096 & $64\,d$ & 85.10M & 680.6M \\
        \trow{lin} Gated DeltaNet-SWA~\citep{yang2025gateddeltanet} & 1{,}792 / 3{,}584 & $64\,d$ & 85.21M & 681.5M \\
        \trow{lin} KDA-SWA~\citep{team2025kimilinear} & 1{,}984 / 4{,}032 & $64\,d$ & 85.71M & 683.0M \\
        \trow{lin} MoM-SWA~\citep{du2025mom} & 1{,}792 / 3{,}584 & $4\times 64\,d$ & 85.25M & 681.6M \\
        \tsep
        \tgroup{mamba}{5}{State space, matched in parameters}
        \trow{mamba} Mamba-2-SWA~\citep{dao2024transformersssms} & 1{,}152 / 2{,}304 & $0.75\,d^2$ & 85.23M & 681.1M \\
        \tsep
        \tgroup{fw}{5}{Fast-weight models, matched in state}
        \trow{fw} LaCT~\citep{zhang2025testtimetrainingright} & 2{,}048 / 4{,}096 & $0.75\,d^2$ & 88.13M & 699.3M \\
        \trow{fw} TTT-Dense & 2{,}048 / 4{,}096 & $0.75\,d^2$ & 88.13M & 699.3M \\
        \trow{fw} TTT-MoE & 2{,}048 / 4{,}096 & $0.75\,d^2$ & 88.11M & 699.2M \\
        \thl{ours} uTTT-MoE & 2{,}048 / 4{,}096 & $0.75\,d^2$ & 88.21M & 700.1M \\
        \bottomrule
    \end{tabular}
\end{table}

\begin{table}[!htbp]
    \centering
    \small
    \caption{The fast-weight block ($d$: model width; two values are 124M / 760M). LaCT and TTT-Dense compute the same block in a different order: LaCT runs the whole sequence through one layer at a time, TTT-Dense one chunk at a time through all layers.}
    \label{tab:lm-memory}
    \begin{tabular*}{\textwidth}{@{\extracolsep{\fill}}lccc@{}}
        \toprule
        & LaCT, TTT-Dense & TTT-MoE & uTTT-MoE \\
        \midrule
        Heads $\times$ width & \multicolumn{3}{c}{$4 \times d/4$} \\
        Function & \multicolumn{3}{c}{SwiGLU; initial $W_0, W_2$: rank 32 $+\,0.5\,I$} \\
        $q, k, v$ & \multicolumn{3}{c}{window attention's, pre-RoPE; SiLU; L2, then RoPE on $q,k$} \\
        Step size & \multicolumn{3}{c}{per token, head, and matrix; base $10^{-3}$} \\
        Update & \multicolumn{3}{c}{momentum, Muon~\citep{jordan2024muon} (5 steps), row renormalization} \\
        Momentum coefficient & per head & per chunk & per chunk \\
        Chunk & \multicolumn{3}{c}{4{,}096 tokens; read, then write} \\
        \midrule
        Pool & per layer & 4 experts / layer & $4L$ experts, shared \\
        Router & --- & softmax top-1, per layer & softmax top-1, per layer \\
        Load balancing & --- & loss-free $10^{-5}$ & loss-free, $3{\times}10^{-5}$ / $10^{-3}$ \\
        Write & own pool & own pool & aggregated, once per chunk \\
        \bottomrule
    \end{tabular*}
\end{table}

\paragraph{Other language models.}
uTTT-Dense (Appendix~\ref{app:uttt-dense-llm}) replaces the per-layer pools of TTT-Dense with one dense state shared by all layers, $r$ times as wide as one layer's ($r \in \{1, 2, 4, 6\}$ in Table~\ref{tab:lm-uttt-dense}), with a full-rank learned initial state and an aggregated chunk schedule; each layer keeps its own step-size, momentum, gate, and output modules, and the recipe is that of Table~\ref{tab:lm-config}. Its layer averaging and blockwise update are specified in Appendix~\ref{app:fw-update}. The seven 124M models of Appendix~\ref{app:extrapolation} are separate training runs: five use the configurations of Transformer, Transformer-SWA, LaCT, TTT-MoE, and uTTT-MoE, and the two 64K-trained attention references change only the sequence length and train 16 sequences per step, so each step still sees 1{,}048{,}576 tokens and the budget is unchanged.

\subsection{Novel view synthesis}

Every fast-weight novel-view-synthesis model of Section~\ref{sec:exp-nvs} shares one backbone and recipe (Table~\ref{tab:nvs-config}) and differs only in the fast-weight pool, as defined in Section~\ref{sec:nvs:setup} and Appendix~\ref{app:nvs-perview}. Object- and scene-level fast-weight models differ only in data. Object models are trained for roughly 71B tokens and scene models for roughly 58B --- a global batch of 128 sequences, each of 22 chunks of 1{,}024 tokens (11 posed views and 11 target cameras) --- in a single stage at fixed resolution and view count, on NVIDIA A100 GPUs.

\begin{table}[!htbp]
    \centering
    \small
    \caption{What distinguishes each novel-view model. The fast-weight models share Table~\ref{tab:nvs-config}; the references change only the memory sublayer and the FFN width. Parameters count every trainable weight, including learnable initial states; the fast-weight row is for 64 total experts or total width 8 (36.25M outside the state and router, 0.79M initial state, and a router of at most 4{,}096 weights).}
    \label{tab:nvs-models}
    \begin{tabular}{lccr}
        \toprule
        Model & FFN width & State / layer & Params \\
        \midrule
        \tgroup{ref}{4}{References}
        \trow{ref} Transformer & 1{,}536 & KV cache ($+1{,}024$ tokens per view) & 27.76M \\
        \trow{ref} DeltaNet & 1{,}584 & $8 \times 64^2$, delta rule & 37.04M \\
        \trow{ref} Gated DeltaNet & 1{,}408 & $8 \times 64^2$, gated delta rule & 37.01M \\
        \tsep
        \tgroup{fw}{4}{Fast-weight models of Section~\ref{sec:exp-nvs}}
        \trow{fw} \shortstack[l]{64 total experts or\\total width 8} & 1{,}536 & $8 \times 3 \times 64^2$ per width unit & 37.04--37.05M \\
        \bottomrule
    \end{tabular}
\end{table}

\begin{table}[!htbp]
    \centering
    \small
    \caption{Settings shared by every novel-view-synthesis fast-weight model of Section~\ref{sec:exp-nvs}; 36.25M parameters outside the fast-weight state and router.}
    \label{tab:nvs-config}
    \begin{tabular*}{\textwidth}{@{\extracolsep{\fill}}ll@{}}
        \toprule
        Width / layers / attention heads & 512 / 8 / 8 \\
        Block & attention $\to$ fast-weight read $\to$ SwiGLU FFN (1{,}536) \\
        Attention & within one view \\
        Input & $256^2$, patch 8: 1{,}024 tokens per view \\
        Views & training 12 (11 posed + 11 target chunks); evaluation 24 \\
        Fast-weight heads & $8 \times 64$ \\
        Unit of state & expert: $3 \times 64^2$ weights, shared by heads; width unit: 8 experts \\
        $q, k, v$ & SiLU; L2 on $q,k$ \\
        Step size & per token, head, and matrix; base 0.01 \\
        Update & no momentum or Muon; rows L2-normalized \\
        Router & softmax top-$a$ on $q$ (read) or $k$ (write); balancing loss $10^{-5}$ \\
        \midrule
        Optimizer & AdamW, $\beta = (0.9, 0.95)$, weight decay 0.05 \\
        Learning rate & $4{\times}10^{-4}$, 2{,}500 warmup steps, cosine to 0 \\
        Gradient clip / global batch & 1.0 / 128 \\
        Training steps & 24{,}476 / 20{,}046 (objects / scenes; 20{,}000 rounded up to an epoch) \\
        Loss & pixel MSE \\
        \bottomrule
    \end{tabular*}
\end{table}

\paragraph{The two instantiations.}
Figure~\ref{fig:uttt-main} draws the language-modeling block: the window attention and the fast-weight memory run in parallel on one set of $q,k,v$ projections, and their outputs are summed before one output projection (Table~\ref{tab:lm-memory}). The novel-view block instead has three sequential pre-norm sublayers with a memory that has its own $q,k,v$ projections, the layout of LaCT's view-synthesis model~\citep{zhang2025testtimetrainingright}, and eight fast-weight heads of dimension 64 rather than four of dimension $d/4$, updated without momentum or Muon (Table~\ref{tab:nvs-config}). What the two share is the axis this paper studies, which layers may read and write a pool.

\paragraph{Reference models.}
Three references keep the backbone, block order, recipe and global batch of Table~\ref{tab:nvs-config} and change only the memory sublayer and the FFN width that pays for it (Table~\ref{tab:nvs-models}). The Transformer, an LVSM-style model~\citep{jin2025lvsm}, drops the fast-weight sublayer: each layer attends to the current view and to the cached keys and values of every earlier posed view, so its memory grows by 1{,}024 tokens per view, and target chunks never enter the cache. DeltaNet and Gated DeltaNet replace the fast-weight sublayer by a linear-attention memory with the same $8 \times 64$ heads, SiLU and L2 on $q,k$ and no short convolution, updated once per view chunk by the delta rule, $U \leftarrow e^{\bar g}\,U + \tfrac{1}{n}\sum_{i} \beta_i\,(v_i - W k_i)\,k_i^{\top}$ over the $n$ tokens of the chunk, whose rows are L2-normalized into the read copy $W$ as in \eqref{eq:fw-nvs-update}; $\bar g$ is the chunk mean of the forget gate, Gated DeltaNet also has a gated output, and DeltaNet has neither gate; both start from a learnable state, as LaCT does. Their FFN widths set them to the parameter count of the fast-weight models once the learnable initial states are included. The Transformer is evaluated at step 24{,}000 (objects) and 20{,}000 (scenes), the linear-attention models at their final step (24{,}476 and 20{,}046), all under the protocol of Appendix~\ref{app:nvs-perview}.

\FloatBarrier
\section{Language-Modeling Experiments}
\label{app:lm-results}

\label{app:lm-curves}
\begin{figure}[tb]
    \centering
    \includegraphics[width=\textwidth]{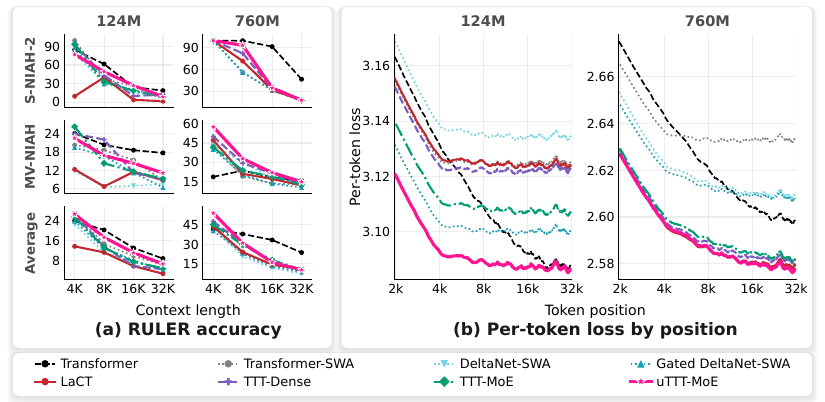}
    \caption{\textbf{Retrieval by task and loss by position}, at 124M and 760M. \emph{(a)}~RULER accuracy at 4K--32K on the second single-needle task (S-NIAH-2), the multi-value task (MV-NIAH), and, in the last row, the unweighted mean over all thirteen RULER tasks, which is the RULER axis of Figure~\ref{fig:teaser-pareto}. \emph{(b)}~Per-token validation loss on Books3 by position within a 32K-token sequence, on a log-scaled axis with positions below 2K omitted as warm-up.}
    \label{fig:ruler-tasks}
\end{figure}

Figure~\ref{fig:ruler-tasks} resolves retrieval by task and validation loss by token position at both model scales.

\subsection{Load Balancing}
\label{app:balancing-mechanisms}
Both balancing mechanisms act on expert \emph{utilization}. For each routing direction $a\in\{\mathrm{read},\mathrm{write}\}$, let $S^{(a)}$ be the token heads routed within a scope and $c_e^{(a)}$ the number that select expert $e$. Its utilization is $f_e^{(a)}=c_e^{(a)}/\sum_{e'}c_{e'}^{(a)}$, and its mean routing probability is $\bar p_e^{(a)}=|S^{(a)}|^{-1}\sum_{t\in S^{(a)}}p_{t,e}^{(a)}$, where $p_t^{(a)}$ is the softmax of the corresponding logits of \eqref{eq:topk} over the visible domain of token head $t$; the superscript is omitted for one direction. With \emph{layer} scope, each layer carries its own penalty or bias vector; with \emph{pool} scope, the statistics cover all $L$ layers that share the pool and one penalty or bias vector serves the whole pool, as in shared expert pools~\citep{huang2026unipool,chen2026moue}.

The \emph{auxiliary loss}~\citep{shazeer2017outrageously,fedus2022switch} adds to the training loss the load-balancing term of \citet{fedus2022switch},
\begin{equation}
\mathcal{L}_{\mathrm{bal}} \;=\; \lambda \, E \sum_{a\in\{\mathrm{read},\mathrm{write}\}}\sum_{e=1}^{E} \bar p_e^{(a)} \, f_e^{(a)} ,
\label{eq:aux-loss}
\end{equation}
with $\lambda$ the coefficient of Table~\ref{tab:lm-balancing}. Its statistics are computed over each data-parallel worker's local batch, read and write are normalized separately, and under layer scope the $L$ per-layer terms are added. The \emph{auxiliary-loss-free bias}~\citep{wang2024auxlossfree} instead uses the combined counts $c_e=c_e^{(\mathrm{read})}+c_e^{(\mathrm{write})}$, summed across data-parallel workers. It keeps a bias $b_e$ per expert in the scope and adds it to the scores that the top-$K$ selection of \eqref{eq:topk} ranks; after every training step
\begin{equation}
b_e \;\leftarrow\; b_e + u \,\operatorname{sign}\!\big(\bar c - c_e\big), \qquad \bar c = \tfrac{1}{E}\textstyle\sum_{e'} c_{e'},
\label{eq:loss-free}
\end{equation}
followed by re-centering to zero mean and clipping to $|b_e| \le 2$, with $u$ the \emph{bias-update rate}; inference keeps the trained bias fixed. TTT-MoE, whose layers each route over $E/L$ experts, uses $E/L$ in place of $E$ in \eqref{eq:aux-loss} and \eqref{eq:loss-free}. The gate of a selected expert is its routing probability times the \emph{routing-weight scale} $\gamma$, \eqref{eq:gate-scale} of Section~\ref{sec:method:uttt-moe}, with $\gamma=1$ everywhere except Table~\ref{tab:lm-alpha}.

\begin{figure}[tb]
    \centering
    \includegraphics[width=\textwidth]{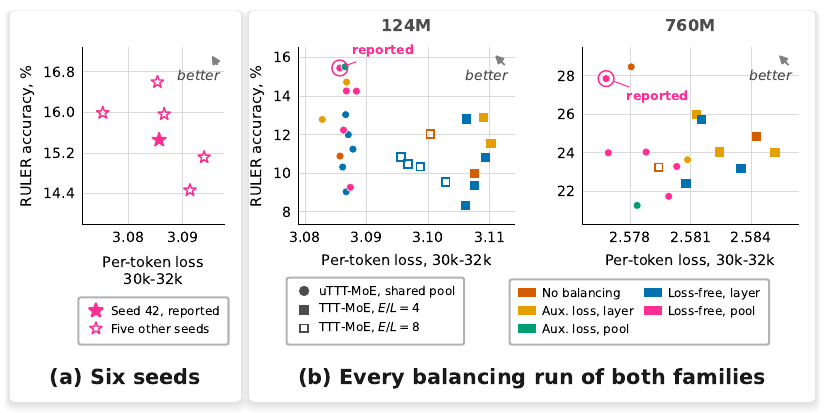}
    \caption{\textbf{Training seeds and every balancing run of both routed families}, on the axes of Figure~\ref{fig:teaser-pareto}. (a)~Six training seeds of the 124M uTTT-MoE configuration of Table~\ref{tab:lm-results}; the filled star is seed 42, the reported run. (b)~Every run of Table~\ref{tab:lm-balancing} (uTTT-MoE, circles) and Table~\ref{tab:lm-ttt-moe} (TTT-MoE; filled squares $E/L{=}4$, hollow $E/L{=}8$) at 124M and 760M. Colour is the balancing mechanism and the scope over which utilization is measured; the layer-private control has a single scope and takes the layer-scope colours. The ringed circle is the configuration reported as uTTT-MoE at that scale.}
    \label{fig:lm-sweeps}
\end{figure}

\begin{table}[tb]
    \centering
    \begin{minipage}[t]{0.483\textwidth}
        \centering
        \caption{\textbf{Every load-balancing run of uTTT-MoE}, at the default routing-weight scale $\gamma=1$ with a per-layer router; columns as in Table~\ref{tab:lm-results}. Mechanism, scope and the number in each row label are defined in Appendix~\ref{app:balancing-mechanisms}. $^{\ast}$ marks the row reported in Table~\ref{tab:lm-results}; bold marks the best value in each column within a scale.}
        \label{tab:lm-balancing}
        \fontsize{6pt}{7.3pt}\selectfont
        \setlength{\tabcolsep}{1.8pt}
        \renewcommand{\arraystretch}{1.12}
        \begin{tabular}{lcccccc}
        \toprule
        Balancing & PTL $\downarrow$ & RULER $\uparrow$ & 4K & 8K & 16K & 32K \\
        \midrule
        \tgroup{fw}{7}{124M, $E{=}48$}
        \trow{fw} No balancing & 3.0857 & 10.89 & 22.43 & 10.26 & 7.33 & 3.56 \\
        \trow{fw} Aux.\ loss $10^{-4}$, layer & 3.0868 & 14.73 & 24.73 & 16.93 & 10.94 & 6.32 \\
        \trow{fw} Aux.\ loss $10^{-3}$, layer & \textbf{3.0828} & 12.79 & 20.05 & 16.17 & 8.59 & 6.37 \\
        \trow{fw} Aux.\ loss $10^{-3}$, pool & 3.0865 & \textbf{15.52} & 26.44 & \textbf{17.70} & \textbf{11.44} & 6.49 \\
        \trow{fw} Loss-free $10^{-5}$, layer & 3.0866 & 13.04 & 25.56 & 13.12 & 9.07 & 4.42 \\
        \trow{fw} Loss-free $3{\times}10^{-5}$, layer & 3.0878 & 11.25 & 19.88 & 10.79 & 9.08 & 5.25 \\
        \trow{fw} Loss-free $10^{-4}$, layer & 3.0870 & 12.00 & 23.03 & 12.47 & 7.15 & 5.35 \\
        \trow{fw} Loss-free $3{\times}10^{-4}$, layer & 3.0861 & 10.32 & 19.79 & 10.63 & 6.15 & 4.71 \\
        \trow{fw} Loss-free $10^{-3}$, layer & 3.0867 & 9.04 & 18.79 & 9.02 & 4.83 & 3.54 \\
        \trow{fw} Loss-free $10^{-5}$, pool & 3.0863 & 12.24 & 22.28 & 14.73 & 7.61 & 4.34 \\
        \thl{ours} \textbf{Loss-free $3{\times}10^{-5}$, pool} $^{\ast}$ & 3.0857 & 15.46 & 26.50 & 17.32 & 11.23 & \textbf{6.78} \\
        \trow{fw} Loss-free $10^{-4}$, pool & 3.0867 & 14.27 & \textbf{26.95} & 15.53 & 9.43 & 5.19 \\
        \trow{fw} Loss-free $3{\times}10^{-4}$, pool & 3.0883 & 14.26 & 24.45 & 16.07 & 9.80 & 6.72 \\
        \trow{fw} Loss-free $10^{-3}$, pool & 3.0874 & 9.27 & 17.02 & 8.48 & 6.43 & 5.15 \\
        \tsep
        \tgroup{fw}{7}{760M, $E{=}96$}
        \trow{fw} No balancing & 2.5780 & \textbf{28.46} & 49.32 & 30.11 & \textbf{19.23} & \textbf{15.17} \\
        \trow{fw} Aux.\ loss $10^{-3}$, layer & 2.5808 & 23.64 & 48.64 & 23.27 & 14.42 & 8.23 \\
        \trow{fw} Aux.\ loss $10^{-3}$, pool & 2.5783 & 21.27 & 38.88 & 20.36 & 14.19 & 11.66 \\
        \trow{fw} Loss-free $10^{-5}$, pool & 2.5769 & 24.00 & 44.21 & 26.07 & 16.47 & 9.24 \\
        \trow{fw} Loss-free $3{\times}10^{-5}$, pool & 2.5803 & 23.29 & 45.21 & 22.55 & 14.18 & 11.23 \\
        \trow{fw} Loss-free $10^{-4}$, pool & 2.5788 & 24.04 & 49.79 & 24.73 & 13.85 & 7.79 \\
        \trow{fw} Loss-free $3{\times}10^{-4}$, pool & 2.5799 & 21.74 & 40.72 & 23.11 & 13.02 & 10.11 \\
        \thl{ours} \textbf{Loss-free $10^{-3}$, pool} $^{\ast}$ & \textbf{2.5768} & 27.85 & \textbf{54.15} & \textbf{31.00} & 16.03 & 10.23 \\
        \bottomrule
    \end{tabular}
    \end{minipage}\hfill
    \begin{minipage}[t]{0.494\textwidth}
        \centering
        \caption{\textbf{Every layer-private routed run (TTT-MoE)}, the control this paper compares against; columns as in Table~\ref{tab:lm-results}. $E/L$ is the number of experts per layer, one expert active per token head; the balancing setting reads as in Table~\ref{tab:lm-balancing}, without a scope. The tinted row is the one reported in Table~\ref{tab:lm-results} at each scale; bold as in Table~\ref{tab:lm-balancing}.}
        \label{tab:lm-ttt-moe}
        \fontsize{6pt}{7.3pt}\selectfont
        \setlength{\tabcolsep}{1.8pt}
        \renewcommand{\arraystretch}{1.12}
        \begin{tabular}{lcccccc}
        \toprule
        Configuration & PTL $\downarrow$ & RULER $\uparrow$ & 4K & 8K & 16K & 32K \\
        \midrule
        \tgroup{fw}{7}{124M}
        \trow{fw} $E/L{=}4$, no balancing & 3.1075 & 9.99 & 18.53 & 10.75 & 6.97 & 3.70 \\
        \trow{fw} $E/L{=}4$, loss-free $3{\times}10^{-6}$ & 3.1076 & 9.37 & 18.68 & 10.71 & 4.78 & 3.33 \\
        \thl{fw} \textbf{$E/L{=}4$, loss-free $10^{-5}$} & 3.1062 & 12.82 & \textbf{25.35} & 13.26 & 7.60 & 5.08 \\
        \trow{fw} $E/L{=}4$, loss-free $3{\times}10^{-5}$ & 3.1061 & 8.33 & 16.50 & 7.97 & 4.90 & 3.96 \\
        \trow{fw} $E/L{=}4$, loss-free $10^{-4}$ & 3.1093 & 10.82 & 23.33 & 9.30 & 6.33 & 4.32 \\
        \trow{fw} $E/L{=}4$, aux.\ loss $10^{-6}$ & 3.1102 & 11.53 & 23.15 & 11.40 & 6.59 & 4.97 \\
        \trow{fw} $E/L{=}4$, aux.\ loss $10^{-5}$ & 3.1090 & \textbf{12.90} & 22.01 & \textbf{15.64} & 7.72 & \textbf{6.21} \\
        \trow{fw} $E/L{=}8$, no balancing & 3.1003 & 12.04 & 24.61 & 10.83 & \textbf{7.79} & 4.94 \\
        \trow{fw} $E/L{=}8$, loss-free $3{\times}10^{-6}$ & 3.0967 & 10.48 & 21.64 & 9.49 & 6.41 & 4.40 \\
        \trow{fw} $E/L{=}8$, loss-free $10^{-5}$ & 3.0987 & 10.34 & 20.95 & 10.23 & 5.71 & 4.47 \\
        \trow{fw} $E/L{=}8$, loss-free $3{\times}10^{-5}$ & \textbf{3.0956} & 10.84 & 20.15 & 10.80 & 7.74 & 4.67 \\
        \trow{fw} $E/L{=}8$, loss-free $10^{-4}$ & 3.1028 & 9.54 & 17.56 & 10.55 & 6.46 & 3.59 \\
        \tsep
        \tgroup{fw}{7}{760M}
        \trow{fw} $E/L{=}4$, no balancing & 2.5843 & 24.86 & 45.57 & 25.37 & 16.17 & 12.33 \\
        \thl{fw} \textbf{$E/L{=}4$, loss-free $10^{-5}$} & 2.5815 & 25.73 & 45.27 & \textbf{29.44} & 17.85 & 10.35 \\
        \trow{fw} $E/L{=}4$, loss-free $3{\times}10^{-5}$ & 2.5835 & 23.17 & 44.05 & 24.49 & 14.33 & 9.83 \\
        \trow{fw} $E/L{=}4$, loss-free $10^{-4}$ & 2.5808 & 22.40 & 38.94 & 26.04 & 14.64 & 9.98 \\
        \trow{fw} $E/L{=}4$, aux.\ loss $10^{-5}$ & 2.5813 & \textbf{25.98} & 45.82 & 25.34 & \textbf{18.79} & \textbf{13.98} \\
        \trow{fw} $E/L{=}4$, aux.\ loss $3{\times}10^{-5}$ & 2.5824 & 24.05 & 41.90 & 24.58 & 17.27 & 12.46 \\
        \trow{fw} $E/L{=}4$, aux.\ loss $10^{-4}$ & 2.5852 & 24.02 & 43.52 & 24.58 & 15.65 & 12.35 \\
        \trow{fw} $E/L{=}8$, no balancing & \textbf{2.5794} & 23.24 & \textbf{46.29} & 22.31 & 14.50 & 9.88 \\
        \bottomrule
    \end{tabular}
    \end{minipage}
\end{table}

\begin{table}[tb]
    \centering
    \small
    \caption{\textbf{The routing-weight scale at 760M.} Pool scope and a per-layer router throughout; $\gamma$ scales the gates as in \eqref{eq:gate-scale} and is 1 everywhere else in this paper. Each cell is PTL and RULER (\%); a dash marks a combination that was not trained. The $\gamma{=}1$ column is the pool-scope part of the 760M block of Table~\ref{tab:lm-balancing}.}
    \label{tab:lm-alpha}
    \setlength{\tabcolsep}{5pt}
    \begin{tabular}{@{}lcccccc@{}}
        \toprule
        & \multicolumn{2}{c}{$\gamma=1$} & \multicolumn{2}{c}{$\gamma=2$} & \multicolumn{2}{c}{$\gamma=4$} \\
        \cmidrule(lr){2-3} \cmidrule(lr){4-5} \cmidrule(l){6-7}
        Balancing & PTL & RULER & PTL & RULER & PTL & RULER \\
        \midrule
        Aux.\ loss $10^{-4}$ & --- & --- & 2.5744 & 23.16 & --- & --- \\
        Aux.\ loss $5{\times}10^{-4}$ & --- & --- & 2.5780 & 23.96 & 2.5805 & 24.19 \\
        Aux.\ loss $10^{-3}$ & 2.5783 & 21.27 & 2.5783 & 25.50 & 2.5791 & 26.07 \\
        Aux.\ loss $2{\times}10^{-3}$ & --- & --- & 2.5793 & 20.57 & 2.5794 & 25.56 \\
        Aux.\ loss $5{\times}10^{-3}$ & --- & --- & 2.5771 & 25.11 & --- & --- \\
        Loss-free $10^{-5}$ & 2.5769 & 24.00 & --- & --- & --- & --- \\
        Loss-free $3{\times}10^{-5}$ & 2.5803 & 23.29 & 2.5764 & 25.57 & --- & --- \\
        Loss-free $10^{-4}$ & 2.5788 & 24.04 & 2.5735 & 25.21 & 2.5752 & 22.37 \\
        Loss-free $3{\times}10^{-4}$ & 2.5799 & 21.74 & 2.5757 & 25.68 & --- & --- \\
        Loss-free $10^{-3}$ & 2.5768 & 27.85 & --- & --- & --- & --- \\
        \bottomrule
    \end{tabular}
\end{table}

\paragraph{Router utilization.}
The busiest-expert ratio $E\max_e f_e$ of Section~\ref{sec:llm:balancing} is taken from the read routing at the last training step; the write routing gives $6.3\times$ and $1.5\times$ at 124M and $10.8\times$ and $2.1\times$ at 760M for the same unbalanced and reported runs. The unbalanced routers leave $3$ of $48$ experts below a quarter of an even share at 124M and $74$ of $96$ at 760M, the reported ones $0$ and $3$. Counted per layer, layer scope leaves under $10\%$ of the experts below that share at 124M, whereas pool scope leaves $41$--$49\%$ at 124M and $89$--$92\%$ at 760M while at most $5\%$ of the whole pool falls below it: each layer draws on about half of the pool at 124M and on eight to eleven of the $96$ experts at 760M. The two smallest loss-free rates at 760M leave $44\%$ and $12\%$ of the whole pool below that share. The four-expert layer-private routers leave none at 124M and $3$--$19\%$ at 760M.

\begin{figure}[tb]
    \centering
    \includegraphics[width=\textwidth]{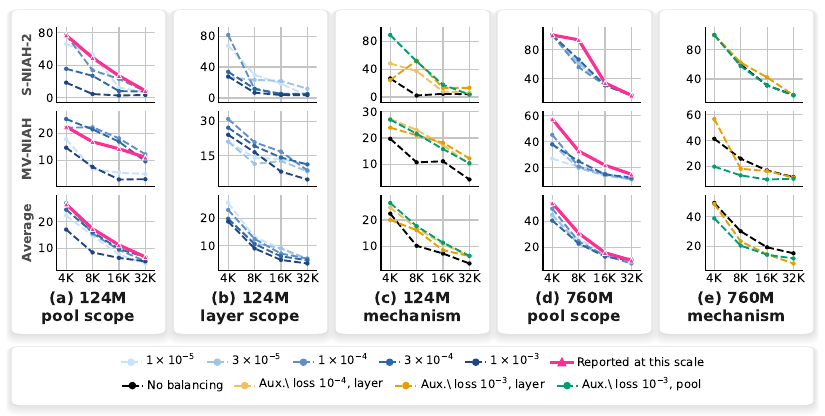}
    \caption{\textbf{Every run of Table~\ref{tab:lm-balancing}, resolved by task.} Cards (a), (b) and (d) sweep the loss-free bias-update rate at 124M pool scope, 124M layer scope and 760M pool scope; cards (c) and (e) draw the unbalanced and auxiliary-loss runs at 124M and 760M. Each card holds the two tasks of Figure~\ref{fig:ruler-tasks}(a) and the unweighted mean over all thirteen RULER tasks, whose four points are the 4K--32K columns of Table~\ref{tab:lm-balancing}; the cards share their rows, so a task reads across the figure. Rates are ordinal and drawn light to dark; the pink line is the configuration reported at that scale, present only in the two pool-scope sweeps. Each panel carries its own accuracy range.}
    \label{fig:balancing-tasks}
\end{figure}

\paragraph{Seeds and sweeps.}
\label{app:lm-seeds}
\label{app:balancing-tasks}
Figure~\ref{fig:lm-sweeps}a shows six training seeds of the 124M uTTT-MoE configuration, spanning $14.47$ to $16.60$ RULER and $3.0753$ to $3.0939$ in per-token loss. Tables~\ref{tab:lm-balancing} and~\ref{tab:lm-ttt-moe} list every balancing run of the two routed families, Figure~\ref{fig:lm-sweeps}b plots them, Figure~\ref{fig:balancing-tasks} resolves the uTTT-MoE runs by task, and Table~\ref{tab:lm-alpha} varies the routing-weight scale at 760M. The 760M uTTT-MoE configuration keeps the mechanism and scope selected at 124M, with the best of the five bias-update rates tried at 760M. Eight private experts per layer, twice the total state of the shared pool, lower the 124M loss of TTT-MoE at every matched setting but do not raise its best RULER.

\subsection{A Dense Shared State at 124M}
\label{app:uttt-dense-llm}
Table~\ref{tab:lm-uttt-dense} reports the dense shared state of Section~\ref{sec:method:dense}, trained at 124M over a ladder of widths $r$.

\begin{table}[tb]
    \centering
    \small
    \caption{\textbf{A dense shared state at 124M.} The shared state is $r$ times the width of one layer-private state, so $r{=}1$ matches TTT-Dense in per-token fast-weight compute and $r{=}12$, the depth of the model, would match it in size; columns as in Table~\ref{tab:lm-results}; the TTT-Dense row repeats Table~\ref{tab:lm-results}.}
    \label{tab:lm-uttt-dense}
    \setlength{\tabcolsep}{5pt}
    \begin{tabular*}{\textwidth}{@{\extracolsep{\fill}}lccccccc@{}}
        \toprule
        Model & State (M) & PTL $\downarrow$ & RULER $\uparrow$ & 4K & 8K & 16K & 32K \\
        \midrule
        uTTT-Dense, $r{=}1$ (FLOPs-matched) & 0.44 & 3.1290 & 10.64 & 20.27 & 12.23 & 6.46 & 3.61 \\
        uTTT-Dense, $r{=}2$ & 0.88 & 3.1232 & 13.20 & 23.57 & 14.66 & 9.79 & 4.78 \\
        uTTT-Dense, $r{=}4$ & 1.77 & 3.1249 & 13.41 & 21.65 & 15.12 & 10.69 & 6.20 \\
        uTTT-Dense, $r{=}6$ & 2.65 & 3.1165 & 13.63 & 23.60 & 16.01 & 9.92 & 5.01 \\
        \midrule
        TTT-Dense (Table~\ref{tab:lm-results}) & 5.31 & 3.1219 & 12.65 & 26.31 & 13.58 & 6.00 & 4.72 \\
        \bottomrule
    \end{tabular*}
\end{table}

\FloatBarrier

\section{Length Extrapolation}
\label{app:extrapolation}

This appendix evaluates seven 124M models beyond the $32$K training length, per-token loss to $128$K and retrieval to $64$K (Table~\ref{tab:extrap-ruler}): five trained at $32$K and two attention references trained at $64$K (Appendix~\ref{app:configs}). These models are trained separately from those of Table~\ref{tab:lm-results}, so their scores are not on the scale of that table.

\begin{figure}[htbp]
    \centering
    \includegraphics[width=\linewidth]{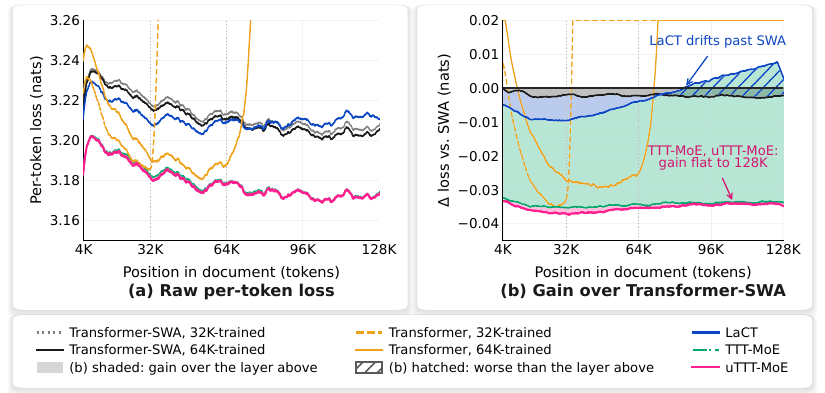}
    \caption{\textbf{Per-token loss along $3{,}000$ Books3 documents (positions $4$K--$128$K)}, smoothed over $6{,}144$ tokens. \emph{(a)} Loss of the seven models. \emph{(b)} Loss relative to the 32K-trained Transformer-SWA with a $4$K attention window. Shaded bands show successive loss differences in the order SWA (64K), LaCT, TTT-MoE, uTTT-MoE, starting from zero; hatching marks a reversal of the ordering. Full-attention models are lines only, clipped at the top. Dotted lines mark the $32$K and $64$K training lengths; LaCT and both expert pools are 32K-trained.}
    \label{fig:extrap-ptl}
\end{figure}

Per-position loss confounds context length with document depth: later tokens may be easier even without access to more history. We therefore pair the full-document curves (Figure~\ref{fig:extrap-ptl}; $3{,}000$ Books3 documents of at least $256$K tokens, each read from its first token to $131{,}072$) with two controls on the same documents (Table~\ref{tab:extrap-context}): a \emph{fixed-target ablation}, which scores the same $8{,}192$ tokens at positions $120$--$128$K while the visible prefix grows from $4$K to $120$K, on a $512$-document subset; and \emph{reset segments}, which evaluate the same $128$K as four consecutive $32$K pieces, each from position $0$. Means exclude each row's final position.

In the fixed-target ablation (Table~\ref{tab:extrap-context}, left), the 32K-trained Transformer-SWA stays at $3.222$ nats at the reported precision from a $4$K to a $120$K prefix, so it serves as a position-matched reference. With context reset every $32$K (Table~\ref{tab:extrap-context}, right), every model still improves by $0.010$--$0.015$ nats from the first segment to the fourth: later text is easier by itself, so we compare models against this reference rather than by raw loss.
\begin{table}[htbp]
    \centering
    \footnotesize
    \setlength{\tabcolsep}{1.8pt}
    \caption{\textbf{Fixed-target ablation and reset segments}, loss in nats. Left: the same $8{,}192$ target tokens as the visible prefix grows, each prefix and its targets forming one sequence; $\dagger$: sequence beyond the model's trained length; bold marks each row's minimum before rounding. Right: the four consecutive $32$K pieces of the same $128$K. Transformer-SWA has a $4$K attention window.}
    \label{tab:extrap-context}
    \begin{tabular*}{\textwidth}{@{\extracolsep{\fill}}l rrrrrrr @{\hspace{14pt}} rrrr@{}}
\toprule
 & \multicolumn{7}{c}{Fixed-target ablation: visible prefix} & \multicolumn{4}{c}{Reset segments} \\
\cmidrule(lr){2-8} \cmidrule(l){9-12}
Model & 4K & 8K & 16K & 32K & 48K & 64K & 120K & Seg.\ 1 & Seg.\ 2 & Seg.\ 3 & Seg.\ 4 \\
\midrule
Transformer-SWA (32K) & 3.222 & \textbf{3.222} & \textbf{3.222} & \textbf{3.222}$^\dagger$ & \textbf{3.222}$^\dagger$ & \textbf{3.222}$^\dagger$ & \textbf{3.222}$^\dagger$ & 3.224 & 3.224 & 3.218 & 3.214 \\
Transformer-SWA (64K) & 3.220 & \textbf{3.220} & 3.220 & 3.220 & 3.220 & \textbf{3.220}$^\dagger$ & 3.220$^\dagger$ & 3.223 & 3.222 & 3.216 & 3.212 \\
Transformer (32K) & 3.211 & 3.198 & \textbf{3.188} & 3.329$^\dagger$ & 4.920$^\dagger$ & 5.692$^\dagger$ & 5.838$^\dagger$ & 3.205 & 3.201 & 3.194 & 3.190 \\
Transformer (64K) & 3.226 & 3.212 & 3.201 & \textbf{3.193} & 3.194 & 3.208$^\dagger$ & 5.563$^\dagger$ & 3.218 & 3.215 & 3.208 & 3.204 \\
\midrule
LaCT & 3.215 & 3.214 & \textbf{3.213} & 3.213$^\dagger$ & 3.216$^\dagger$ & 3.219$^\dagger$ & 3.230$^\dagger$ & 3.216 & 3.216 & 3.209 & 3.205 \\
TTT-MoE & 3.189 & 3.188 & 3.187 & \textbf{3.187}$^\dagger$ & 3.188$^\dagger$ & 3.188$^\dagger$ & 3.189$^\dagger$ & 3.190 & 3.190 & 3.184 & 3.180 \\
uTTT-MoE & 3.189 & 3.187 & 3.186 & \textbf{3.185}$^\dagger$ & 3.186$^\dagger$ & 3.187$^\dagger$ & 3.188$^\dagger$ & 3.189 & 3.189 & 3.183 & 3.179 \\
\bottomrule
\end{tabular*}

\end{table}

Figure~\ref{fig:extrap-ptl}(b) shows each model's loss relative to the reference at the same positions, $\Delta(p) = L_{\text{model}}(p) - L_{\text{SWA}}(p)$. Within $32$K, uTTT-MoE has the lowest curve; the expert pools sit about $0.035$ nats below SWA and LaCT about $0.010$, while the two SWA references stay close. Both expert pools keep this advantage to $128$K, whereas LaCT crosses zero near $85$K and full attention deteriorates sharply beyond its trained length, later for the 64K-trained reference.
\begin{table}[htbp]
    \centering
    \small
    \caption{\textbf{RULER accuracy (\%) by evaluation length}, $500$ samples per task per length, averaged over the $13$ tasks of Figure~\ref{fig:ruler-tasks}. Bold marks column maxima at the displayed precision. $\dagger$: beyond the trained length. Model configurations are given in Appendix~\ref{app:configs}.}
    \label{tab:extrap-ruler}
    \begin{tabular}{lrrrrr}
\toprule
Model & 4K & 8K & 16K & 32K & 64K \\
\midrule
Transformer-SWA, 4K window (32K) & 29.1 & 15.6 & 9.8 & 6.0 & 3.4$^\dagger$ \\
Transformer-SWA, 4K window (64K) & 28.5 & 16.0 & 9.5 & 6.0 & 3.8 \\
Transformer (32K) & \textbf{30.9} & \textbf{26.4} & \textbf{23.0} & \textbf{13.8} & 1.4$^\dagger$ \\
Transformer (64K) & 20.5 & 10.7 & 11.4 & 8.4 & \textbf{5.8} \\
\midrule
LaCT & 23.6 & 14.8 & 9.3 & 5.2 & 3.3$^\dagger$ \\
TTT-MoE & 28.7 & 13.3 & 6.4 & 4.2 & 3.3$^\dagger$ \\
uTTT-MoE & \textbf{30.9} & 17.7 & 9.8 & 5.9 & 3.7$^\dagger$ \\
\bottomrule
\end{tabular}

\end{table}
\FloatBarrier

\section{Efficiency measurements}
\label{app:efficiency}

Figure~\ref{fig:efficiency-main}a reports median per-GPU throughput over steps 50--100 of a 100-step run. Forward FLOPs per token differ by at most $0.6\%$ across the three models, and all use blockwise recomputation to reduce training memory.

\begin{figure}[tb]
    \centering
    \includegraphics[width=\textwidth]{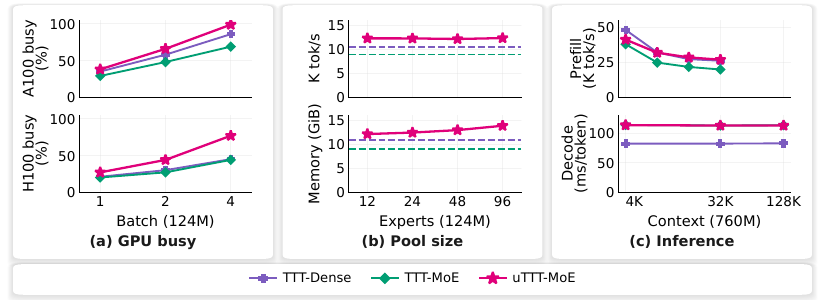}
    \caption{\textbf{Efficiency beyond Figure~\ref{fig:efficiency-main}.} \emph{(a)}~GPU busy time at 124M against batch per GPU, on one node of eight A100s (top) or H100s (bottom). \emph{(b)}~uTTT-MoE training throughput (top) and peak memory (bottom) against shared-pool size at 124M on one A100, with TTT-Dense and TTT-MoE in the same setting dashed. \emph{(c)}~760M inference against context length: batch-1 prefill throughput (top, one A100, 4K--32K contexts) and per-token decoding latency (bottom, one A100; TTT-MoE and uTTT-MoE coincide).}
    \label{fig:eff-appendix}
\end{figure}

At 124M, the throughput advantage depends on hardware and batch size. On H100, uTTT-MoE is $1.14$--$1.16\times$ faster than TTT-Dense across the tested batches. In the eight-GPU training runs at 124M and 760M with batch~1 per GPU, peak memory is $1.29$--$1.37\times$ TTT-Dense's. Increasing the shared pool from 12 to 96 experts at 124M changes training throughput by less than $2\%$ and adds $1.7$\,GiB of peak memory (Figure~\ref{fig:eff-appendix}b). With top-$K$ fixed, the larger pool expands accessible fast-weight state while keeping prefill throughput and per-token decoding latency nearly unchanged at batch~1 and 32K context on one A100.

TTT-MoE and uTTT-MoE use the same fused Triton kernels. Table~\ref{tab:eff-kernels} ablates these kernels in uTTT-MoE by replacing each group with split references of the same arithmetic. Together they improve throughput by $15\%$ at 124M and $18\%$ at 760M, chiefly through routing permutation and, at 760M, the backward pass. With all fusion disabled, uTTT-MoE matches TTT-Dense at 124M and is $6\%$ slower at 760M in this single-GPU setting. It remains $19\%$ and $15\%$ faster, respectively, than TTT-MoE with fusion.

\begin{table}[tb]
    \centering
    \small
    \caption{\textbf{Fused-kernel ablation of uTTT-MoE training} at batch $1\times32$K on one A100. Training tokens per second per GPU, median of three repeats, and the speed-up the group gives; peak memory is the same in every row.}
    \label{tab:eff-kernels}
    \begin{tabular}{@{}lrrrr@{}}
        \toprule
        & \multicolumn{2}{c}{124M} & \multicolumn{2}{c}{760M} \\
        \cmidrule(lr){2-3} \cmidrule(l){4-5}
        Kernels switched off & tok/s & speed-up & tok/s & speed-up \\
        \midrule
        none & 12{,}071 & -- & 5{,}542 & -- \\
        all & 10{,}533 & 1.15 & 4{,}710 & 1.18 \\
        routing permutation of $k$, $v$ and learning rates & 11{,}282 & 1.07 & 5{,}088 & 1.09 \\
        grouped GEMMs of the outer-gradient backward & 11{,}816 & 1.02 & 5{,}158 & 1.07 \\
        pointwise second-order backward & 11{,}950 & 1.01 & 5{,}162 & 1.07 \\
        inner gradient of the fast-weight write & 12{,}033 & 1.00 & 5{,}478 & 1.01 \\
        fast-weight read & 12{,}197 & 0.99 & 5{,}498 & 1.01 \\
        \bottomrule
    \end{tabular}
\end{table}

At 760M on one A100, uTTT-MoE prefills at batch~1 as fast as TTT-Dense and $1.3\times$ faster than TTT-MoE at the tested 16K and 32K prompt lengths (Figure~\ref{fig:eff-appendix}c). Increasing batch size from 1 to 4 brings little improvement in prefill throughput at these lengths. It decodes at TTT-MoE's speed, $0.72$--$0.74\times$ TTT-Dense's across the tested 4K--128K contexts. At batch~1, time to first token is $29$--$33\%$ lower than TTT-MoE's at 32K and 128K. Chunk-boundary pauses, measured as the maximum boundary-token step latency per run, are $26$--$27\%$ shorter at the tested 4K, 32K and 128K contexts. Both latency metrics are comparable to TTT-Dense in these settings; comparisons use medians of three runs.

\FloatBarrier

\begingroup\let\FloatBarrier\relax
\section{Additional Novel View Synthesis Comparisons}
\label{app:nvs-perview}

This appendix holds the novel-view-synthesis results that Section~\ref{sec:exp-nvs} refers to but does not display: the modules of Figure~\ref{fig:nvs-psnr-by-view} on DL3DV, the remaining modules on both datasets, every configuration with its intervals, and the comparison of the two write schedules. Protocol and notation are those of Section~\ref{sec:nvs:setup} and Figure~\ref{fig:nvs-psnr-by-view}.

\subsection{Sharing and Compute on Scenes}
\label{app:nvs-dl3dv}
Figure~\ref{fig:nvs-s2} repeats the four modules of Figure~\ref{fig:nvs-psnr-by-view} on DL3DV scenes. Row~B carries over: activating more experts raises PSNR on scenes as on objects, by $0.54$~dB over all views from $a{=}1$ to $a{=}8$ with non-overlapping intervals, and widening the dense state raises it by $0.31$~dB. On DL3DV, the four sharing levels in row~A differ by at most $0.10$~dB over all views. The sharing conclusion therefore rests on objects, whose test set is seven times larger; the compute conclusion holds on both datasets.

\begin{figure}[!htbp]
    \centering
    \includegraphics[width=\textwidth]{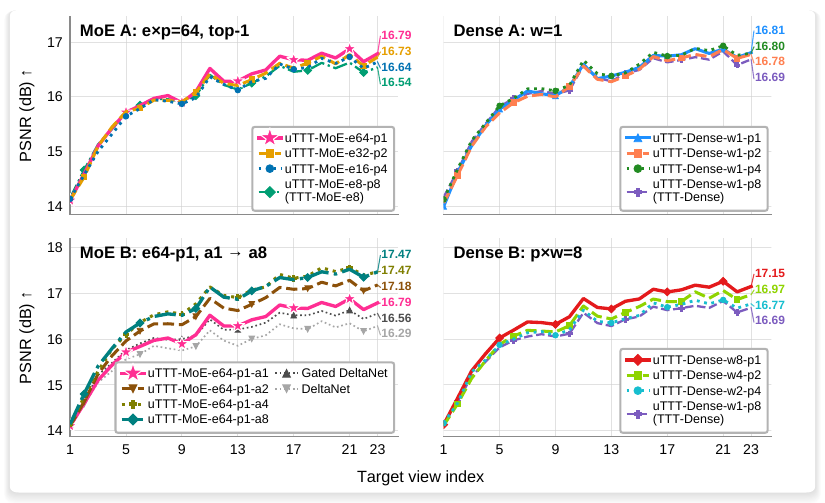}
    \caption{\textbf{The four modules of Figure~\ref{fig:nvs-psnr-by-view} on DL3DV}; conventions as there.}
    \label{fig:nvs-s2}
\end{figure}

\subsection{Total Experts and Pool Size}
\label{app:nvs-modules}
Rows~A and~B hold the total at 64 experts. MoE~A$'$ and B$'$ (Figures~\ref{fig:nvs-s1} and~\ref{fig:nvs-s3}) repeat both sweeps at 32 experts in total, with TTT-MoE-e4 as the layer-private end point, to check that the orderings are properties of sharing and of compute rather than of one total. Both hold: quality rises at every merge on Objaverse/GSO and the activation sweep is monotone on both datasets, with non-overlapping intervals between the end points of both sweeps on objects and of the activation sweep on scenes.

Module~C separates sharing from the wider routing choice that comes with it: sharing is fixed at $p{=}1$, per-layer compute at $a{=}1$, and only the pool size varies, from 8 to 64 experts. If choice or stored state were what mattered, quality would rise with the pool; the four pools instead finish within $0.25$~dB of one another on objects, the smallest the best, and within $0.16$~dB on scenes, and on Objaverse/GSO the 8-expert pool leads TTT-MoE-e8 over all views with non-overlapping intervals at one eighth of its experts and the same eight candidates per token head (Section~\ref{sec:nvs:ownership}). Row~A therefore measures that the state a layer reads was written by every layer, not that it has more to choose from. The dense state has no such control, since widening it raises per-layer compute together with state; the width sweep of module~C is the dense counterpart of row~B and behaves like one, width 8 leading on both datasets while the intermediate widths are not ordered on objects.

\begin{figure}[!htbp]
    \centering
    \includegraphics[width=\textwidth]{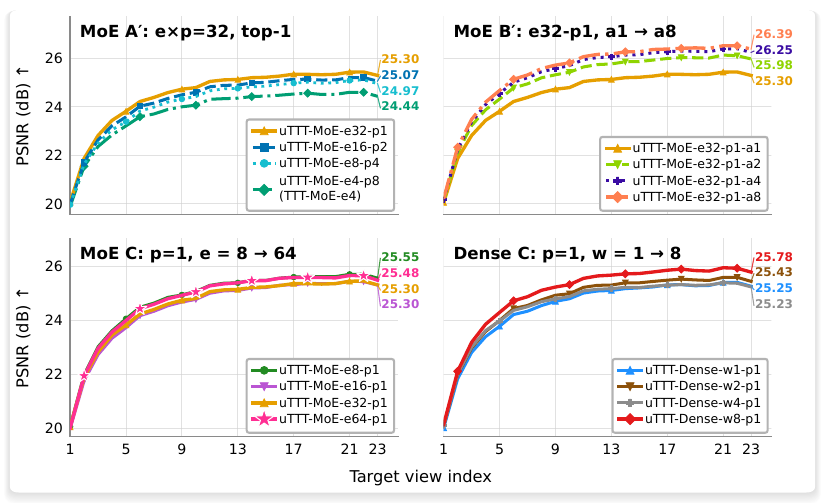}
    \caption{\textbf{The remaining modules on Objaverse/GSO}; conventions as Figure~\ref{fig:nvs-psnr-by-view}. MoE~A$'$ and B$'$ repeat rows~A and~B at 32 experts in total, with TTT-MoE-e4 as the $p{=}8$ endpoint. C sweeps the pool size under full sharing ($p{=}1$): per-layer compute is fixed for the routed pool ($a{=}1$) and scales with $w$ for the dense state.}
    \label{fig:nvs-s1}
\end{figure}

\begin{figure}[!htbp]
    \centering
    \includegraphics[width=\textwidth]{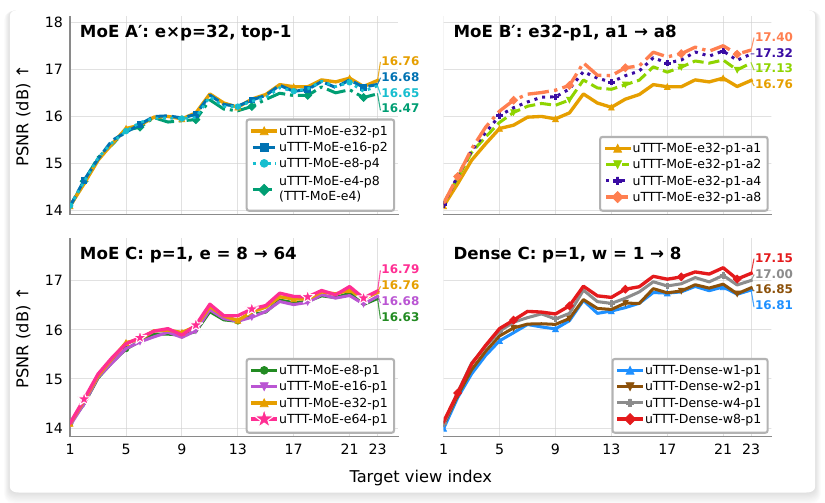}
    \caption{\textbf{The modules of Figure~\ref{fig:nvs-s1} on DL3DV}; conventions as Figure~\ref{fig:nvs-psnr-by-view}.}
    \label{fig:nvs-s3}
\end{figure}

\subsection{Every Configuration}
\label{app:nvs-full-tables}
Tables~\ref{tab:nvs-gso-full} and~\ref{tab:nvs-dl3dv-full} list every fast-weight configuration and the reference models on each dataset in one order: the layer-private pools, the universal pool at $a{=}1$, the depth partitions at fixed total, the activation sweeps, the universal dense state and the dense partitions.

\begin{table}[!t]
    \centering
    \scriptsize
    \caption{\textbf{Every fast-weight configuration on Objaverse/GSO} ($n{=}1{,}019$ objects). Cells: mean over objects of each object's mean over target views 1--23, with its 95\% bootstrap interval over objects; the second PSNR column is view~23 alone. Bold: best per column among fast-weight models; $^\dagger$full-attention reference, not bolded.}
    \label{tab:nvs-gso-full}
    \setlength{\tabcolsep}{3.0pt}
    \renewcommand{\arraystretch}{1.12}
    \begin{tabularx}{\textwidth}{l*{4}{>{\centering\arraybackslash}X}}
        \toprule
        Model & PSNR $\uparrow$ & View 23 PSNR $\uparrow$ & SSIM $\uparrow$ & LPIPS $\downarrow$ \\
        \midrule
        \tgroup{fw}{5}{Layer-private pools ($p{=}8$)}
        \trow{fw} uTTT-Dense-w1-p8 (TTT-Dense) & 23.99 {\tiny[23.81, 24.16]} & 24.49 {\tiny[24.29, 24.69]} & 0.852 {\tiny[0.847, 0.856]} & 0.185 {\tiny[0.180, 0.189]} \\
        \trow{fw} uTTT-MoE-e4-p8 (TTT-MoE-e4) & 23.80 {\tiny[23.62, 23.97]} & 24.44 {\tiny[24.24, 24.64]} & 0.849 {\tiny[0.845, 0.854]} & 0.188 {\tiny[0.183, 0.193]} \\
        \trow{fw} uTTT-MoE-e8-p8 (TTT-MoE-e8) & 23.90 {\tiny[23.73, 24.08]} & 24.56 {\tiny[24.36, 24.77]} & 0.850 {\tiny[0.845, 0.855]} & 0.186 {\tiny[0.181, 0.191]} \\
        \tsep
        \tgroup{moeC}{5}{MoE, universal pool, top-1 (C)}
        \trow{moeC} uTTT-MoE-e8-p1 & 24.72 {\tiny[24.54, 24.90]} & 25.55 {\tiny[25.34, 25.77]} & 0.861 {\tiny[0.856, 0.865]} & 0.172 {\tiny[0.168, 0.177]} \\
        \trow{moeC} uTTT-MoE-e16-p1 & 24.45 {\tiny[24.27, 24.63]} & 25.30 {\tiny[25.09, 25.50]} & 0.857 {\tiny[0.853, 0.862]} & 0.176 {\tiny[0.172, 0.181]} \\
        \trow{moeC} uTTT-MoE-e32-p1 & 24.49 {\tiny[24.32, 24.67]} & 25.30 {\tiny[25.10, 25.51]} & 0.858 {\tiny[0.854, 0.863]} & 0.175 {\tiny[0.170, 0.179]} \\
        \trow{moeC} uTTT-MoE-e64-p1 & 24.68 {\tiny[24.51, 24.86]} & 25.48 {\tiny[25.27, 25.68]} & 0.860 {\tiny[0.855, 0.864]} & 0.173 {\tiny[0.168, 0.177]} \\
        \tsep
        \tgroup{moeA}{5}{MoE, depth partitions at fixed total (A, A$'$)}
        \trow{moeA} uTTT-MoE-e32-p2 & 24.20 {\tiny[24.03, 24.37]} & 24.98 {\tiny[24.78, 25.18]} & 0.854 {\tiny[0.850, 0.859]} & 0.181 {\tiny[0.177, 0.186]} \\
        \trow{moeA} uTTT-MoE-e16-p4 & 24.11 {\tiny[23.94, 24.29]} & 24.89 {\tiny[24.69, 25.10]} & 0.854 {\tiny[0.849, 0.858]} & 0.182 {\tiny[0.177, 0.186]} \\
        \trow{moeA} uTTT-MoE-e16-p2 & 24.28 {\tiny[24.11, 24.46]} & 25.07 {\tiny[24.87, 25.27]} & 0.855 {\tiny[0.850, 0.859]} & 0.180 {\tiny[0.176, 0.185]} \\
        \trow{moeA} uTTT-MoE-e8-p4 & 24.14 {\tiny[23.97, 24.32]} & 24.97 {\tiny[24.76, 25.17]} & 0.854 {\tiny[0.849, 0.858]} & 0.181 {\tiny[0.177, 0.186]} \\
        \tsep
        \tgroup{moeB}{5}{MoE, active width (B, B$'$)}
        \trow{moeB} uTTT-MoE-e64-p1-a2 & 25.18 {\tiny[25.00, 25.36]} & 26.12 {\tiny[25.91, 26.33]} & 0.867 {\tiny[0.862, 0.871]} & 0.164 {\tiny[0.159, 0.168]} \\
        \trow{moeB} uTTT-MoE-e64-p1-a4 & 25.38 {\tiny[25.20, 25.56]} & 26.35 {\tiny[26.14, 26.56]} & 0.869 {\tiny[0.864, 0.873]} & 0.161 {\tiny[0.156, 0.165]} \\
        \trow{moeB} uTTT-MoE-e64-p1-a8 & \textbf{25.56} {\tiny[25.38, 25.74]} & \textbf{26.55} {\tiny[26.34, 26.76]} & \textbf{0.871} {\tiny[0.867, 0.876]} & \textbf{0.158} {\tiny[0.154, 0.163]} \\
        \trow{moeB} uTTT-MoE-e32-p1-a2 & 25.07 {\tiny[24.89, 25.25]} & 25.98 {\tiny[25.77, 26.20]} & 0.865 {\tiny[0.861, 0.870]} & 0.165 {\tiny[0.161, 0.170]} \\
        \trow{moeB} uTTT-MoE-e32-p1-a4 & 25.32 {\tiny[25.14, 25.50]} & 26.25 {\tiny[26.05, 26.47]} & 0.868 {\tiny[0.863, 0.872]} & 0.163 {\tiny[0.158, 0.167]} \\
        \trow{moeB} uTTT-MoE-e32-p1-a8 & 25.44 {\tiny[25.26, 25.62]} & 26.39 {\tiny[26.18, 26.61]} & 0.869 {\tiny[0.865, 0.873]} & 0.160 {\tiny[0.156, 0.165]} \\
        \tsep
        \tgroup{denB}{5}{Dense, universal state (C)}
        \trow{denB} uTTT-Dense-w1-p1 & 24.45 {\tiny[24.27, 24.62]} & 25.25 {\tiny[25.05, 25.46]} & 0.858 {\tiny[0.853, 0.862]} & 0.176 {\tiny[0.172, 0.181]} \\
        \trow{denB} uTTT-Dense-w2-p1 & 24.64 {\tiny[24.46, 24.82]} & 25.43 {\tiny[25.22, 25.64]} & 0.860 {\tiny[0.855, 0.864]} & 0.174 {\tiny[0.169, 0.178]} \\
        \trow{denB} uTTT-Dense-w4-p1 & 24.53 {\tiny[24.34, 24.71]} & 25.23 {\tiny[25.02, 25.44]} & 0.858 {\tiny[0.854, 0.863]} & 0.175 {\tiny[0.170, 0.179]} \\
        \trow{denB} uTTT-Dense-w8-p1 & 24.95 {\tiny[24.77, 25.13]} & 25.78 {\tiny[25.57, 25.99]} & 0.864 {\tiny[0.859, 0.868]} & 0.167 {\tiny[0.163, 0.172]} \\
        \tsep
        \tgroup{denA}{5}{Dense, depth partitions (B, A)}
        \trow{denA} uTTT-Dense-w4-p2 & 24.75 {\tiny[24.58, 24.93]} & 25.56 {\tiny[25.36, 25.77]} & 0.861 {\tiny[0.857, 0.866]} & 0.172 {\tiny[0.167, 0.176]} \\
        \trow{denA} uTTT-Dense-w2-p4 & 24.53 {\tiny[24.35, 24.71]} & 25.30 {\tiny[25.09, 25.51]} & 0.859 {\tiny[0.854, 0.863]} & 0.174 {\tiny[0.169, 0.179]} \\
        \trow{denA} uTTT-Dense-w1-p2 & 24.35 {\tiny[24.17, 24.53]} & 25.14 {\tiny[24.93, 25.34]} & 0.856 {\tiny[0.852, 0.861]} & 0.179 {\tiny[0.175, 0.184]} \\
        \trow{denA} uTTT-Dense-w1-p4 & 24.37 {\tiny[24.20, 24.55]} & 25.19 {\tiny[24.98, 25.40]} & 0.858 {\tiny[0.853, 0.862]} & 0.177 {\tiny[0.172, 0.182]} \\
        \tsep
        \tgroup{ref}{5}{Baseline}
        \trow{ref} Transformer$^\dagger$ & 25.36 {\tiny[25.19, 25.54]} & 26.54 {\tiny[26.33, 26.74]} & 0.870 {\tiny[0.865, 0.874]} & 0.162 {\tiny[0.158, 0.166]} \\
        \trow{ref} Gated DeltaNet & 24.32 {\tiny[24.14, 24.50]} & 25.01 {\tiny[24.80, 25.21]} & 0.856 {\tiny[0.851, 0.860]} & 0.178 {\tiny[0.173, 0.182]} \\
        \trow{ref} DeltaNet & 23.94 {\tiny[23.76, 24.12]} & 24.55 {\tiny[24.34, 24.75]} & 0.851 {\tiny[0.846, 0.856]} & 0.185 {\tiny[0.180, 0.189]} \\
        \bottomrule
    \end{tabularx}
\end{table}

\begin{table}[!t]
    \centering
    \scriptsize
    \caption{\textbf{Every fast-weight configuration on DL3DV} ($n{=}140$ test scenes); format as Table~\ref{tab:nvs-gso-full}, with means and bootstrap intervals computed over scenes.}
    \label{tab:nvs-dl3dv-full}
    \setlength{\tabcolsep}{3.0pt}
    \renewcommand{\arraystretch}{1.12}
    \begin{tabularx}{\textwidth}{l*{4}{>{\centering\arraybackslash}X}}
        \toprule
        Model & PSNR $\uparrow$ & View 23 PSNR $\uparrow$ & SSIM $\uparrow$ & LPIPS $\downarrow$ \\
        \midrule
        \tgroup{fw}{5}{Layer-private pools ($p{=}8$)}
        \trow{fw} uTTT-Dense-w1-p8 (TTT-Dense) & 16.14 {\tiny[15.91, 16.39]} & 16.69 {\tiny[16.35, 17.03]} & 0.383 {\tiny[0.365, 0.402]} & 0.681 {\tiny[0.673, 0.690]} \\
        \trow{fw} uTTT-MoE-e4-p8 (TTT-MoE-e4) & 15.98 {\tiny[15.74, 16.22]} & 16.47 {\tiny[16.14, 16.81]} & 0.381 {\tiny[0.362, 0.399]} & 0.691 {\tiny[0.682, 0.699]} \\
        \trow{fw} uTTT-MoE-e8-p8 (TTT-MoE-e8) & 16.02 {\tiny[15.78, 16.27]} & 16.54 {\tiny[16.20, 16.89]} & 0.379 {\tiny[0.361, 0.397]} & 0.688 {\tiny[0.679, 0.696]} \\
        \tsep
        \tgroup{moeC}{5}{MoE, universal pool, top-1 (C)}
        \trow{moeC} uTTT-MoE-e8-p1 & 16.02 {\tiny[15.78, 16.27]} & 16.63 {\tiny[16.30, 16.98]} & 0.382 {\tiny[0.364, 0.401]} & 0.686 {\tiny[0.677, 0.694]} \\
        \trow{moeC} uTTT-MoE-e16-p1 & 16.01 {\tiny[15.77, 16.26]} & 16.68 {\tiny[16.35, 17.02]} & 0.382 {\tiny[0.364, 0.401]} & 0.686 {\tiny[0.678, 0.695]} \\
        \trow{moeC} uTTT-MoE-e32-p1 & 16.09 {\tiny[15.85, 16.34]} & 16.76 {\tiny[16.41, 17.11]} & 0.384 {\tiny[0.366, 0.403]} & 0.683 {\tiny[0.674, 0.692]} \\
        \trow{moeC} uTTT-MoE-e64-p1 & 16.12 {\tiny[15.88, 16.37]} & 16.79 {\tiny[16.44, 17.14]} & 0.385 {\tiny[0.367, 0.403]} & 0.682 {\tiny[0.673, 0.690]} \\
        \tsep
        \tgroup{moeA}{5}{MoE, depth partitions at fixed total (A, A$'$)}
        \trow{moeA} uTTT-MoE-e32-p2 & 16.06 {\tiny[15.83, 16.31]} & 16.73 {\tiny[16.39, 17.07]} & 0.383 {\tiny[0.365, 0.401]} & 0.685 {\tiny[0.676, 0.693]} \\
        \trow{moeA} uTTT-MoE-e16-p4 & 16.02 {\tiny[15.78, 16.27]} & 16.64 {\tiny[16.30, 16.99]} & 0.382 {\tiny[0.364, 0.401]} & 0.688 {\tiny[0.679, 0.696]} \\
        \trow{moeA} uTTT-MoE-e16-p2 & 16.07 {\tiny[15.84, 16.32]} & 16.68 {\tiny[16.34, 17.02]} & 0.382 {\tiny[0.363, 0.400]} & 0.685 {\tiny[0.677, 0.694]} \\
        \trow{moeA} uTTT-MoE-e8-p4 & 16.05 {\tiny[15.81, 16.31]} & 16.65 {\tiny[16.31, 16.99]} & 0.382 {\tiny[0.363, 0.400]} & 0.687 {\tiny[0.679, 0.696]} \\
        \tsep
        \tgroup{moeB}{5}{MoE, active width (B, B$'$)}
        \trow{moeB} uTTT-MoE-e64-p1-a2 & 16.45 {\tiny[16.20, 16.71]} & 17.18 {\tiny[16.82, 17.54]} & 0.394 {\tiny[0.376, 0.413]} & 0.668 {\tiny[0.659, 0.678]} \\
        \trow{moeB} uTTT-MoE-e64-p1-a4 & \textbf{16.68} {\tiny[16.43, 16.94]} & 17.47 {\tiny[17.09, 17.84]} & \textbf{0.401} {\tiny[0.383, 0.420]} & \textbf{0.659} {\tiny[0.650, 0.668]} \\
        \trow{moeB} uTTT-MoE-e64-p1-a8 & 16.66 {\tiny[16.41, 16.92]} & \textbf{17.47} {\tiny[17.10, 17.84]} & 0.400 {\tiny[0.382, 0.419]} & 0.660 {\tiny[0.651, 0.669]} \\
        \trow{moeB} uTTT-MoE-e32-p1-a2 & 16.38 {\tiny[16.13, 16.63]} & 17.13 {\tiny[16.78, 17.49]} & 0.393 {\tiny[0.375, 0.412]} & 0.674 {\tiny[0.665, 0.683]} \\
        \trow{moeB} uTTT-MoE-e32-p1-a4 & 16.52 {\tiny[16.27, 16.79]} & 17.32 {\tiny[16.95, 17.70]} & 0.397 {\tiny[0.378, 0.415]} & 0.666 {\tiny[0.657, 0.675]} \\
        \trow{moeB} uTTT-MoE-e32-p1-a8 & 16.63 {\tiny[16.38, 16.89]} & 17.40 {\tiny[17.04, 17.78]} & 0.399 {\tiny[0.381, 0.418]} & 0.661 {\tiny[0.652, 0.671]} \\
        \tsep
        \tgroup{denB}{5}{Dense, universal state (C)}
        \trow{denB} uTTT-Dense-w1-p1 & 16.17 {\tiny[15.93, 16.43]} & 16.81 {\tiny[16.47, 17.16]} & 0.386 {\tiny[0.368, 0.404]} & 0.683 {\tiny[0.674, 0.692]} \\
        \trow{denB} uTTT-Dense-w2-p1 & 16.22 {\tiny[15.98, 16.48]} & 16.85 {\tiny[16.51, 17.20]} & 0.387 {\tiny[0.369, 0.406]} & 0.680 {\tiny[0.671, 0.689]} \\
        \trow{denB} uTTT-Dense-w4-p1 & 16.36 {\tiny[16.11, 16.62]} & 17.00 {\tiny[16.66, 17.36]} & 0.392 {\tiny[0.373, 0.410]} & 0.676 {\tiny[0.666, 0.685]} \\
        \trow{denB} uTTT-Dense-w8-p1 & 16.45 {\tiny[16.21, 16.71]} & 17.15 {\tiny[16.80, 17.50]} & 0.394 {\tiny[0.376, 0.413]} & 0.669 {\tiny[0.660, 0.678]} \\
        \tsep
        \tgroup{denA}{5}{Dense, depth partitions (B, A)}
        \trow{denA} uTTT-Dense-w4-p2 & 16.28 {\tiny[16.04, 16.53]} & 16.97 {\tiny[16.62, 17.33]} & 0.390 {\tiny[0.372, 0.409]} & 0.676 {\tiny[0.666, 0.685]} \\
        \trow{denA} uTTT-Dense-w2-p4 & 16.20 {\tiny[15.96, 16.45]} & 16.77 {\tiny[16.43, 17.11]} & 0.386 {\tiny[0.367, 0.404]} & 0.679 {\tiny[0.670, 0.688]} \\
        \trow{denA} uTTT-Dense-w1-p2 & 16.13 {\tiny[15.89, 16.39]} & 16.78 {\tiny[16.44, 17.12]} & 0.385 {\tiny[0.367, 0.403]} & 0.683 {\tiny[0.674, 0.692]} \\
        \trow{denA} uTTT-Dense-w1-p4 & 16.22 {\tiny[15.98, 16.47]} & 16.80 {\tiny[16.46, 17.15]} & 0.385 {\tiny[0.367, 0.404]} & 0.680 {\tiny[0.671, 0.688]} \\
        \tsep
        \tgroup{ref}{5}{Baseline}
        \trow{ref} Transformer$^\dagger$ & 17.41 {\tiny[17.14, 17.68]} & 18.55 {\tiny[18.12, 18.99]} & 0.426 {\tiny[0.408, 0.445]} & 0.626 {\tiny[0.617, 0.636]} \\
        \trow{ref} Gated DeltaNet & 16.05 {\tiny[15.81, 16.29]} & 16.56 {\tiny[16.23, 16.89]} & 0.382 {\tiny[0.364, 0.400]} & 0.681 {\tiny[0.672, 0.690]} \\
        \trow{ref} DeltaNet & 15.82 {\tiny[15.59, 16.06]} & 16.29 {\tiny[15.96, 16.62]} & 0.376 {\tiny[0.358, 0.395]} & 0.689 {\tiny[0.680, 0.697]} \\
        \bottomrule
    \end{tabularx}
\end{table}

\subsection{Write Schedules}
\label{app:nvs-schedules}
We compare the 64-expert, $a{=}1$ shared pool under the two schedules of Appendix~\ref{app:fw-schedules} using full-test results on both datasets (Table~\ref{tab:nvs-aggwrite-efficiency}). At target view~23, the aggregated schedule has about $0.1$~dB lower PSNR on Objaverse/GSO and $0.01$~dB lower on DL3DV, trains $2.1\%$ faster and uses $26.3\%$ less peak memory on one H100; the canonical sweeps use it.

\begin{table}[!htbp]
    \centering
    \scriptsize
    \caption{\textbf{Measured cost and final quality of the two write schedules} at 64 experts, $a{=}1$. Aggregated-schedule quality is taken from Tables~\ref{tab:nvs-gso-full} and~\ref{tab:nvs-dl3dv-full}. View-23 PSNR is the final target view; \emph{min/1k steps} is wall-clock minutes per thousand optimizer steps. Cost measurements use one H100 80GB with the same $256{\times}256$ shape, 8 layers, 12 views, batch 32, and no activation checkpointing. Bold marks the best value per column.}
    \label{tab:nvs-aggwrite-efficiency}
    \setlength{\tabcolsep}{3.5pt}
    \begin{tabular*}{\textwidth}{@{\extracolsep{\fill}}lrrrrr@{}}
        \toprule
        Schedule & GSO V23 PSNR $\uparrow$ & DL3DV V23 PSNR $\uparrow$ & k patch tok/s/GPU $\uparrow$ & min/1k steps $\downarrow$ & Peak GiB $\downarrow$ \\
        \midrule
        Sequential schedule & \textbf{25.58} & \textbf{16.80} & 100.2 & 119.9 & 40.75 \\
        Aggregated schedule & 25.48 & 16.79 & \textbf{102.3} (+2.1\%) & \textbf{117.5} (-2.1\%) & \textbf{30.04} (-26.3\%) \\
        \bottomrule
    \end{tabular*}
\end{table}

\FloatBarrier

\endgroup
\section{Analysis of Shared Memory}
\label{app:gradient-routes}
We analyze the shared memory in three steps. Appendix~\ref{app:fw-recurrence} derives the mechanism: with a shared state, a read can depend on the writes of every layer in earlier write-enabled chunks, deeper layers included (Proposition~\ref{prop:reach}). Appendix~\ref{app:gr:validation} specifies the experiments of Figure~\ref{fig:core-evidence} that test these dependencies in trained models, and Appendix~\ref{app:redundancy} shows a behavior that sharing enables: repeated observations of a surface preferentially reuse the same experts across views and depths.

\subsection{Mechanism: recurrence and cross-depth reach}
\label{app:fw-recurrence}
\label{app:gr:proof}
Both write schedules of Appendix~\ref{app:fw-schedules} read the chunk-start state throughout each chunk. Once the state is shared, depth joins the time axis through a joint fast-weight state recurrence over (chunk, layer) pairs (Figure~\ref{fig:recurrence}a shows the aggregated case). Indexing the pairs in execution order by $t = (c-1)L + \ell$, the sequential schedule becomes
\begin{equation}
W^{(t)} \;=\; \operatorname{Update}\big(W^{(t-1)},\, G_t\big),
\label{eq:unrolled}
\end{equation}
a \emph{single} recurrence of $(N-1)L$ steps over the whole fast-weight memory in language modeling; the chunk-start state is read throughout each chunk. Step $t$ runs layer $\ell(t) = 1 + ((t-1) \bmod L)$ with that layer's own slow weights; tying them across layers would give a Universal-Transformer-style recurrence~\citep{dehghani2019universal}, which we do not require. Depth thus adds test-time optimization: in each chunk the shared memory takes $L$ steps under the sequential schedule, or one step that combines the contributions of every depth under the aggregated schedule, whereas a layer-private model advances each of its $L$ memories by one step. A write by layer $r$ in chunk $s$ can be read in the next chunk by any layer, including those below $r$; this opens the top-down route of Section~\ref{sec:method:recurrent} alongside the bottom-up one. Layer-private TTT is a strict special case: at $p = L$, each $G_t$ updates one private state block as in \eqref{eq:prelim-inner-loop}; intermediate $p$ gives $p$ group blocks. With producer inputs fixed, the direct state-transition Jacobian is block diagonal under either schedule. Query and write producers remain coupled across layers through the residual stream.

\paragraph{Full and memory-chain derivatives.}
Let $u_{s,r}$ be the write inputs (keys, values and step sizes) of layer $r$ in chunk $s$, and $o_{c,\ell}$ the fast-weight read of layer $\ell$ in chunk $c$. Nonzero-reach statements below assume that chunk $s$ is write-enabled. We write $\partial o_{c,\ell}/\partial u_{s,r}$ for the \emph{full} derivative: the response of the read to an independent perturbation added at the write inputs, with the graph that produces the queries and the other write inputs left connected. Figure~\ref{fig:recurrence}b draws a restricted part of it, the \emph{memory-chain} derivative, which holds the queries and predicted momentum coefficients at their forward values and treats the write inputs, step sizes included, as independent variables; the momentum buffers remain recurrent and differentiable. Let $S_{\mathrm{mem}}^{(j)}$ collect the fast weights, master accumulators and momentum buffers after chunk $j$, and $J_j=\partial S_{\mathrm{mem}}^{(j)}/\partial S_{\mathrm{mem}}^{(j-1)}$ with producer inputs fixed. Since $u_{s,r}$ enters the state through its contribution $G_{s,r}$ in \eqref{eq:sequential}--\eqref{eq:aggregated}, for write-enabled $s<c$
\begin{equation}
D^{\mathrm{mem}}_{u_{s,r}} o_{c,\ell}
\;=\;
\frac{\partial o_{c,\ell}}{\partial S_{\mathrm{mem}}^{(c-1)}}
\bigl(J_{c-1}\cdots J_{s+1}\bigr)
\frac{\partial S_{\mathrm{mem}}^{(s)}}{\partial G_{s,r}}\,
\frac{\partial G_{s,r}}{\partial u_{s,r}}.
\label{eq:route-chain}
\end{equation}
The product runs from the latest update to the earliest and is the identity when $c=s+1$. Every state Jacobian keeps the complete update dependence, including the inner gradient, momentum and renormalization, and the final factors describe how $G_{s,r}$ enters the boundary update: in one step under \eqref{eq:aggregated}, or at step $r$ followed by the remaining layer steps under \eqref{eq:sequential}. It is zero for $s\geq c$, because a chunk never reads its own writes, and it excludes paths that leave the memory chain through a hidden state, a query or a predicted coefficient, including paths that later re-enter the chain through another write. Unless the memory-chain derivative is named, the proposition below and the probe of Appendix~\ref{app:gr:reach} refer to the full derivative, which includes these paths. A historical write of a shallower layer can therefore receive gradient in a layer-private model even though its node is hollow in Figure~\ref{fig:recurrence}b; the figure depicts potential paths, whose existence does not establish their magnitude or their contribution to performance. Keeping the boundary update in the graph is backpropagation through time over chunks (Section~\ref{sec:method:recurrent}), and \eqref{eq:route-chain} is one of its terms.

\begin{proposition}[Reach of a read]\label{prop:reach}
\emph{(i)} In every model, $\partial o_{c,\ell}/\partial u_{s,r} = 0$ for $s \geq c$; the language-modeling sequence omits the final-chunk write. \emph{(ii)} With private memories ($p = L$), the derivative is identically zero for deeper writers, $r > \ell$; shallower writers can reach the read through paths that also traverse the residual stream. \emph{(iii)} With a universal pool ($p = 1$) and write-enabled $s < c$, under either schedule, the memory-chain derivative~\eqref{eq:route-chain} is generically nonzero for a dense pool, and for a routed pool when the writer and the reader select a common expert. Pooled momentum, when used, can add paths through its input-dependent coefficient.
\end{proposition}

\paragraph{Proof of Proposition~\ref{prop:reach}.}
A model with $L$ layers processes $N$ chunks. The memory state carried across a chunk boundary consists of fast weights $W$, master accumulators $U$ and, where used, momentum buffers $M$, updated as specified in Appendix~\ref{app:fw-update}; a layer-private model carries $L$ blocks $(W_r,U_r,M_r)$, and a universal model carries one pool. Within chunk $c$ every layer reads $W^{(c-1)}$. The write inputs $u_{c,r} = (k, v, \eta)$ of layer $r$ are computed from the layer's input hidden states and enter the computation only through $G_{c,r}$; the momentum coefficient, when used, is predicted from each layer's inputs and averaged over the layers of a pool. Information crosses a chunk boundary only through the memory state and, in language modeling, through each layer's sliding-window attention cache; within a chunk it moves upward through the residual stream.

\paragraph{(i)} The write inputs $u_{s,r}$ have a single child, $G_{s,r}$, which enters the state at the boundary after chunk $s$. A read in chunk $c \leq s$ is a function of $W^{(c-1)}$, with $c - 1 < s$, and of chunk-$c$ hidden states that do not involve $G_{s,r}$; hence $\partial o_{c,\ell} / \partial u_{s,r} = 0$ for $s \geq c$. In the language-modeling sequence, no boundary update is computed after the last chunk, so $u_{N,r}$ has no child at all.

\paragraph{(ii)} Layer-private, $p = L$. Along every edge of the forward graph the layer index is non-decreasing: within a chunk the residual stream carries the output of layer $j$ to the layers above it, and across chunks the memory block $(W_j,U_j,M_j)$ and the attention cache of layer $j$ are written and read by layer $j$ alone. A path from $u_{s,r}$ therefore never reaches a quantity of a layer $\ell < r$, and $\partial o_{c,\ell} / \partial u_{s,r} = 0$ for $r > \ell$ whatever the weights and the input. For $r < \ell$ the memory-chain path is closed as well, since $G_{s,r}$ does not enter the update of $W_\ell$, but the residual stream is open: $u_{s,r}$ enters $W_r^{(s)}$, and its effect can persist in layer $r$'s memory through subsequent updates or read-only chunks. In any later chunk $c>s$, layer $r$ can read this state and pass the effect up the residual stream to $o_{c,\ell}$. At an intervening write-enabled chunk, the effect may also enter layer $\ell$'s own memory through its write inputs. These derivatives are generically nonzero.

\paragraph{(iii)} Universal, $p = 1$. Under \eqref{eq:aggregated}, $W^{(s)} = \operatorname{Update}\big(W^{(s-1)}, \sum_r G_{s,r}\big)$ depends on every $G_{s,r}$; under \eqref{eq:sequential}, $G_{s,r}$ enters step $r$, and the later steps carry its effect to $W^{(s)}$. Because $W^{(s')}$ is a function of $W^{(s)}$ for $s' > s$, $W^{(c-1)}$ is a function of $u_{s,r}$ for every layer $r$ and every write-enabled chunk $s < c$. A layer reading a dense pool uses all $H$ head blocks, while each head reads its corresponding block; neither uses expert routing. The layer-level read-state Jacobian in \eqref{eq:route-chain} therefore has support across these blocks, and the memory-chain product is nonzero for generic weights and inputs. In a routed pool, $G_{s,r}$ touches the experts that layer $r$'s write router selects in chunk $s$, and $o_{c,\ell}$ reads the experts that layer $\ell$'s read router selects, so the memory-chain path is open exactly when the two selections share an expert. Cross-depth memory reuse does not require momentum. When momentum is used, its pooled coefficient depends on the input and adds a path through the coefficient predictor. The coefficient is predicted from every layer's inputs, averaged over the layers sharing the pool and applied to every expert's momentum buffer at the boundary. A write of layer $r$ in chunk $s$ that changes the inputs of the layers above $r$ in chunk $s+1$ therefore changes the coefficient at boundary $s+1$, and with it the state of every expert from chunk $s+2$ on; Appendix~\ref{app:gr:reach} measures both paths. In language modeling, the structural maxima are $(c-1)\,\ell$ write cells $\{(s, r) : s < c,\ r \leq \ell\}$ for a layer-private read and $(c-1)\,L$ cells $\{(s, r) : s < c\}$ for a universal read; realized reach in routed pools also requires active routes. The difference, $r > \ell$, identifies the potential top-down routes. $\square$

\subsection{Validation: probe, training control and interventions}
\label{app:gr:validation}
\label{app:gr:reach}
\label{app:gr:cut}
\label{app:tp:knockout}
Proposition~\ref{prop:reach} states which dependencies can exist; Figure~\ref{fig:core-evidence} tests them in trained models with a read-output gradient probe (panel a), a training control that cuts gradients through other-layer write contributions while keeping the forward pass (panel b), and inference-time interventions that remove writes from fixed checkpoints (panels b and c). The language-model diagnostics use the first 32K tokens of single documents from the $3{,}000$-book Books3 cohort of Appendix~\ref{app:extrapolation}, without packing.

\paragraph{Read-output gradient probe.}
We add independent zero-valued perturbations at the write inputs $(k, v, \eta)$ of every (chunk, layer) cell, without detaching the graph that produces them. The probe is a fixed random projection of layer $\ell$'s fast-weight read in the last chunk, taken before the TTT scale and the output projection and summed over the last 32 positions. Its coefficients are independent Rademacher signs ($\pm1$ with equal probability), sampled once and shared across documents and models. For each cell we sum the squared gradient of the probe over the cell's write inputs and report it as a share of the total over all cells. The box in Figure~\ref{fig:core-evidence}a contains the cells of the layers above $\ell$ in earlier chunks, the top-down routes into the read; a boxed cell counts as reached when its share is nonzero in every document. Figure~\ref{fig:core-evidence}a applies the $\ell = 1$ probe to the 124M TTT-MoE and uTTT-MoE (12 layers, 8 chunks of 4{,}096 tokens). It uses 300 documents: 32 evenly spaced ones plus 268 drawn without replacement from the remaining 2{,}968.

\paragraph{Probe checks.} Randomly initialized six-layer probe models serve as checks. Finite-difference perturbations confirm the tested zero cells of the layer-private models, and routing controls in which no two layers share an expert isolate the additional paths through pooled momentum.

\paragraph{Cross-layer gradient cut.}
The training intervention gives each reading layer $\ell$ its own backward stream of the pool, including master accumulators and momentum buffers. All streams share the learned initialization. Writer $r$ computes $G_{s,r}$ at its own stream's chunk-start state $W^{(s-1)}_{\langle r\rangle}$; its predicted momentum coefficient is $\mu_{s,r}$ from \eqref{eq:fw-momentum-predictor}. Stream $\ell$ then uses
\begin{equation}
\begin{aligned}
\mu^{(s)}_{\langle\ell\rangle}
&=\frac{1}{L}\Big(\mu_{s,\ell}+\textstyle\sum_{r\neq\ell}\operatorname{sg}(\mu_{s,r})\Big),\\
W^{(s)}_{\langle\ell\rangle}
&=\operatorname{Update}\Big(W^{(s-1)}_{\langle\ell\rangle},\;
G_{s,\ell}+\textstyle\sum_{r\neq\ell}\operatorname{sg}(G_{s,r});\;
\mu^{(s)}_{\langle\ell\rangle}\Big).
\end{aligned}
\label{eq:cross-layer-cut}
\end{equation}
Here $\operatorname{sg}$ is the stop-gradient, and the argument after the semicolon makes the momentum coefficient of the stateful update explicit. Layer $\ell$ reads $W^{(c-1)}_{\langle\ell\rangle}$ in chunk $c$. In the forward pass $\operatorname{sg}$ is the identity, and evaluation preserves the original writer order and reductions, so every stream equals the full shared state and the logits are unchanged. In backward, the update retains the gradient to each stream's previous state and routes current contribution gradients only to its own $G_{s,\ell}$ and to its own coefficient with the mean's factor $1/L$. Stopping $G_{s,r}$ for $r \neq \ell$ removes both its write-input derivative and its state derivative $\partial G_{s,r}/\partial W$, changing the transition Jacobians. Same-layer and residual-stream paths remain available, but gradients to earlier same-layer writes can change magnitude. We call \eqref{eq:cross-layer-cut} the \emph{cross-layer gradient cut}. Stopping $G$ alone would leave the additional producer paths through input-dependent pooled momentum; the implemented cut also stops the other layers' coefficient contributions. On the probe models the tested cut preserves the logits exactly and removes the gradient from every boxed cell while keeping the reading layer's own row.

\paragraph{Training controls.} We train the 124M uTTT-MoE configuration of Table~\ref{tab:lm-results} with the cross-layer gradient cut and call the result Cut. It keeps the configuration, batch shape (one node of eight GPUs, four sequences per GPU), $10{,}240$ steps and evaluation stream; only the backward pass differs.

\paragraph{Inference route knock-outs.}
A route is one (writer, reader) pair of layers. To remove routes, each reading layer receives its own copy of the pool. At each write-enabled boundary, writer $r$ computes $G_{s,r}$ once at the chunk-start state of the copy it reads; each reader's copy then receives the sum of these contributions over the writers permitted by $\mathrm{allowed}[\text{reader},\text{writer}]$. In language modeling, its momentum coefficient is recomputed as the mean over those same allowed writers; NVS omits momentum (Appendix~\ref{app:fw-update}). Thus reading only own writes uses the reader's own contribution and, in language modeling, its own coefficient. In the language-model no-write condition, every copy has no allowed writers, skips the entire update including its momentum buffers, and remains at the learned initialization. Allowing every route reproduces the unmodified model exactly. The trained parameters stay fixed while the permitted writes and, in language modeling, their pooled coefficient determine each copy's forward updates.

\paragraph{Language modeling.} Figure~\ref{fig:core-evidence}b applies two conditions to the Full and Cut checkpoints: freezing state writes, which disables every boundary update of the pool, and reading only own writes, which removes every cross-layer route. We use the first $1{,}024$ documents of the cohort. The loss increase of a condition is measured against the same checkpoint's unmodified predictions: we average the loss within each 4K chunk, then over chunks 2--8, then over documents. Chunk 1 reads only the initial state and is identical in every condition. The unmodified tail PTL of Full, Cut and TTT is averaged over positions 30{,}000--32{,}766 of the same documents.

\paragraph{View synthesis.} Figure~\ref{fig:core-evidence}c removes the 64 writer--reader routes of uTTT-MoE-e64-p1 one at a time, including the eight same-layer routes on the diagonal, and reports the mean paired PSNR drop, unmodified minus knock-out, over the $1{,}019$ GSO objects and their 23 target views.

\subsection{Behavior: expert reuse by redundant tokens}
\label{app:redundancy}
\suppressfloats[t]   %
\begin{figure}[tb]
    \centering
    \includegraphics[width=\textwidth]{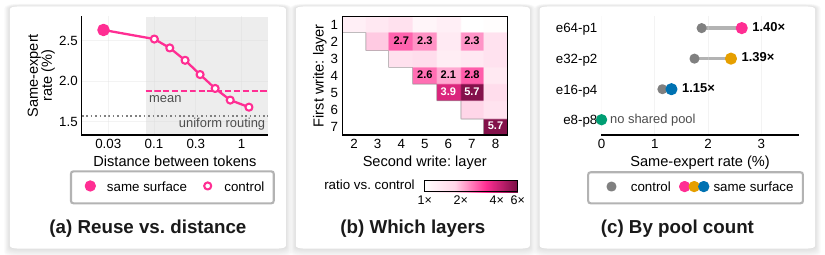}
    \caption{\textbf{Quantitative reuse of shared experts by redundant tokens} (GSO, $1{,}019$ objects, $8.6$M token pairs). All panels compare the write routing of an earlier-view token at one layer with that of a later-view token at a deeper layer, over 28 layer pairs and 8 heads.
    \emph{(a)}~Same-expert rate versus 3D Euclidean distance for e64-p1; objects are scaled to a longest bounding-box side of 1.8. Filled: geometry-matched pairs; open: controls on the object surface from the same later view, at distance $\geq0.08$, grouped by distance. Dashed: pooled control rate; dotted: uniform routing ($1/64$).
    \emph{(b)}~Same-surface/control ratio for each layer pair in e64-p1, maximum over the 8 heads. Colors are logarithmic; only ratios $\geq2$ are labeled.
    \emph{(c)}~Same-surface and control rates for four sharing partitions, with their ratios. Rates use all 28 layer pairs, including pairs in different pools that cannot share a physical expert.}
    \label{fig:redundancy-reuse}
\end{figure}

\begin{figure}[tb]
    \centering
    \includegraphics[width=\textwidth]{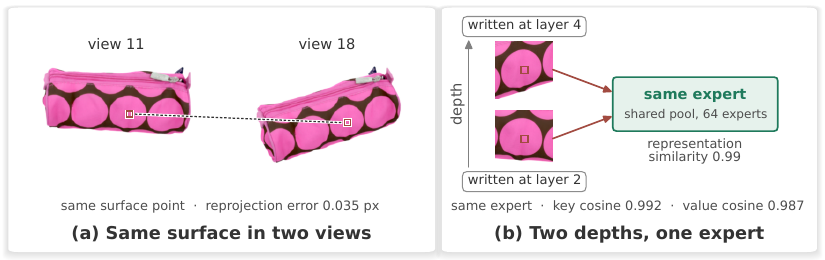}
    \caption{\textbf{A redundant token pair reuses one physical expert across depths} (uTTT-MoE-e64-p1, one GSO object). \emph{(a)}~Two source views of the object; the red squares mark two tokens that see the same surface point (mean reprojection error $0.035$ source pixels). \emph{(b)}~The same two tokens at the two depths compared: the write of the view-11 token at layer 2 and that of the view-18 token at layer 4. Red arrows mark these writes; both are routed to the same expert of the 64-expert shared pool. Compared at the same layer and head and averaged over the 8 layers and 8 heads, their keys and values have cosine similarities of $0.992$ and $0.987$, a mean representation similarity of $0.99$.}
    \label{fig:redundancy-example}
\end{figure}

We now turn from dependence to behavior: are repeated observations of the same surface routed to the same shared experts across views and depths? Two source views that see the same surface region produce geometry-matched tokens; we call such a pair \emph{redundant}. We report population statistics over all GSO objects, then one example pair. Over the $1{,}019$ objects, $8.6$M same-surface pairs are compared at 224 (layer pair, head) combinations each: the earlier view's token, written at some layer, against the later view's token written at a deeper layer. In uTTT-MoE-e64-p1 the two are routed to the same physical expert in $2.63\%$ of the comparisons, against $1.88\%$ for a control token from the same later view at least one token spacing away from the first token's surface point, and $1.56\%$ for uniform routing over 64 experts. The same-surface/control ratio is $1.40$ [$1.39$, $1.41$] (95\% interval from $10{,}000$ object resamples) and exceeds 1 on every object.

Figure~\ref{fig:redundancy-reuse}a shows that the same-expert rate decreases with the 3D distance between the two tokens, and panel b shows which layer pairs have strong per-head enrichment. Across sharing partitions (panel c), the same-surface/control ratio is $1.40\times$, $1.39\times$ and $1.15\times$ for e64-p1, e32-p2 and e16-p4, respectively; the layer-private e8-p8 shares no physical expert across layers. The absolute rates include all 28 layer pairs, so their decrease also reflects that fewer pairs share a pool. The later token's read routing, taken while its view is rendered from camera rays alone and before its image is written, also favors the earlier token's write experts, with a same-surface/control ratio of $1.34$ in e64-p1. These statistics measure how often repeated observations are routed together; Figure~\ref{fig:core-evidence}c separately tests how prediction depends on individual state-read routes. Figure~\ref{fig:redundancy-example} shows one such pair.

\FloatBarrier

\end{document}